\documentclass[journal]{IEEEtran}

\usepackage{stfloats}
\usepackage{amsmath}
\usepackage{amssymb,latexsym}
\usepackage{graphicx}
\usepackage{color,cite,times}
\usepackage{epstopdf}
\usepackage{algorithmic}
\usepackage[ruled,linesnumbered] {algorithm2e}
\usepackage{booktabs}
\usepackage{bbm} 
\usepackage{url}
\usepackage{fancyhdr}
\usepackage{verbatim}
\usepackage{multirow}
\usepackage{makecell}
\usepackage{graphicx}
\usepackage{indentfirst}
\usepackage{cases}
\newcommand{\RNum}[1]{\uppercase\expandafter{\romannumeral #1\relax}}
\usepackage{colortbl} 
\usepackage{extarrows}
\usepackage{threeparttable}
\usepackage{diagbox}
\usepackage{pifont}
\usepackage{cancel}
\usepackage{framed}
\usepackage{overpic}
\usepackage{mathtools} 
\usepackage{hyperref}
\hypersetup{
	colorlinks=true,
	linkcolor=blue,
	citecolor=blue
}
\usepackage{cleveref}
\usepackage{subfigure}
\usepackage{booktabs}

\usepackage[framemethod=tikz]{mdframed}
\usepackage{lipsum}
\usepackage[nolist]{acronym}
\begin{acronym}

\acro{PC}{predictor-corrector}
\acro{CT}{computed tomography}
\acro{SAR}{synthetic aperture radar}
\acro{MRI}{magnetic resonance imaging}
\acro{NFE}{neural function evaluation}

\acro{w.r.t.}{with respect to}

\acro{AMP}{approximate message passing}
\acro{D-AMP}{denoising-based AMP}
\acro{D-Turbo-CS}{denoising-based Turbo-CS}
\acro{BM3D-prGAMP}{BM3D phase-retrieval GAMP}
\acro{LDAMP}{learned D-AMP}
\acro{VAMP}{vector AMP}
\acro{OAMP}{orthogonal AMP}
\acro{BP}{belief propagation}
\acro{CS}{compressive sensing}
\acro{TV}{total variation}

\acro{VAE}{variational autoencoder}
\acro{GAN}{generative adversarial network}
\acro{MAMP}{memory AMP}
\acro{GAMP}{generalized AMP}

\acro{EP}{expectation propagation}
\acro{EC}{expectation consistent}
\acro{MSE}{mean-squared error}
\acro{MMSE}{minimum mean-squared error}
\acro{NMSE}{normalized MSE}
\acro{LMMSE}{linear MMSE}
\acro{SVD}{singular value decomposition}
\acro{KL}{Kullback-Leibler}
\acro{KKT}{Karush-Kuhn-Tucker}
\acro{AWGN}{additive white Gaussian noise}
\acro{ROI}{right-orthogonally invariant}
\acro{FFT}{fast Fourier transform}
\acro{DCT}{discrete cosine transform}
\acro{DFT}{discrete Fourier transform}
\acro{i.i.d.}{independent and identically distributed}
\acro{SE}{state evolution}
\acro{NFEs}{neural function evaluations}

\acro{STMP}{score-based turbo message passing}
\acro{Q-STMP}{quantized STMP}
\acro{ADCs}{analog-to-digital converters}
\acro{SDE}{stochastic differencial equation}
\acro{VE}{variance exploding}
\acro{GLMs}{generalized linear models}
\acro{SE}{state evolution}
\acro{TMP}{turbo message passing}

\acro{DnCNN}{denoising convolutional neural network}
\acro{E2E}{end-to-end}

\acro{PDF}{probability density function}
\acro{CDF}{cumulative distribution function}

\acro{MAP}{maximum \textit{a posteriori}}
\acro{FISTA}{fast iterative shrinkage-thresholding algorithm}
\acro{ADMM}{alternating direction method of multipliers}
\acro{RED}{regularization by denoising}
\acro{HQS}{half-quadratic splitting}
\acro{GTurbo-SR}{generalized turbo signal recovery}
\acro{PnP}{plug-and-play}

\acro{DPS}{diffusion posterior sampling}
\acro{DiffPIR}{diffusion models for PnP image restoration}
\acro{DDRM}{denoising diffusion restoration models}
\acro{DMPS}{diffusion model based posterior sampling}
\acro{MCG}{manifold constrained gradient}
\acro{Score-ALD}{score-based annealed Langevin dynamics}
\acro{Score-MI}{score-based medical imaging}
\acro{ILVR}{iterative latent variable refinement}
\acro{DDIM}{denoising diffusion implicit models}
\acro{ODE}{ordinary differential equation}

\acro{CG}{conjugate gradient}

\acro{MPDQ}{message-passing de-quantization}
\acro{QSP}{quantized subspace pursuit}
\acro{QIHT}{quantized iterative hard thresholding}
\acro{QCoSaMP}{quantized compressive sampling matching pursuit}

\acro{CNN}{convolutional neural network}

\acro{RMP}{reverse mean propagation}
\end{acronym}

\newcommand{\bsm}{\boldsymbol}

\newcommand{\diag}{\mathrm{diag}}

\newtheorem{theo}{Theorem}

\newtheorem{remk}{Remark}

\newtheorem{assump}{Assumption}

\definecolor{gray}{RGB}{192,192,192}

\begin{document}
\title{Score-Based Turbo Message Passing for Plug-and-Play Compressive Imaging}

\author{Chang Cai, \IEEEmembership{Member, IEEE,}
	Hao Jiang, \IEEEmembership{Graduate Student Member, IEEE,} \\ 
	Xiaojun Yuan, \IEEEmembership{Fellow, IEEE,} 
	and Ying-Jun Angela Zhang, \IEEEmembership{Fellow, IEEE} 
		\thanks{
			Received 16 December 2025; revised 13 May 2026; accepted 29 July 2026.
			This work was supported in part by the National Natural Science Foundation of China under Grant 62571087;
			in part by the National Key Laboratory of Wireless Communications Foundation under Grant IFN202507;
			in part by the General Research Fund from the Research Grants Council of Hong Kong under Project 14214122, Project 14202723, and Project 14207624;
			in part by the Area of Excellence Scheme Grant from the Research Grants Council of Hong Kong under Project AoE/E-601/22-R;
			and in part by the NSFC/RGC Collaborative Research Scheme from the Research Grants Council of Hong Kong under Project CRS\_HKUST603/22 and Project CRS\_HKU702/24.
			An earlier version of this paper was presented in part at the IEEE 26th International Workshop on Signal Processing and Artificial Intelligence for Wireless Communications (SPAWC), 2025 \cite{cai2025spawc}.
			The associate editor coordinating the review of this article and approving it for publication was Thomas Oberlin.
			\textit{(Corresponding author: Xiaojun Yuan.)}
			
			Chang Cai was with the Department of Information Engineering, The Chinese University of Hong Kong, Hong Kong SAR.
			He is now with the Department of Electrical and Computer Engineering, The University of Hong Kong, Hong Kong SAR (e-mail: changcai@hku.hk).

			Hao Jiang and Xiaojun Yuan are with the National Key Laboratory of Wireless Communications,
			University of Electronic Science and Technology of China, Chengdu 611731, China (e-mail: jh@std.uestc.edu.cn; xjyuan@uestc.edu.cn). 
			
			Ying-Jun Angela Zhang is with the Department of Information Engineering, The Chinese University of Hong Kong, Hong Kong SAR (e-mail: yjzhang@ie.cuhk.edu.hk).
			
			Code is available online at https://github.com/chang-cai/STMP
		}
	}
\maketitle

\begin{abstract}
	Message-passing algorithms have been adapted for compressive imaging by incorporating various off-the-shelf image denoisers.
	However, these denoisers rely largely on generic or hand-crafted priors and often fall short in accurately capturing the complex statistical structure of natural images.
	As a result, traditional plug-and-play (PnP) methods often lead to suboptimal reconstruction, especially in highly underdetermined regimes.
	Recently, score-based generative models have emerged as a powerful framework for accurately characterizing sophisticated image distributions.
	Yet, their direct use for posterior sampling typically incurs prohibitive computational complexity.
	In this paper, by exploiting the close connection between score-based generative modeling and empirical Bayes denoising, we devise a message-passing framework that integrates a score-based minimum mean-squared error (MMSE) denoiser for compressive image recovery.
	The resulting algorithm, named score-based turbo message passing (STMP), combines the fast convergence of message passing with the expressive power of score-based generative priors.
	For practical systems with quantized measurements, we further propose quantized STMP (Q-STMP), which augments STMP with a component-wise MMSE dequantization module.
	\color{black}
	We present state evolution (SE) predictions for the asymptotic performance of STMP and Q-STMP.
	Experiments on the FFHQ and LDCT datasets under various sensing models demonstrate that STMP strikes a significantly better performance--complexity tradeoff compared with competing baselines, and that \mbox{Q-STMP} remains robust even under 1-bit quantization.
	\color{black}
	Remarkably, both STMP and \mbox{Q-STMP} typically converge within 20 iterations.
\end{abstract}

\section{Introduction}
Compressive image recovery~\cite{from_denoising_to_cs} refers to the reconstruction of high-dimensional images from a reduced set of (possibly linear) measurements.
This paradigm underlies a broad range of practical systems including \ac{MRI}~\cite{vlaardingerbroek2013magnetic}, \ac{SAR}~\cite{sar2013magzine}, and \ac{CT}~\cite{buzug2011computed}, where physical constraints or hardware limitations restrict the number of available measurements.
As a result, the associated inverse problem is highly underdetermined.
Successful image recovery thus requires incorporating strong prior knowledge of the underlying image to regularize the solution space.

Message passing provides a highly efficient paradigm for Bayesian \ac{CS}, enabling the principled incorporation of various sparsity-inducing priors.
For linear inverse problems, representative algorithms include~\ac{AMP}, \ac{TMP}~\cite{turbo_cs} (a.k.a. Turbo-CS), \ac{VAMP}~\cite{vamp}, and \ac{OAMP}~\cite{oamp}.
These algorithms have also been extended beyond the linear setting, most notably through \ac{GAMP}~\cite{gamp}, which generalizes the framework to a broad class of~\ac{GLMs}.
However, since natural images do not have an exact sparse representation in any known transform domain, these sparsity-based approaches often exhibit poor performance in many imaging applications.
To address this limitation, off-the-shelf image denoisers have been plugged into message-passing algorithms to enhance reconstruction performance, giving rise to methods such as \ac{D-AMP}\cite{from_denoising_to_cs}, \ac{D-Turbo-CS}~\cite{xue2017access}, and \ac{BM3D-prGAMP}~\cite{bm3d_prgamp2016icip}.
Commonly used denoisers include \ac{TV}, SURE-LET~\cite{sure-let}, and BM3D~\cite{bm3d}.
These off-the-shelf denoisers are non-parametric or heuristic in nature and do not rely on an explicit statistical model of the underlying image.
While effective in practice, such \ac{PnP} methods still fall short in fully exploiting the rich and highly structured priors inherent in natural images.
To benefit from deep learning for more accurate prior modeling, methods such as \ac{LDAMP}~\cite{metzler2017learned_damp} and AMP-Net~\cite{amp_net2021tip} unroll message-passing iterations into layer-wise trainable architectures, in which traditional denoisers are replaced by \ac{CNN}-based modules.
However, these approaches typically require training separate end-to-end models for different sampling ratios and noise levels, thereby limiting their flexibility and generalizability.

Recently, score-based generative models~\cite{song2019generative, score_sde} (a.k.a. diffusion models~\cite{ho2020denoising}) have achieved unparalleled success in generating high-quality samples across a wide range of domains, including images, videos, and molecular conformations.
These results highlight their exceptional capacity to capture the complex structure of high-dimensional data distributions.
During training, score-based generative models corrupt the data with random Gaussian noise of different magnitudes, and learn to approximate the gradient of the log-density of the perturbed data, known as the score function.
The score function is then applied at each reverse-diffusion step for sample generation.
Score-based generative models can also be leveraged for solving inverse problems without task-specific training \cite{score_ald2021nips}, \cite{ilvr2021iccv}, \cite{dmps}, \cite{kawar2022denoising}, \cite{mcg}, \cite{dps}, \cite{song2023pseudoinverse}, \cite{diffpir}.
These approaches combine a prior term provided by the learned score function with a likelihood term derived from the degradation model, forming a Bayesian posterior via Bayes' rule.
Sampling from this posterior enables high-quality reconstructions across a variety of imaging tasks.
However, these methods typically require traversing the full reverse-diffusion trajectory, often involving hundreds or even thousands of \ac{NFEs}, which results in substantial computational cost.

To address the above limitations, we propose integrating score-based generative priors into the message-passing framework for compressive imaging.
This hybrid approach inherits the fast convergence and computational efficiency of message passing while fully leveraging the expressive power of learned score networks to capture complex image priors.
Specifically, we exploit the fact that, at each iteration, message-passing algorithms model the intermediate estimate as an \ac{AWGN} observation of the ground-truth image.
This enables the direct use of the score network for denoising via Tweedie's formula~\cite{efron2011tweedie}, with only one forward pass needed per iteration.
In addition, we employ a second-order score network~\cite{high_order_gradients, lu2022maximum} that outputs the diagonals of the log-density Hessian, allowing the calculation of the denoised variance required for message passing.
These components together facilitate a turbo-type message-passing algorithm with score-based denoising, named \ac{STMP}.
Owing to its \ac{PnP} structure, \ac{STMP} naturally adapts to different sampling ratios and measurement noise levels without requiring retraining.
In practice, measurements are often quantized due to the finite precision of \ac{ADCs} and data storage constraints.
Motivated by \ac{EP}~\cite{EP2001Minka} and its application to \ac{GLMs}~\cite{jiangzhu2018spl}, we extend \ac{STMP} to handle quantized measurements, resulting in the \ac{Q-STMP} algorithm.
\Ac{Q-STMP} iteratively exchanges messages between \ac{STMP} and an additional dequantization module, each producing extrinsic estimates that serve as inputs to the other.
Moreover, we characterize the asymptotic \ac{MSE} performance of \ac{STMP} and \ac{Q-STMP} via \ac{SE}.
The main contributions and results of this paper are summarized as follows.
\color{black}
\begin{itemize}
	\item
	We propose \ac{STMP}, a turbo-type message-passing algorithm for compressive imaging that embeds score-based generative priors into the \ac{TMP} framework.
	The key algorithmic innovation is to exploit the denoising interpretation of \ac{TMP} and implement it efficiently using learned first- and second-order score functions.
	Under standard assumptions, we use \ac{SE} to characterize the asymptotic \ac{MSE} performance of \ac{STMP} and discuss the practical approximations that may cause mismatched predictions.
	
	\item We develop \ac{Q-STMP} for quantized compressive image recovery by augmenting \ac{STMP} with a component-wise \ac{MMSE} dequantization module.
	\ac{Q-STMP} iteratively refines the dequantized signal estimate and the reconstructed image through extrinsic information exchange.
	Under standard assumptions, we further provide the corresponding \ac{SE} recursions to characterize the asymptotic \ac{MSE} performance of \ac{Q-STMP}.
	
	\item Extensive experiments on the FFHQ~\cite{ffhq} and LDCT~\cite{ldct2021} datasets demonstrate the effectiveness and generality of the proposed methods under diverse sensing models.
	\ac{STMP} achieves a favorable performance--complexity tradeoff against various baselines, with strong visual fidelity on natural images and reliable anatomical preservation in highly compressed medical imaging settings.
	For quantized compressive imaging, \ac{Q-STMP} delivers faithful reconstructions across different bit depths and remains robust even in the challenging 1-bit setting.
	Notably, both algorithms typically converge in fewer than 20 iterations.
\end{itemize}
\color{black}

The rest of this paper is organized as follows.
Section~\ref{sec:related_work} reviews the related work.
Section~\ref{sec:tmp_framework} introduces the basics of the \ac{TMP} framework.
In Section~\ref{sec:learning_score_priors}, we establish the connection between score functions and \ac{MMSE} denoising, and describe the learning of the first- and second-order score models.
Section~\ref{sec:stmp_algorithm} presents the proposed \ac{STMP} algorithm and its \ac{SE} characterization.
In Section~\ref{sec:q_stmp}, we extend \ac{STMP} to quantized systems, developing the \ac{Q-STMP} algorithm and deriving its corresponding \ac{SE} equations.
Experimental results on compressive and quantized compressive imaging are provided in Section~\ref{sec:experiments}.
Finally, Section~\ref{sec:conclusion} concludes the paper.

\section{Related Work}\label{sec:related_work}
\subsection{Optimization-Based \ac{PnP} Methods}
A large body of prior work approaches the \ac{MAP} solution from an optimization perspective by formulating the linear inverse problem as a regularized least-squares program and solving it via proximal methods.
Building on the denoising interpretation of \ac{ADMM}, the \ac{PnP} framework replaces explicit proximal operators with powerful image denoisers, leading to PnP-ADMM~\cite{pnp_admm_globalsip} and numerous related variants.
Subsequent studies have analyzed convergence properties~\cite{pnp_admm}, proposed parameter-free solutions~\cite{parameter_free_admm}, and extended \mbox{PnP-ADMM} to online and large-scale imaging settings~\cite{online_pnp}.
In addition, \ac{PnP} versions of the \ac{HQS} algorithm and other variable-splitting schemes have been developed, exemplified by the PnP-HQS framework~\cite{zhang2017learning} and its extensions using deep denoisers~\cite{zhangkai2022tpami}.
In parallel, the \ac{RED} framework~\cite{romano2017little} explicitly constructs a regularizer whose gradient is linked to a chosen denoiser, allowing efficient implementation via first-order optimization methods.
These formulations give rise to algorithms such as RED by fixed-point projection (\mbox{RED-PRO})~\cite{cohen2021regularization} and RED-ADMM~\cite{red_clarifications}.

\subsection{Diffusion-Based \ac{PnP} Methods}
In diffusion-based inverse solvers, sampling from the exact posterior distribution is intractable due to the time dependence of the likelihood.
This challenge has motivated a variety of approximation strategies.
\Ac{Score-ALD}~\cite{score_ald2021nips} enforces data consistency via a matched-filter-like approximation of the likelihood score at each step.
\Ac{ILVR}~\cite{ilvr2021iccv} and pseudoinverse-guided diffusion models ($\Pi$GDM)~\cite{song2023pseudoinverse} incorporate the Moore-Penrose pseudoinverse of the measurement operator to guide the reverse-diffusion trajectory toward the observation-consistent subspace.
\Ac{DMPS}~\cite{dmps} employs a \ac{LMMSE}-type approximation of the likelihood score.
\Ac{RMP}~\cite{xue2025rmp} optimizes the reverse \ac{KL} divergence with natural gradient descent and propagates the mean at each reverse step.
These approaches typically introduce additional hyperparameters that are sensitive and require careful tuning.
The approximation errors accumulate over iterations and can potentially drive the trajectory away from the underlying data manifold.
To mitigate this issue, \ac{MCG}~\cite{mcg} introduces an additional correction term that can be seamlessly integrated with existing solvers to steer the iterates back toward the data manifold.
\Ac{DPS}~\cite{dps} generalizes the above work to nonlinear inverse problems, which requires backpropagation through the unconditional score network to compute likelihood gradients.
\color{black}
Moreover, \ac{Score-MI}~\cite{score_mri} develops a posterior sampling solver tailored for \ac{CT} and \ac{MRI} tasks.
\color{black}
Most of the above methods require hundreds to thousands of reverse-diffusion steps, resulting in significant computational burden.
\Ac{DDRM}~\cite{kawar2022denoising} alleviates this cost by employing \ac{DDIM}~\cite{ddim2021iclr}, an \ac{ODE}-based accelerated sampler, enabling high-quality reconstructions with as few as 20 NFEs.
DiffPIR~\cite{diffpir} operates in an optimization-inspired \ac{PnP}-\ac{HQS} framework.
At each iteration, it heuristically injects Gaussian noise following diffusion principles to improve flexibility, which in turn often slows convergence.
Meanwhile, variational-inference approaches such as RED-diff~\cite{red_diff2024iclr} reinterpret diffusion priors through the lens of the \ac{RED} framework, offering a complementary optimization-based perspective.

\subsection{Quantized Compressive Sensing}
Early efforts in quantized compressed sensing focused on adapting classical sparse recovery algorithms to low-bit measurements, leading to methods such as \ac{QSP}, \ac{QIHT}, and \ac{QCoSaMP}, as surveyed in~\cite{shi2016methods}.
A seminal contribution from the message-passing perspective is \ac{MPDQ}~\cite{mpdq2012tsp}, which incorporates regular and non-regular scalar quantizers into the \ac{GAMP} framework.
Building on the turbo principle, \ac{GTurbo-SR}~\cite{GTurboSR} adapts Turbo-CS to quantized measurements by introducing an additional component-wise \ac{MMSE} denoiser tailored to the quantization likelihood.
A further unifying perspective is offered by the Bayesian inference framework in~\cite{jiangzhu2018spl}, which generalizes \ac{EP} to arbitrary \ac{GLMs}.
Recently, the advent of score-based generative models has inspired new approaches to quantized compressed sensing.
\mbox{QCS-SGM}~\cite{qcs_sgm} formulates quantized recovery as posterior sampling using a learned score prior, while QCS-SGM+~\cite{meng2024qcs+} extends this framework beyond row-orthogonal sensing matrices for broader applicability.

\begin{figure}
	[t]
	\centering
	\includegraphics[width=0.7\columnwidth]{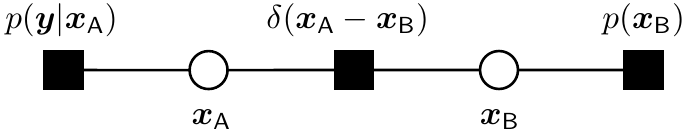}
	\caption{Factor graph for \ac{TMP} derivation.
		The circles represent variable nodes and the squares represent factor nodes.}
	\label{factor_graph}
\end{figure}

\section{TMP Framework}\label{sec:tmp_framework}
In this section, we briefly review the \ac{TMP} framework~\cite{turbo_cs} for solving linear inverse problems.
Consider the real-valued linear model
\begin{align} \label{linear_inverse_problem}
	\bsm{y} = \bsm{A}\bsm{x} + \bsm{n},
\end{align}
where $\bsm{x} \in \mathbb{R}^{N}$ is an unknown signal vector, $\bsm{y} \in \mathbb{R}^{M}$ denotes noisy linear measurements, $\bsm{A} \in \mathbb{R}^{M \times N}$ is a known measurement matrix, and $\bsm{n} \sim \mathcal{N}(\bsm{0}, \delta_0^2 \bsm{I})$ is an \ac{AWGN}.
Our goal is to recover $\bsm{x}$ from $\bsm{y}$, particularly in the underdetermined regime where $M < N$.

{\color{black}
Message passing is a Bayesian inference framework for estimating unknown variables by combining measurement data with prior knowledge.
According to Bayes' rule, the posterior distribution can be factorized as
\begin{align}\label{posterior_distribution}
	p(\bsm{x}|\bsm{y}) \propto p (\bsm{x}) p(\bsm{y}|\bsm{x}).
\end{align}
In this framework, the posterior is represented by a factor graph and approximated via iterative local message exchanges, where each message is a function (or its parametric approximation) passed between nodes that summarizes the contribution of a subset of factors to a variable’s marginal distribution.

\ac{TMP} approximates the exchanged messages as Gaussian distributions characterized by their means and variances, enabling a tractable and efficient evaluation of the posterior mean $\mathbb{E}[\bsm{x}|\bsm{y}]$.
Specifically, \ac{TMP} operates on the vector factor graph shown in Fig.~\ref{factor_graph}, which is constructed by introducing an auxiliary variable splitting $\bsm{x}_{\sf A} = \bsm{x}_{\sf B} = \bsm{x}$.
Under this reformulation, the posterior distribution admits the equivalent factorization
\begin{align}
	p(\bsm{x}_{\sf A}, \bsm{x}_{\sf B} |\bsm{y}) \propto p(\bsm{x}_{\sf B}) \delta (\bsm{x}_{\sf A} - \bsm{x}_{\sf B}) p(\bsm{y}|\bsm{x}_{\sf A}),
\end{align}
where $\delta(\cdot)$ is the Dirac delta function.
Unlike scalar factor graphs commonly used in conventional message passing, which typically rely on component-wise prior factorization, the vector formulation in Fig.~\ref{factor_graph} preserves the image as a whole at the prior node.
This enables the incorporation of expressive high-dimensional priors that capture strong spatial dependencies across pixels.

At a high level, \ac{TMP} performs iterative inference on the factor graph by exchanging Gaussian-approximated messages between the likelihood node and the prior node.
This leads to a modular interpretation of~\ac{TMP} that resembles the turbo decoding principle~\cite{turbo_codes}, which consists of two modules.
Module A performs \ac{LMMSE} estimation of $\bsm{x}$ from $\bsm{y}$, using the messages from module B as the effective prior mean and variance.
Module B acts as an \ac{MMSE} denoiser that incorporates the prior information of $\bsm{x}$ together with the messages from module A.
The two modules are executed iteratively to refine the estimate of $\bsm{x}$.
In the following, we provide a concise derivation of \ac{TMP} based on the \ac{EP} principle~\cite{EP2001Minka}.
}

\subsection{Module A: \ac{LMMSE} Estimator}
We start by initializing the message from the factor node $\delta (\bsm{x}_{\sf A} - \bsm{x}_{\sf B})$ to the variable node $\bsm{x}_{\sf A}$ as $\mu_{\delta \rightarrow \sf{A}} (\bsm{x}_{\sf A}) = \mathcal{N}\big(\bsm{x}_{\sf A}; \bsm{x}_{\sf A}^{\mathtt{pri}}, v_{\sf A}^{\mathtt{pri}}\bsm{I}\big)$.
According to \ac{EP}, the belief on $\bsm{x}_{\sf A}$ 
is given by $b(\bsm{x}_{\sf A}) = \mathcal{N}\big(\bsm{x}_{\sf A}; \bsm{x}_{\sf A}^{\mathtt{post}}, \bsm{V}_{\sf A}^{\mathtt{post}}\big) =  \mathrm{proj}_G\left[p(\bsm{y}|\bsm{x}_{\sf A}) \mu_{\delta \rightarrow \sf{A}} (\bsm{x}_{\sf A})\right]$,
where $\mathrm{proj}_G[\cdot]$ represents the projection of a probability distribution to a Gaussian density with matched first- and second-order moments.
By noting that the product of two Gaussian messages is proportional to a Gaussian \ac{PDF},
the projection in $b(\bsm{x}_{\sf A})$ can thus be removed.
\ac{TMP} further approximates $\bsm{V}_{\sf A}^\mathtt{post}$ using a shared scalar variance \cite{turbo_cs}, resulting in the belief of the form $b(\bsm{x}_{\sf A}) \approx \mathcal{N}\big(\bsm{x}_{\sf A}; \bsm{x}_{\sf A}^{\mathtt{post}}, v_{\sf A}^{\mathtt{post}}\bsm{I}\big)$, where 
\begin{align}
	\!\!\bsm{x}_{\sf A}^{\mathtt{post}} \!&= \bsm{x}_{\sf A}^\mathtt{pri} + v_{\sf A}^\mathtt{pri}\bsm{A}^\top \big(v_{\sf A}^\mathtt{pri}\bsm{A}\bsm{A}^\top + \delta_0^2\bsm{I}\big)^{-1} \big(\bsm{y} - \bsm{A} \bsm{x}_{\sf A}^\mathtt{pri}\big), \label{lmmse_mean}\\
	\!\!v_{\sf A}^\mathtt{post} \!&= v_{\sf A}^\mathtt{pri} - \frac{(v_{\sf A}^\mathtt{pri})^2}{N} \mathrm{tr} \left(\bsm{A}^\top \big(v_{\sf A}^\mathtt{pri}\bsm{A} \bsm{A}^\top + \delta_0^2 \bsm{I}\big)^{-1}\bsm{A}\right). \label{lmmse_variance}
\end{align}
Eqns.~\eqref{lmmse_mean} and \eqref{lmmse_variance} are the posterior mean and variance from the \ac{LMMSE} estimator of $\bsm{x}_{\sf A}$, respectively.
{\color{black}
	While per-element or full covariance estimation could capture heterogeneous uncertainty across image pixels, such estimation is computationally prohibitive and statistically unreliable, especially in highly compressed settings.
	In contrast, scalar averaging is a standard design choice to stabilize message-passing updates~\cite{turbo_cs, vamp}.}

{\color{black}
	The following remarks discuss efficient matrix-inverse-free implementations of the \ac{LMMSE} update for different classes of measurement matrices.
	\begin{remk}[Partial orthogonal $\bsm{A}$]
	 The measurement matrix $\bsm{A}$ in compressive imaging is typically partial orthogonal (i.e., $\bsm{A}\bsm{A}^\top = \bsm{I}$), constructed by randomly selecting rows from an orthogonal transform such as the \ac{DCT}, \ac{DFT}, or Hadamard matrix.
	The partial orthogonality of $\bsm{A}$ enables the following simplified \ac{LMMSE} updates that avoid explicit matrix inversion:
	\begin{align}
		\bsm{x}_{\sf A}^{\mathtt{post}} &=  \bsm{x}_{\sf A}^\mathtt{pri} + \frac{v_{\sf A}^\mathtt{pri}}{v_{\sf A}^\mathtt{pri} + \delta_0^2}\bsm{A}^\top \big(\bsm{y} - \bsm{A} \bsm{x}_{\sf A}^\mathtt{pri}\big), \\
		v_{\sf A}^\mathtt{post} &= v_{\sf A}^\mathtt{pri} - \frac{M}{N} \frac{(v_{\sf A}^\mathtt{pri})^2}{v_{\sf A}^\mathtt{pri} + \delta_0^2}.
	\end{align}
\end{remk}

\begin{remk}[\Ac{SVD} of $\bsm{A}$ available]
	Even when $\bsm{A}$ is not partial orthogonal, the \ac{LMMSE} update in module A can still be implemented efficiently without explicit matrix inversion.
	Let $\bsm{A} \triangleq \bsm{U}\bsm{\Sigma}\bsm{V}^\top$, where $\bsm{U} \in \mathbb{R}^{M \times M}$ and $\bsm{V} \in \mathbb{R}^{N \times N}$ are orthogonal matrices, and $\bsm{\Sigma} \in \mathbb{R}^{M \times N}$ has singular values $\{\lambda_i\}_{i=1}^M$ on its diagonal.
	Then, 
	\begin{align}
		&\big(v_{\sf A}^\mathtt{pri}\bsm{A}\bsm{A}^\top + \delta_0^2\bsm{I}\big)^{-1} \nonumber \\
		= &\, \bsm{U} \diag \left\{\frac{1}{v_{\sf A}^\mathtt{pri}\lambda_1^2 + \delta_0^2}, \dots, \frac{1}{v_{\sf A}^\mathtt{pri}\lambda_M^2 + \delta_0^2}\right\} \bsm{U}^\top.
	\end{align}
	Thus, the inverse reduces to two orthogonal transforms and a diagonal scaling.
	Moreover, the trace term in~\eqref{lmmse_variance} simplifies to
	\begin{align}
		\mathrm{tr} \left(\bsm{A}^\top \big(v_{\sf A}^\mathtt{pri}\bsm{A} \bsm{A}^\top + \delta_0^2 \bsm{I}\big)^{-1}\bsm{A}\right) 
		= \sum_{i=1}^M \frac{\lambda_i^2}{v_{\sf A}^\mathtt{pri}\lambda_i^2 + \delta_0^2}.
	\end{align}
\end{remk}

\begin{remk}[\Ac{SVD} of $\bsm{A}$ not available]
	When the \ac{SVD} of $\bsm{A}$ is not available or is too costly to compute, the \ac{LMMSE} update can be iteratively calculated by, e.g., the \ac{CG} method.
	Specifically, define $\bsm{H} \triangleq v_{\sf A}^\mathtt{pri}\bsm{A} \bsm{A}^\top + \delta_0^2 \bsm{I}$ and
	$\bsm{r} \triangleq \bsm{y} - \bsm{A} \bsm{x}_{\sf A}^\mathtt{pri}$.
	Then, \eqref{lmmse_mean} can be rewritten as $\bsm{x}_{\sf A}^{\mathtt{post}} = \bsm{x}_{\sf A}^\mathtt{pri} + v_{\sf A}^\mathtt{pri}\bsm{A}^\top \bsm{u}$, where $\bsm{u}$ is the solution to the linear system $\bsm{H} \bsm{u} = \bsm{r}$.
	Since $\bsm{H}$ is symmetric positive definite for $\delta_0 > 0$, the vector $\bsm{u}$ can be efficiently obtained using the \ac{CG} method.
	For the variance update in~\eqref{lmmse_variance}, the trace term can be efficiently estimated using the Hutchinson trace estimator with random probe vectors.
\end{remk}
}
The message passed from the variable node $\bsm{x}_{\sf A}$ to the factor node $\delta (\bsm{x}_{\sf A} - \bsm{x}_{\sf B})$, referred to as the extrinsic message, is given by
$\mu_{\sf{A} \rightarrow \delta} (\bsm{x}_{\sf A}) = \mathcal{N}\big(\bsm{x}_{\sf A}; \bsm{x}_{\sf A}^{\mathtt{ext}}, v_{\sf A}^{\mathtt{ext}}\bsm{I}\big) = \mathcal{N}\big(\bsm{x}_{\sf A}; \bsm{x}_{\sf A}^{\mathtt{post}}, v_{\sf A}^{\mathtt{post}}\bsm{I}\big)/\mathcal{N}\big(\bsm{x}_{\sf A}; \bsm{x}_{\sf A}^{\mathtt{pri}}, v_{\sf A}^{\mathtt{pri}}\bsm{I}\big)$.
This yields the update rules in Lines~\ref{ext_v_A} and~\ref{ext_x_A} of Algorithm~\ref{alg:turbo_cs}.

\begin{algorithm}[t]
	\caption{\ac{TMP} Framework}
	\label{alg:turbo_cs}
	\begin{algorithmic}[1]
		\STATE {\bfseries Input:} $\bsm{A}$, $\bsm{y}$, $\delta_0^2$, $\bsm{x}_{\sf A}^\mathtt{pri}$, $v_{\sf A}^\mathtt{pri}$
		\STATE {\bfseries Output:} $\bsm{x}_{\sf B}^\mathtt{post}$
		\REPEAT
		\STATE \% LMMSE estimator (module A)  \color{black}
		\STATE $\bsm{x}_{\sf A}^{\mathtt{post}} = \bsm{x}_{\sf A}^\mathtt{pri} + v_{\sf A}^\mathtt{pri}\bsm{A}^\top \big(v_{\sf A}^\mathtt{pri}\bsm{A}\bsm{A}^\top + \delta_0^2\bsm{I}\big)^{-1} \big(\bsm{y} - \bsm{A} \bsm{x}_{\sf A}^\mathtt{pri}\big)$ \label{post_x_A}
		\STATE $v_{\sf A}^\mathtt{post} = v_{\sf A}^\mathtt{pri} - \frac{(v_{\sf A}^\mathtt{pri})^2}{N} \mathrm{tr} \big(\bsm{A}^\top \big(v_{\sf A}^\mathtt{pri}\bsm{A} \bsm{A}^\top + \delta_0^2 \bsm{I}\big)^{-1}\bsm{A}\big)$ \label{post_v_A} \color{black}
		\STATE $v_{\sf B}^{\mathtt{pri}} = v_{\sf A}^\mathtt{ext} = \left(\frac{1}{v_{\sf A}^{\mathtt{post}}} - \frac{1}{v_{\sf A}^{\mathtt{pri}}}\right)^{-1}$ \label{ext_v_A}
		\STATE $\bsm{x}_{\sf B}^{\mathtt{pri}} = \bsm{x}_{\sf A}^\mathtt{ext} = v_{\sf A}^{\mathtt{ext}} \left(\frac{\bsm{x}_{\sf A}^{\mathtt{post}}}{v_{\sf A}^{\mathtt{post}}} - \frac{\bsm{x}_{\sf A}^{\mathtt{pri}}}{v_{\sf A}^{\mathtt{pri}}}\right)$ \label{ext_x_A}
		\STATE \% MMSE denoiser (module B)
		\STATE $\bsm{x}_{\sf B}^{\mathtt{post}} = \mathbb{E} \big[\bsm{x}_{\sf B}\big|\bsm{x}_{\sf B}^{\mathtt{pri}}\big]$ 
		\STATE $v_{\sf B}^{\mathtt{post}} = \frac{1}{N} \mathrm{tr} \big(\mathrm{Cov} \big[\bsm{x}_{\sf B}\big|\bsm{x}_{\sf B}^{\mathtt{pri}}\big] \big)$ \label{post_v_B}
		\STATE $ v_{\sf A}^{\mathtt{pri}} = v_{\sf B}^\mathtt{ext} = \left(\frac{1}{v_{\sf B}^{\mathtt{post}}} - \frac{1}{v_{\sf B}^{\mathtt{pri}}}\right)^{-1}$  \label{ext_v_B}
		\STATE $ \bsm{x}_{\sf A}^{\mathtt{pri}} = \bsm{x}_{\sf B}^\mathtt{ext} = v_{\sf B}^{\mathtt{ext}} \left(\frac{\bsm{x}_{\sf B}^{\mathtt{post}}}{v_{\sf B}^{\mathtt{post}}} - \frac{\bsm{x}_{\sf B}^{\mathtt{pri}}}{v_{\sf B}^{\mathtt{pri}}}\right)$ \label{ext_x_B}
		\UNTIL{the stopping criterion is met}
	\end{algorithmic}
\end{algorithm}

\subsection{Module B: \ac{MMSE} denoiser}
The message $\mu_{\sf{A} \rightarrow \delta} (\bsm{x}_{\sf A}) = \mathcal{N}\big(\bsm{x}_{\sf A}; \bsm{x}_{\sf A}^{\mathtt{ext}}, v_{\sf A}^{\mathtt{ext}}\bsm{I}\big)$ flows rightward through the factor node $\delta (\bsm{x}_{\sf A} - \bsm{x}_{\sf B})$ unchanged, manifesting as $\mu_{\delta \rightarrow \sf{B}} (\bsm{x}_{\sf B}) = \mathcal{N}\big(\bsm{x}_{\sf B}; \bsm{x}_{\sf B}^{\mathtt{pri}}, v_{\sf B}^{\mathtt{pri}}\bsm{I}\big)$ on the other side.
The belief on $\bsm{x}_{\sf B}$ is calculated as $b(\bsm{x}_{\sf B}) = \mathcal{N}\big(\bsm{x}_{\sf B}; \bsm{x}_{\sf B}^{\mathtt{post}}, \bsm{V}_{\sf B}^{\mathtt{post}}\big) = \mathrm{proj}_G\left[p(\bsm{x}_{\sf B}) \mu_{\delta \rightarrow \sf{B}} (\bsm{x}_{\sf B})\right]$, where
\begin{align}
	\bsm{x}_{\sf B}^{\mathtt{post}} &= \frac{1}{Z}\int \bsm{x}_{\sf B} p(\bsm{x}_{\sf B})  \mathcal{N}\big(\bsm{x}_{\sf B}; \bsm{x}_{\sf B}^{\mathtt{pri}}, v_{\sf B}^{\mathtt{pri}}\bsm{I}\big) \mathrm{d} \bsm{x}_{\sf B}, \label{moduleB_integral_mean}\\
	\bsm{V}_{\sf B}^{\mathtt{post}} &= \frac{1}{Z} \int \big(\bsm{x}_{\sf B} - \bsm{x}_{\sf B}^{\mathtt{post}}\big) \big(\bsm{x}_{\sf B} - \bsm{x}_{\sf B}^{\mathtt{post}}\big)^\top \nonumber \\
	&~~~~~~~~~~~~~~~~~\times p(\bsm{x}_{\sf B})  \mathcal{N}\big(\bsm{x}_{\sf B}; \bsm{x}_{\sf B}^{\mathtt{pri}}, v_{\sf B}^{\mathtt{pri}}\bsm{I}\big) \mathrm{d} \bsm{x}_{\sf B} , \label{moduleB_integral_variance}
\end{align}
with $Z = \int p(\bsm{x}_{\sf B})  \mathcal{N}\big(\bsm{x}_{\sf B}; \bsm{x}_{\sf B}^{\mathtt{pri}}, v_{\sf B}^{\mathtt{pri}}\bsm{I}\big) \mathrm{d} \bsm{x}_{\sf B}$ being the normalization constant.
\ac{TMP} models $\bsm{x}_{\sf B}^\mathtt{pri}$ as an \ac{AWGN} observation of the ground-truth signal $\bsm{x}_{\sf B}$: 
\begin{align}
	\bsm{x}_{\sf B}^\mathtt{pri} = \bsm{x}_{\sf B} + \bsm{w}, \quad \bsm{w} \sim \mathcal{N}\big(\bsm{0}, v_{\sf B}^\mathtt{pri}\bsm{I}\big). \label{AWGN_observation}
\end{align}
From \eqref{AWGN_observation}, we have $p\big(\bsm{x}_{\sf B}^\mathtt{pri}|\bsm{x}_{\sf B}\big) = \mathcal{N}\big(\bsm{x}_{\sf B}^{\mathtt{pri}}; \bsm{x}_{\sf B}, v_{\sf B}^{\mathtt{pri}}\bsm{I}\big)$.
By noting that $\mathcal{N}\big(\bsm{x}_{\sf B}; \bsm{x}_{\sf B}^{\mathtt{pri}}, v_{\sf B}^{\mathtt{pri}}\bsm{I}\big)$ in \eqref{moduleB_integral_mean} and \eqref{moduleB_integral_variance} takes the same form as $p\big(\bsm{x}_{\sf B}^\mathtt{pri}|\bsm{x}_{\sf B}\big)$,
the belief update on $\bsm{x}_{\sf B}$ can be interpreted as an \ac{MMSE} denoising problem under the AWGN observation model \eqref{AWGN_observation}, i.e.,
\begin{align}
	\bsm{x}_{\sf B}^{\mathtt{post}} &= \mathbb{E} \big[\bsm{x}_{\sf B}\big|\bsm{x}_{\sf B}^{\mathtt{pri}}\big], \label{post_mean}\\
	\bsm{V}_{\sf B}^{\mathtt{post}} &= \mathrm{Cov} \big[\bsm{x}_{\sf B}\big|\bsm{x}_{\sf B}^{\mathtt{pri}}\big]. \label{post_variance}
\end{align}
Similarly, we approximate $\bsm{V}_{\sf B}^{\mathtt{post}}$ by $v_{\sf B}^\mathtt{post}\bsm{I}$ with $v_{\sf B}^\mathtt{post} = \mathrm{tr} \big(\mathrm{Cov} \big[\bsm{x}_{\sf B}\big|\bsm{x}_{\sf B}^{\mathtt{pri}}\big] \big) /N$ to simplify the computation.
The extrinsic message of $\bsm{x}_{\sf B}$ is denoted as $\mu_{{\sf B} \rightarrow \delta}(\bsm{x}_{\sf B}) = \mathcal{N}\big(\bsm{x}_{\sf B}; \bsm{x}_{\sf B}^{\mathtt{ext}}, v_{\sf B}^{\mathtt{ext}}\bsm{I}\big)$, where the variance and mean are respectively given in Lines~\ref{ext_v_B} and~\ref{ext_x_B} of Algorithm~\ref{alg:turbo_cs}.
Finally, the extrinsic message of $\bsm{x}_{\sf B}$ flows leftward through the factor node $\delta(\bsm{x}_{\sf A}-\bsm{x}_{\sf B})$ unchanged, serving as the prior message of $\bsm{x}_{\sf A}$ in the next iteration, i.e., $\bsm{x}_{\sf A}^\mathtt{pri} = \bsm{x}_{\sf B}^\mathtt{ext}, v_{\sf A}^\mathtt{pri} = v_{\sf B}^\mathtt{ext}$.

The main challenge in implementing the \ac{TMP} framework arises from the \ac{MMSE} denoising part in module B, primarily due to i) the lack of a tractable prior image distribution, and ii) the high-dimensional integrals involved in calculating the posterior mean and variance. 
To sidestep these difficulties, various off-the-shelf image denoisers (e.g., TV, SURE-LET \cite{sure-let}, BM3D \cite{bm3d}) are plugged into message-passing algorithms in place of the \ac{MMSE} denoiser.
These \ac{PnP} denoisers typically rely on generic or hand-crafted priors, which could lead to substantial imperfection in compressive imaging applications due to the oversimplification or misspecification of the prior distribution.
Motivated by the above, we propose the integration of score-based generative models into \ac{TMP} as accurate and expressive image priors, leading to a data-driven Bayesian inference approach to compressive imaging.

\section{Learning Score-Based Generative Priors for \ac{MMSE} Denoising} \label{sec:learning_score_priors}
\subsection{Bridging Score Function and \ac{MMSE} Denoising}
\begin{theo}[Tweedie's formula~\cite{robbins1992empirical, efron2011tweedie} and its second-order generalization~\cite{high_order_gradients}] \label{tweedie_mean_covariance}
	Consider $\tilde{\bsm{x}}$ to be an \ac{AWGN} observation of $\bsm{x}$, i.e., $\tilde{\bsm{x}} = \bsm{x} + \bsm{w}$, where the prior distribution $p(\bsm{x})$ and the noise distribution $\bsm{w} \sim \mathcal{N}(\bsm{0}, \sigma^2 \bsm{I})$ are given.
	The posterior mean and covariance are respectively given by
	\begin{align}
		\mathbb{E} \left[\bsm{x}|\tilde{\bsm{x}}\right] &= \tilde{\bsm{x}} + \sigma^2 \nabla_{\tilde{\bsm{x}}} \log p(\tilde{\bsm{x}}), \label{1st_order_tweedie} \\
		\mathrm{Cov} \left[\bsm{x}|\tilde{\bsm{x}}\right] &= \sigma^2 \bsm{I} + \sigma^4 \nabla_{\tilde{\bsm{x}}}^2 \log p(\tilde{\bsm{x}}). \label{2nd_order_tweedie}
	\end{align}
\end{theo}
Theorem~\ref{tweedie_mean_covariance} bridges score-based generative modeling and empirical Bayes method, providing the theoretical foundation for using score functions to perform \ac{MMSE} denoising of \ac{AWGN} observations.
With the score functions available, Theorem~\ref{tweedie_mean_covariance} enables the efficient implementation of the \ac{MMSE} denoiser in \eqref{post_mean} and \eqref{post_variance}.
We next discuss how to obtain the first- and second-order score estimates, $\nabla_{\tilde{\bsm{x}}} \log p(\tilde{\bsm{x}})$ and $\nabla_{\tilde{\bsm{x}}}^2 \log p(\tilde{\bsm{x}})$, via the denoising score matching technique~\cite{vincent2011connection, song2019generative}.

\subsection{First-Order Score Matching}\label{1st_sm}
For a given $\sigma$, we aim to learn a first-order score model $\bsm{s}_{\bsm{\theta}}(\cdot, \sigma^2): \mathbb{R}^N \rightarrow \mathbb{R}^N$ parameterized by $\bsm{\theta}$ with the following score matching objective:
\begin{align}
	\min_{\bsm{\theta}} ~\mathbb{E}_{p(\tilde{\bsm{x}})}\left[ \left\|\bsm{s}_{\bsm{\theta}} (\tilde{\bsm{x}}, \sigma^2) - \nabla_{\tilde{\bsm{x}}} \log p(\tilde{\bsm{x}}) \right\|^2_2 \right].
\end{align}
The above formulation is shown to be equivalent to the denoising score matching objective \cite{vincent2011connection, song2019generative}, given by
\begin{align} \label{1st_dsm}
	\min_{\bsm{\theta}}~\mathbb{E}_{p(\bsm{x})p(\tilde{\bsm{x}}|\bsm{x})} \left[ \left\|\bsm{s}_{\bsm{\theta}} (\tilde{\bsm{x}}, \sigma^2) - \nabla_{\tilde{\bsm{x}}} \log p(\tilde{\bsm{x}}|\bsm{x}) \right\|^2_2 \right].
\end{align}
Substituting $\nabla_{\tilde{\bsm{x}}} \log p(\tilde{\bsm{x}}|\bsm{x}) = -\frac{\tilde{\bsm{x}} - \bsm{x}}{\sigma^2}$ into \eqref{1st_dsm} yields
\begin{align} \label{1st_dsm_2}
	\min_{\bsm{\theta}}~ \ell_1(\bsm{\theta}; \sigma) \triangleq \mathbb{E}_{p(\bsm{x})p(\tilde{\bsm{x}}|\bsm{x})} \left[ \left\|\bsm{s}_{\bsm{\theta}} (\tilde{\bsm{x}}, \sigma^2) + \frac{\tilde{\bsm{x}} - \bsm{x}}{\sigma^2} \right\|^2_2 \right].
\end{align}
In practice, the score-based \ac{MMSE} denoiser is required to operate effectively across a range of noise levels.
We therefore combine \eqref{1st_dsm_2} for all possible $\sigma \in \{\sigma_i\}_{i=1}^L$ to formulate the following unified objective for training the first-order score model:
\begin{align}
	\min_{\bsm{\theta}} ~\mathcal{L}_1 \big(\bsm{\theta}; \{\sigma_i\}_{i=1}^L\big) \triangleq \frac{1}{L} \sum_{i=1}^{L} \lambda_1(\sigma_i)  \ell_1(\bsm{\theta}; \sigma_i),
\end{align}
where $\lambda_1(\sigma_i)$ is the weighting factor depending on $\sigma_i$.

\subsection{Second-Order Score Matching}
For a given $\sigma$, we aim to learn a second-order score model $\bsm{S}_{\bsm{\phi}}(\cdot, \sigma^2): \mathbb{R}^N \rightarrow \mathbb{R}^{N \times N}$ parameterized by $\bsm{\phi}$ with the following score matching objective:
\begin{align} \label{2nd_sm}
	\min_{\bsm{\phi}}~\mathbb{E}_{p(\tilde{\bsm{x}})} \left[ \left\|\bsm{S}_{\bsm{\phi}}(\tilde{\bsm{x}}, \sigma^2) - \nabla_{\tilde{\bsm{x}}}^2 \log p(\tilde{\bsm{x}}) \right\|^2_F \right].
\end{align}
The second-order score matching objective also has an equivalent denoising score matching formulation \cite{lu2022maximum}:
\begin{align} 
	\min_{\bsm{\phi}}~&\mathbb{E}_{p(\bsm{x})p(\tilde{\bsm{x}}|\bsm{x})} \bigg[ \Big\|\bsm{S}_{\bsm{\phi}}(\tilde{\bsm{x}}, \sigma^2) \nonumber \\
	&~~~~~~~~~~~~~- \bsm{b}(\bsm{x}, \tilde{\bsm{x}}, \sigma^2) \bsm{b}(\bsm{x},\tilde{\bsm{x}}, \sigma^2)^\top +  \frac{\bsm{I}}{\sigma^2}\Big\|_F^2\bigg], \label{2nd_dsm}
\end{align}
where $\bsm{b}(\bsm{x}, \tilde{\bsm{x}}, \sigma^2) \triangleq \nabla_{\tilde{\bsm{x}}} \log p(\tilde{\bsm{x}}) + \frac{\tilde{\bsm{x}} - \bsm{x}}{\sigma^2} $.
The objective in \eqref{2nd_dsm} relies on the ground-truth of the first-order score function, which is in general not accessible.
In practice, we replace $\nabla_{\tilde{\bsm{x}}} \log p(\tilde{\bsm{x}})$ by the learned $\bsm{s}_{\bsm{\theta}}(\tilde{\bsm{x}}, \sigma^2)$ for efficient computation.
The objective in \eqref{2nd_dsm} then becomes
\begin{align} 
	\min_{\bsm{\phi}}~&\mathbb{E}_{p(\bsm{x})p(\tilde{\bsm{x}}|\bsm{x})} \bigg[ \Big\|\bsm{S}_{\bsm{\phi}}(\tilde{\bsm{x}}, \sigma^2) \nonumber \\
	&~~~~~~~~~~~~~- \hat{\bsm{b}}(\bsm{x}, \tilde{\bsm{x}}, \sigma^2) \hat{\bsm{b}}(\bsm{x}, \tilde{\bsm{x}}, \sigma^2)^\top +  \frac{\bsm{I}}{\sigma^2}\Big\|_F^2\bigg], \label{2nd_dsm_estimated}
\end{align}
where $\hat{\bsm{b}}(\bsm{x}, \tilde{\bsm{x}}, \sigma^2) \triangleq \bsm{s}_{\bsm{\theta}}(\tilde{\bsm{x}}, \sigma^2) + \frac{\tilde{\bsm{x}} - \bsm{x}}{\sigma^2} $.
It is proved in \cite{lu2022maximum} that with this replacement, the second-order score model has the error-bounded property if the first-order score matching error is bounded.

As in Line~\ref{post_v_B} of Algorithm~\ref{alg:turbo_cs}, only the trace of the posterior covariance matrix is needed, rather than the covariance matrix itself.
By noting in \eqref{2nd_order_tweedie} the simple relation between the posterior covariance matrix and the second-order score function, 
it suffices to only match the trace of the second-order score function for posterior variance evaluation.
The simplified objective is expressed as
\begin{align} \label{2nd_dsm_trace}
	\min_{\bsm{\phi}}~ \ell_2(\bsm{\phi}; \sigma) &\triangleq \mathbb{E}_{p(\bsm{x})p(\tilde{\bsm{x}}|\bsm{x})} \bigg[ \bigg|\mathrm{tr}\big(\bsm{S}_{\bsm{\phi}}(\tilde{\bsm{x}}, \sigma^2)\big) \nonumber \\
	&~~~~~~~~~~~~~~- \left\|\hat{\bsm{b}}(\bsm{x}, \tilde{\bsm{x}}, \sigma^2)\right\|_2^2 +  \frac{N}{\sigma^2}\bigg|^2\bigg].
\end{align}
We combine \eqref{2nd_dsm_trace} for all $\sigma \in \{\sigma_i\}_{i=1}^L$ to obtain the following unified objective for training the second-order score model:
\begin{align} \label{2nd_dsm_unified}
	\min_{\bsm{\phi}}~\mathcal{L}_2 \big(\bsm{\phi}; \{\sigma_i\}_{i=1}^L\big) \triangleq \frac{1}{L} \sum_{i=1}^{L} \lambda_2(\sigma_i)  \ell_2(\bsm{\phi}; \sigma_i),
\end{align}
where $\lambda_2(\sigma_i)$ is the weighting factor depending on $\sigma_i$.
In practice, one can first train a first-order score model, and then freeze it when applied to second-order score learning.
Alternatively, the two score models can be trained jointly by disabling gradient backpropagation from the second-order training objective \eqref{2nd_dsm_unified} to the first-order score model.

\section{STMP Algorithm}\label{sec:stmp_algorithm}
\subsection{Overall Algorithm}
\begin{figure}
	[t]
	\centering
	\includegraphics[width=1\columnwidth]{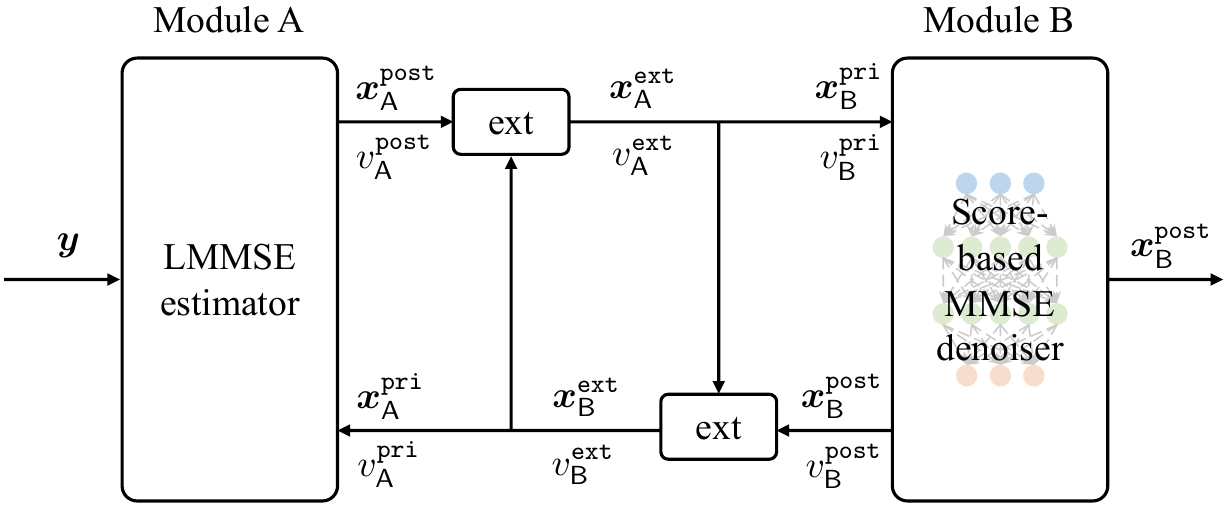}
	\caption{Diagram of the \ac{STMP} algorithm.}
	\label{fig:stmp}
\end{figure}

The learned score models can be plugged into the \ac{TMP} framework, facilitating \ac{MMSE} denoising via
\begin{align}
	\bsm{x}_{\sf B}^\mathtt{post} &= \bsm{x}_{\sf B}^\mathtt{pri} + v_{\sf B}^\mathtt{pri} \bsm{s}_{\bsm{\theta}} \big(\bsm{x}_{\sf B}^\mathtt{pri}, v_{\sf B}^\mathtt{pri}\big), \label{posterior_mean_stmp} \\
	v_{\sf B}^\mathtt{post} &= v_{\sf B}^\mathtt{pri} + \frac{(v_{\sf B}^\mathtt{pri})^2}{N} \mathrm{tr} \Big(\bsm{S}_{\bsm{\phi}} \big(\bsm{x}_{\sf B}^\mathtt{pri}, v_{\sf B}^\mathtt{pri}\big) \Big). \label{posterior_variance_stmp}
\end{align}
This score-based implementation yields the \ac{STMP} algorithm summarized in Algorithm~\ref{alg:score_turbo_mp}.
An illustrative diagram is shown in Fig.~\ref{fig:stmp}.
The iteration proceeds until either a preset maximum number of iterations is reached or the relative change in $\bsm{x}_{\sf B}^\mathtt{post}$ across iterations falls below a predefined threshold.
We have empirically observed that when the sampling ratio $M/N$ is low, adding damping to the message updates helps to stabilize convergence.
In particular, we suggest replacing Line \ref{to_be_damped_A} of Algorithm \ref{alg:score_turbo_mp} by
\begin{align}
	\bsm{x}_{\sf B}^\mathtt{pri} &= \beta \bsm{x}_{\sf A}^\mathtt{ext} + (1-\beta) \bsm{x}_{{\sf A}, \mathtt{old}}^\mathtt{ext}, \\
	v_{\sf B}^\mathtt{pri} &= \beta v_{\sf A}^\mathtt{ext} + (1-\beta) v_{{\sf A},\mathtt{old}}^\mathtt{ext},
\end{align}
where $\bsm{x}_{{\sf A}, \mathtt{old}}^\mathtt{ext}$ and $v_{{\sf A},\mathtt{old}}^\mathtt{ext}$ are the extrinsic mean and variance of module A from the last iteration, and $\beta \in (0, 1]$ denotes the damping factor.
Line \ref{to_be_damped_B} of Algorithm \ref{alg:score_turbo_mp} should be modified similarly.

\begin{algorithm}[t]
	\caption{\ac{STMP} Algorithm}
	\label{alg:score_turbo_mp}
	\begin{algorithmic}[1]
		\STATE {\bfseries Input:} $\bsm{A}$, $\bsm{y}$, $\delta_0^2$, $\bsm{x}_{\sf A}^\mathtt{pri}$, $v_{\sf A}^\mathtt{pri}$
		\STATE {\bfseries Output:} $\bsm{x}_{\sf B}^\mathtt{post}$
		\REPEAT
		\STATE \% LMMSE estimator (module A)
		\color{black}
		\STATE $\bsm{x}_{\sf A}^{\mathtt{post}} = \bsm{x}_{\sf A}^\mathtt{pri} + v_{\sf A}^\mathtt{pri}\bsm{A}^\top \big(v_{\sf A}^\mathtt{pri}\bsm{A}\bsm{A}^\top + \delta_0^2\bsm{I}\big)^{-1} \big(\bsm{y} - \bsm{A} \bsm{x}_{\sf A}^\mathtt{pri}\big)$ \label{stmp_module_a_mean}
		\STATE $v_{\sf A}^\mathtt{post} = v_{\sf A}^\mathtt{pri} - \frac{(v_{\sf A}^\mathtt{pri})^2}{N} \mathrm{tr} \big(\bsm{A}^\top \big(v_{\sf A}^\mathtt{pri}\bsm{A} \bsm{A}^\top + \delta_0^2 \bsm{I}\big)^{-1}\bsm{A}\big)$ \label{stmp_v_A_post}
		\color{black}
		\STATE $v_{\sf A}^\mathtt{ext} = \left(\frac{1}{v_{\sf A}^{\mathtt{post}}} - \frac{1}{v_{\sf A}^{\mathtt{pri}}}\right)^{-1}$ \label{stmp_v_A_ext}
		\STATE $\bsm{x}_{\sf A}^\mathtt{ext} = v_{\sf A}^{\mathtt{ext}} \left(\frac{\bsm{x}_{\sf A}^{\mathtt{post}}}{v_{\sf A}^{\mathtt{post}}} - \frac{\bsm{x}_{\sf A}^{\mathtt{pri}}}{v_{\sf A}^{\mathtt{pri}}}\right)$ 
		\STATE $\bsm{x}_{\sf B}^\mathtt{pri} = \bsm{x}_{\sf A}^\mathtt{ext}, v_{\sf B}^\mathtt{pri} = v_{\sf A}^\mathtt{ext}$ \label{to_be_damped_A}
		\STATE \% Score-based MMSE denoiser (module B)
		\STATE $\bsm{x}_{\sf B}^\mathtt{post} = \bsm{x}_{\sf B}^\mathtt{pri} + v_{\sf B}^\mathtt{pri} \bsm{s}_{\bsm{\theta}} \big(\bsm{x}_{\sf B}^\mathtt{pri}, v_{\sf B}^\mathtt{pri}\big)$ 
		\STATE $v_{\sf B}^\mathtt{post} = v_{\sf B}^\mathtt{pri} + \frac{(v_{\sf B}^\mathtt{pri})^2}{N} \mathrm{tr} \Big(\bsm{S}_{\bsm{\phi}} \big(\bsm{x}_{\sf B}^\mathtt{pri}, v_{\sf B}^\mathtt{pri}\big) \Big)$ 
		\STATE $ v_{\sf B}^\mathtt{ext} = \left(\frac{1}{v_{\sf B}^{\mathtt{post}}} - \frac{1}{v_{\sf B}^{\mathtt{pri}}}\right)^{-1}$ \label{stmp_v_B_post} 
		\STATE $\bsm{x}_{\sf B}^\mathtt{ext} = v_{\sf B}^{\mathtt{ext}} \left(\frac{\bsm{x}_{\sf B}^{\mathtt{post}}}{v_{\sf B}^{\mathtt{post}}} - \frac{\bsm{x}_{\sf B}^{\mathtt{pri}}}{v_{\sf B}^{\mathtt{pri}}}\right)$ 
		\STATE $\bsm{x}_{\sf A}^\mathtt{pri} = \bsm{x}_{\sf B}^\mathtt{ext}, v_{\sf A}^\mathtt{pri} = v_{\sf B}^\mathtt{ext}$ \label{to_be_damped_B}
		\UNTIL{the stopping criterion is met}
	\end{algorithmic}
\end{algorithm}

\begin{remk}
	There exist several alternative approaches for calculating the posterior variance $v_{\sf B}^\mathtt{post}$, such as residual-based variance estimation and Monte-Carlo divergence estimation.
	The former approach relies on the assumption that the denoiser output admits the decomposition
	\begin{align}
		\bsm{x}_{\sf B}^\mathtt{post} = \bsm{x} + \mathcal{N}(\bsm{0}, v_{\sf B}^\mathtt{post}\bsm{I}), 
	\end{align}
	where the effective denoising error is assumed to be Gaussian and independent of the measurement noise $\bsm{n}$.
	Under this assumption, one may form an unbiased estimate of the variance as
	\begin{align}
		v_{\sf B}^\mathtt{post} = \frac{\left\|\bsm{y} - \bsm{A}\bsm{x}_{\sf B}^\mathtt{post}\right\|_2^2 - M\delta_0^2}{\left\|\bsm{A}\right\|_F^2}.
	\end{align}
	However, these Gaussianity and independence assumptions rarely hold for practical denoisers, making this estimate unreliable.
	Another option is to approximate the average Jacobian trace of the denoiser via Monte Carlo sampling.
	Let $\boldsymbol{\varepsilon} \sim\mathcal{N}(\bsm{0},\bsm{I})$ and define
	\begin{align}
		\bsm{x}_{\delta}^\mathtt{pri} 
		&= \bsm{x}_{\sf B}^\mathtt{pri} + \delta \boldsymbol{\varepsilon}, \\
		\bsm{x}_{\delta}^\mathtt{post}
		&= \bsm{x}_{\delta}^\mathtt{pri}
		+ v_{\sf B}^\mathtt{pri}\,
		\bsm{s}_{\boldsymbol{\theta}}
		\big(\bsm{x}_{\delta}^\mathtt{pri}, v_{\sf B}^\mathtt{pri}\big).
	\end{align}
	With $K$ Monte Carlo samples, the divergence is estimated as
	\begin{align}
		\alpha
		= \frac{1}{N K}
		\sum_{k=1}^K
		\frac{\big( (\bsm{x}_{\delta}^\mathtt{post})^{(k)}
		- (\bsm{x}_{{\sf B}}^\mathtt{post})^{(k)}\big)^\top \bsm{\varepsilon}^{(k)}}{\delta},
	\end{align}
	which yields the variance estimate
	\begin{align}
		v_{\sf B}^\mathtt{post}
		= \alpha v_{\sf B}^\mathtt{pri}.
	\end{align}
	While this method is theoretically grounded, it introduces $K$ additional denoiser evaluations, and may suffer from numerical instability for small $\delta$.
\end{remk}

\subsection{State Evolution}
\color{black}
In this subsection, we characterize the asymptotic \ac{MSE} performance of \ac{STMP} through \ac{SE}.  
The analysis basically follows the standard \ac{SE} framework in~\cite{se_analysis2018nips}.
We begin by defining the following asymptotic \ac{MSE} transfer function, which describes the denoising performance of the score-based denoiser:
\begin{align}
	\mathrm{MSE} (v_{\sf B}^\mathtt{pri}) &\triangleq \lim_{N\rightarrow \infty}\frac{1}{N} \mathbb{E} \left[ \left\| \mathsf{D}_{\bsm{\theta}}(\bsm{x}_{\sf B}^\mathtt{pri}, v_{\sf B}^\mathtt{pri}) - \bsm{x}\right\|^2_2 \right], \label{eqn:mse_function} \\ 
	\!\mathsf{D}_{\bsm{\theta}}(\bsm{x}_{\sf B}^\mathtt{pri}, v_{\sf B}^\mathtt{pri})&\triangleq \bsm{x}_{\sf B}^\mathtt{pri} + v_{\sf B}^\mathtt{pri} \bsm{s}_{\bsm{\theta}} (\bsm{x}_{\sf B}^\mathtt{pri}, v_{\sf B}^\mathtt{pri}).
\end{align}
The expectation in~\eqref{eqn:mse_function} is taken over the signal distribution $p(\bsm{x})$ and the effective residual error distribution $p(\bsm{x}_{\sf B}^{\mathtt{pri}}|\bsm{x})$ induced by the message-passing dynamics. 
In the large-system limit, the empirical average of the element-wise posterior variance concentrates to its expectation, yielding
\begin{align}
	\!v_{\sf B}^\mathtt{post} = \lim_{N\rightarrow\infty} \frac{1}{N} \mathbb{E} \left[ \sum_{n=1}^N \mathrm{Var} \big[x_n|\bsm{x}_{\sf B}^\mathtt{pri}\big] \right] = \mathrm{MSE}\big(v_{\sf B}^\mathtt{pri}\big).
\end{align}
We next state the assumptions under which the \ac{SE} recursion is formally established.

\begin{assump}[Right rotationally invariant] \label{assump1}
	The distribution of $\bsm{A}$ is equal to that of $\bsm{A} \bsm{V}$ for any orthogonal matrix $\bsm{V}$, independent of $\bsm{A}$.
\end{assump}
\begin{assump}[Uniformly Lipschitz continuous] \label{assump2}
	The denoiser $\mathsf{D}_{\bsm{\theta}}(\cdot, \cdot)$ is uniformly Lipschitz continuous with parameters $\alpha_1$, $\alpha_2$, $\alpha_3$, i.e., $\left\|\mathsf{D}_{\bsm{\theta}}(\bsm{x}_2, v_2) - \mathsf{D}_{\bsm{\theta}}(\bsm{x}_1, v_1) \right\|_2 
	\leq (\alpha_1 + \alpha_2 |v_2 - v_1|) \left\|\bsm{x}_2 - \bsm{x}_1\right\|_2 + \alpha_3 \sqrt{N} |v_2 - v_1|$,
	for any $\bsm{x}_1$, $\bsm{x}_2$, $v_1$, $v_2$, and $N$.
\end{assump}
\begin{assump}[Convergent under Gaussian noise] \label{assump3}
	The sequence of random vectors $\bsm{x} \in \mathbb{R}^N$ and estimators $\mathsf{D}_{\bsm{\theta}}(\cdot, \cdot)$ are convergent under Gaussian noise.
	Specifically, let $\bsm{w}_1, \bsm{w}_2 \in \mathbb{R}^N$ be two vectors where $(w_{1n}, w_{2n})$ are \ac{i.i.d.} with $(w_{1n}, w_{2n}) = \mathcal{N}(\bsm{0}, \bsm{C})$ for some positive definite covariance $\bsm{C} \in \mathbb{R}^{2\times 2}$. The following limits exist almost surely:
	$\lim_{N\to \infty} (1/N) \mathsf{D}_{\bsm{\theta}}(\bsm{x}+\bsm{w}_1, v_1)^\top \mathsf{D}_{\bsm{\theta}}(\bsm{x}+\bsm{w}_2, v_2)$, 
	$\lim_{N\to \infty} (1/N) \mathsf{D}_{\bsm{\theta}}(\bsm{x}+\bsm{w}_1, v_1)^\top \bsm{x}$,
	$\lim_{N\to \infty} (1/N) \bsm{x}^\top \bsm{w}_1$,
	$\lim_{N\to\infty} (1/N) \left\|\bsm{x}\right\|_2^2$,
	$\lim_{N\to\infty} (1/N) \mathsf{D}_{\bsm{\theta}}(\bsm{x}+\bsm{w}_1, v_1)^\top \bsm{w}_2$,
	for all $v_1$, $v_2$, and covariance matrices $\bsm{C}$.
	Moreover, the values of the limits are continuous in $\bsm{C}$, $v_1$, and $v_2$.
\end{assump}

\begin{theo}\label{theo:stmp_se}
		Under Assumptions~\ref{assump1}, \ref{assump2}, \ref{assump3}, 
		the asymptotic variance evolution of the \ac{STMP} algorithm with partial orthogonal $\bsm{A}$ converges to the following scalar \ac{SE} recursion:
		\begin{align}
			v_{\sf B}^\mathtt{pri} &= \frac{N}{M} \left(v_{\sf A}^\mathtt{pri} + \delta_0^2\right) - v_{\sf A}^\mathtt{pri}, \label{revision_eq:SE_v_B_pri} \\
			v_{\sf A}^\mathtt{pri} &= \left(\frac{1}{\mathrm{MSE} \big(v_{\sf B}^\mathtt{pri}\big)} - \frac{1}{v_{\sf B}^\mathtt{pri}}\right)^{-1}. \label{revision_eq:SE_v_A_pri}
		\end{align}
		Consequently, the asymptotic reconstruction \ac{MSE} at each iteration
		is predicted by $\mathrm{MSE}\big(v_{\sf B}^\mathtt{pri}\big)$.
\end{theo}

\begin{remk}[Practical evaluation of the \ac{MSE} transfer function] \label{remk:empirical_gaussian}
	Under the rigorous SE analysis~\cite{se_analysis2018nips}, the residual error entering the denoiser is characterized in an empirical Gaussian sense: as the system dimension tends to infinity, the empirical distribution formed by collecting all components of the residual vector converges to a scalar Gaussian distribution.
	This empirical Gaussian characterization, however, does not necessarily imply that the residual vector is generated according to an i.i.d. Gaussian model.
	This distinction is important when evaluating the denoiser \ac{MSE} transfer function $\mathrm{MSE} (v_{\sf B}^\mathtt{pri})$.
	For separable denoisers, the corresponding expectation in~\eqref{eqn:mse_function} can often be reduced to a scalar Gaussian integral and evaluated explicitly or numerically.
	In contrast, for a high-dimensional learned score denoiser, the exact transfer function is generally intractable.
	Therefore, in our practical \ac{SE} implementation, we approximate this transfer function by adopting the stronger i.i.d. Gaussian model $\bsm{x}_{\sf B}^\mathtt{pri} = \bsm{x} + \mathcal{N}(\bsm{0}, v_{\sf B}^\mathtt{pri}\bsm{I})$,
	and denote the resulting approximation by
	$\overline{\mathrm{MSE}}(v_{\sf B}^{\mathtt{pri}})$.
	This function is computed offline by applying the learned score denoiser to validation images corrupted by i.i.d. Gaussian noise at different variances, forming a lookup table that is used during SE evaluation.
\end{remk}
\color{black}

\section{Extension to Quantized Compressive Imaging}\label{sec:q_stmp}
Due to the finite precision of \ac{ADCs} and constraints on data storage, measurements in compressive imaging are often quantized.
This section generalizes \ac{STMP} to quantized systems and derives the corresponding \ac{SE} equations to characterize its asymptotic performance.
\subsection{Problem Statement}
Quantized compressive imaging aims to reconstruct an unknown signal $\bsm{x} \in \mathbb{R}^N$ from quantized and noisy measurements $\bsm{y}\in \mathcal{Q}^M$, given by
\begin{align}
	\bsm{y} = \mathsf{Q}(\bsm{z}+\bsm{n}), \quad \bsm{z} = \bsm{A}\bsm{x},
\end{align}
where $\mathsf{Q}(\cdot): \mathbb{R}^M \rightarrow \mathcal{Q}^M$ is an element-wise 
quantization operator.
Specifically, each scalar component $y_m = \mathsf{Q}(z_m + n_m)$ is mapped to one of the $2^B$ quantization levels contained in the discrete codebook $\mathcal{Q}$, where $B$ denotes the quantization bit depth.
The quantizer partitions the real line into $2^B$ contiguous intervals:
\begin{align}
	(-\infty, r_1],~ (r_1, r_2],~ \ldots,~ (r_{2^{B}-1}, \infty),
\end{align}
where $\{r_b\}_{b=0}^{2^{B}}$ are the decision thresholds with $r_0 = -\infty$ and $r_{2^{B}} = \infty$.
For a uniform mid-rise quantizer, the thresholds are evenly spaced with the quantization interval $\Delta > 0$, given by $r_b = (b - 2^{B-1})\Delta$ for $b = 1, 2, \ldots, 2^{B}-1$.
The output of the quantizer is then assigned as
\begin{align}
	y_m = r_b - \tfrac{\Delta}{2}, \quad \text{if } z_m + n_m \in (r_{b-1}, r_b].
\end{align}
When $B=1$, the model reduces to 1-bit quantized compressive sensing with binary measurements.

\subsection{\ac{Q-STMP} Algorithm}
\begin{figure}
	[t]
	\centering
	\includegraphics[width=.92\columnwidth]{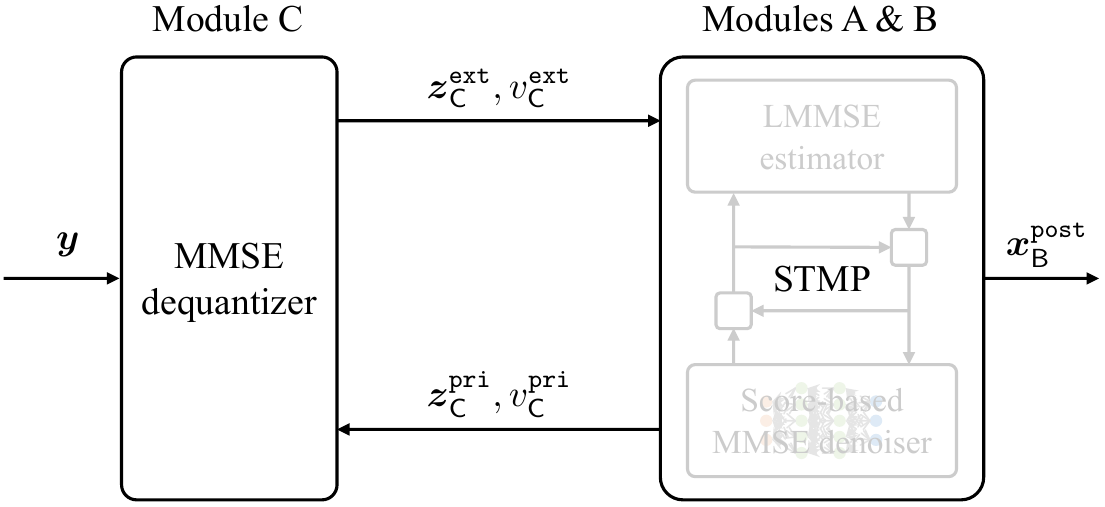}
	\caption{Diagram of the \ac{Q-STMP} algorithm.}
	\label{fig:q_stmp}
\end{figure}

The proposed extension draws inspiration from \ac{EP}~\cite{EP2001Minka} and its application to \ac{GLMs}~\cite{jiangzhu2018spl}.
As shown in Fig.~\ref{fig:q_stmp}, the key idea is to integrate the original \ac{STMP} algorithm with an additional dequantization module (module~C) in a turbo manner~\cite{turbo_codes}.
Module C performs component-wise \ac{MMSE} estimation of $\bsm{z}$ based on the quantization likelihood $p(\bsm{y}|\bsm{z}) = \prod_{m=1}^{M} p(y_m|z_m)$ and a Gaussian pseudo-prior $\mathcal{N}(\bsm{z}; \bsm{z}^\mathtt{pri}_{\sf C}, v^\mathtt{pri}_{\sf C} \bsm{I})$.
Then, it produces an extrinsic estimate of $\bsm{z}$, which is modeled as
\begin{align} \label{eqn:pseudo_linear_model}
	\bsm{z}^\mathtt{ext}_{\sf C} = \bsm{A}\bsm{x} + \bsm{n}^\mathtt{ext}_{\sf C}, \text{with } \bsm{n}^\mathtt{ext}_{\sf C} \sim \mathcal{N}(\bsm{0}, v^\mathtt{ext}_{\sf C} \bsm{I}),
\end{align}
where $\bsm{z}^\mathtt{ext}_{\sf C}$ and $v^\mathtt{ext}_{\sf C}$ represent the effective linear measurements and noise variance after dequantization.
These pseudo-measurements are then fed into the \ac{STMP} algorithm to update the belief on $\bsm{x}$ and consequently refine the prior of $\bsm{z} = \bsm{A}\bsm{x}$ provided to module~C.
The algorithm iterates between module~C and the \ac{STMP} block (modules~A and~B), progressively improving both the dequantized signal estimate and the reconstructed image.

Specifically, we initialize the prior mean and variance for module~C as $\mathcal{N}(\bsm{z}; \bsm{z}^\mathtt{pri}_{\sf C}, v^\mathtt{pri}_{\sf C} \bsm{I})$.
For each scalar entry $z_m$, the posterior mean and variance are given by the component-wise \ac{MMSE} estimator:
\begin{align}
	z^\mathtt{post}_{{\sf C}, m} &= \mathbb{E} \big[z_m \big|z^\mathtt{pri}_{{\sf C}, m}\big], \\
	v^\mathtt{post}_{{\sf C}, m} &= \mathrm{Var} \big[z_m\big|z_{{\sf C}, m}^\mathtt{pri}\big] ,
\end{align}
where the expectation is taken \ac{w.r.t.} the posterior distribution
\begin{align} \label{eqn:module_C_posterior}
	p(z_m|y_m) \propto p(y_m|z_m) \mathcal{N}\big(z_m; z^\mathtt{pri}_{{\sf C}, m}, v^\mathtt{pri}_{\sf C}\big).
\end{align}
For the considered quantization model, measurement $y_m$ indicates that the noisy observation satisfies $z_m + n_m \in (r_{b_m -1}, r_{b_m} ]$.
Equivalently, conditioned on $z_m$, the likelihood is the probability that a Gaussian random variable with mean $z_m$ and variance $\delta_0^2$ falls inside this interval, i.e.,
\begin{align}
	p(y_m|z_m) = \Phi \left(\frac{r_{b_m} - z_m}{\delta_0}\right) - \Phi \left(\frac{r_{b_m - 1} - z_m}{\delta_0}\right),
\end{align}
where $\Phi (x) \triangleq (1/\sqrt{2\pi}) \int_{-\infty}^x e^{-t^2/2} \mathrm{d} t$
denotes the \ac{CDF} of the standard normal distribution.
Therefore, the product in~\eqref{eqn:module_C_posterior} is proportional to a Gaussian distribution truncated to a normalized interval.
Define the normalized truncation bounds:
\begin{align}
	\eta_m^{\sf u} \triangleq \frac{r_{b_m} - z^\mathtt{pri}_{{\sf C}, m}}{\sqrt{v^\mathtt{pri}_{\sf C}+ \delta_0^2}},
	\quad \eta_m^{\sf l} \triangleq \frac{r_{b_m-1} - z^\mathtt{pri}_{{\sf C}, m} }{\sqrt{v^\mathtt{pri}_{\sf C}+ \delta_0^2}}.
\end{align}
Denote the \ac{PDF} of the standard normal distribution by $\phi (x) \triangleq (1/\sqrt{2\pi}) e^{-x^2/2}$.
Using standard identities for truncated Gaussian distributions~\cite{GTurboSR}, the posterior mean and variance become 
\begin{align}
	z^\mathtt{post}_{{\sf C}, m} &= \frac{\int z_m p(y_m|z_m) \mathcal{N}\big(z_m; z^\mathtt{pri}_{{\sf C}, m}, v^\mathtt{pri}_{\sf C}\big) \mathrm{d} z_m }{\int p(y_m|z_m) \mathcal{N}\big(z_m; z^\mathtt{pri}_{{\sf C}, m}, v^\mathtt{pri}_{\sf C}\big) \mathrm{d} z_m} \nonumber \\
	&= z^\mathtt{pri}_{{\sf C}, m} - \frac{v^\mathtt{pri}_{\sf C}}{\sqrt{v^\mathtt{pri}_{\sf C}+ \delta_0^2}} \frac{\phi(\eta_m^{\sf u}) - \phi(\eta_m^{\sf l})}{\Phi(\eta_m^{\sf u}) - \Phi(\eta_m^{\sf l})}, \\
	v^\mathtt{post}_{{\sf C}, m} &= \frac{\int (z_m - z_{{\sf C}, m}^\mathtt{post})^2 p(y_m|z_m) \mathcal{N}\big(z_m; z^\mathtt{pri}_{{\sf C}, m}, v^\mathtt{pri}_{\sf C}\big) \mathrm{d} z_m }{\int p(y_m|z_m) \mathcal{N}\big(z_m; z^\mathtt{pri}_{{\sf C}, m}, v^\mathtt{pri}_{\sf C}\big) \mathrm{d} z_m} \nonumber \\
	&=v^\mathtt{pri}_{\sf C} - \frac{(v^\mathtt{pri}_{\sf C})^2}{v^\mathtt{pri}_{\sf C} + \delta_0^2}
	\Bigg( 
	\frac{\eta_m^{\sf u}\phi(\eta_m^{\sf u}) - \eta_m^{\sf l}\phi(\eta_m^{\sf l})}{\Phi(\eta_m^{\sf u}) - \Phi(\eta_m^{\sf l})} +  \nonumber \\
	&~~~~~~~~~~~~~~~~~~~~~~~~~~~~ \left(\frac{\phi(\eta_m^{\sf u}) - \phi(\eta_m^{\sf l})}{\Phi(\eta_m^{\sf u}) - \Phi(\eta_m^{\sf l})}\right)^2 \Bigg).
\end{align}
The posterior variance from module~C is averaged over all dimensions as
\begin{align} \label{eqn:module_C_post_variance}
	v^\mathtt{post}_{\sf C} = \frac{1}{M} \sum_{m=1}^M v^\mathtt{post}_{{\sf C}, m}.
\end{align}
The corresponding extrinsic variance and mean are respectively given by
\begin{align}
	v^\mathtt{ext}_{\sf C} &= \left(\frac{1}{v^\mathtt{post}_{\sf C}} - \frac{1}{v^\mathtt{pri}_{\sf C}}\right)^{-1}, \label{eqn:v_C_ext} \\
	\bsm{z}^\mathtt{ext}_{\sf C} &= v^\mathtt{ext}_{\sf C} \left(\frac{\bsm{z}^\mathtt{post}_{\sf C}}{v^\mathtt{post}_{\sf C}} - \frac{\bsm{z}^\mathtt{pri}_{\sf C}}{v^\mathtt{pri}_{\sf C}} \right).
\end{align}
The extrinsic output is then used as the pseudo measurement in~\eqref{eqn:pseudo_linear_model} and fed into the \ac{STMP} block.
Within modules~A and~B, one or multiple \ac{STMP} iterations may be executed before passing updated messages back to module~C.
Specifically, the \ac{STMP} block produces $\bsm{x}_{\sf B}^\mathtt{ext}$ and $v_{\sf B}^\mathtt{ext}$, which are used to construct the prior for the next module-C update as $\bsm{z}_{\sf C}^\mathtt{pri} = \bsm{A}\bsm{x}_{\sf B}^\mathtt{ext}$, and
 $v_{\sf C}^\mathtt{pri} = \mathrm{tr}\big(v_{\sf B}^\mathtt{ext}\bsm{A}\bsm{A}^\top\big)/M$.
For partial orthogonal $\bsm{A}$, this further reduces to $v_{\sf C}^\mathtt{pri}  = v_{\sf B}^\mathtt{ext}$.
The complete algorithm is summarized in Algorithm~\ref{alg:quantized_stmp}.

\subsection{State Evolution}
In the large-system limit, the empirical average in~\eqref{eqn:module_C_post_variance} concentrates around its ensemble expectation, allowing $v_{\sf C}^\mathtt{post}$ to be expressed as an expectation \ac{w.r.t.} the joint distribution $p(z_{\sf C}^\mathtt{pri}, y)$.
As all measurement components are statistically identical in this limit, the component index $m$ is dropped hereafter.
The resulting transfer function that characterizes the prior-to-posterior variance mapping of module~C is given by
\begin{align}
	v_{\sf C}^\mathtt{post} &= v^\mathtt{pri}_{\sf C} - \frac{(v^\mathtt{pri}_{\sf C})^2}{v^\mathtt{pri}_{\sf C} + \delta_0^2} \mathbb{E}_{p(z_{\sf C}^\mathtt{pri}, y)}
	\Bigg[\frac{\eta^{\sf u}\phi(\eta^{\sf u}) - \eta^{\sf l}\phi(\eta^{\sf l})}{\Phi(\eta^{\sf u}) - \Phi(\eta^{\sf l})} + \nonumber \\
	&~~~~~~~~~~~~~~~~~~~~~~~~~~~~ \left(\frac{\phi(\eta^{\sf u}) - \phi(\eta^{\sf l})}{\Phi(\eta^{\sf u}) - \Phi(\eta^{\sf l})}\right)^2 \Bigg] \label{eqn:v_C_post_SE} \\
	&=v^\mathtt{pri}_{\sf C} - (v^\mathtt{pri}_{\sf C})^2 \mathbb{E}_{p(z_{\sf C}^\mathtt{pri}, y)}
	\Bigg[-\frac{\Psi'' \big(y; z_{{\sf C}}^{\mathtt{pri}}, v_{\sf C}^{\mathtt{pri}} + \delta_0^2\big)}{\Psi \big(y; z_{{\sf C}}^{\mathtt{pri}}, v_{\sf C}^{\mathtt{pri}} + \delta_0^2\big)} + \nonumber \\
	&~~~~~~~~~~~~~~~~~~~ \left(\frac{\Psi^{\prime} \big(y; z_{{\sf C}}^{\mathtt{pri}}, v_{\sf C}^{\mathtt{pri}} + \delta_0^2\big)}{\Psi \big(y; z_{{\sf C}}^{\mathtt{pri}}, v_{\sf C}^{\mathtt{pri}} + \delta_0^2\big)}\right)^2 \Bigg],
\end{align}
where 
\begin{align}
	\Psi \big(y; z_{{\sf C}}^{\mathtt{pri}}, v_{\sf C}^{\mathtt{pri}} + \delta_0^2\big) &\triangleq \Phi\big(\eta^{\sf u}\big) - \Phi\big(\eta^{\sf l}\big), \\
	\Psi' \big(y; z_{{\sf C}}^{\mathtt{pri}}, v_{\sf C}^{\mathtt{pri}} + \delta_0^2\big) & =  -\frac{\phi(\eta^{\sf u}) - \phi(\eta^{\sf l})}{\sqrt{v^\mathtt{pri}_{\sf C}+ \delta_0^2}}, \\
	\Psi'' \big(y; z_{{\sf C}}^{\mathtt{pri}}, v_{\sf C}^{\mathtt{pri}} + \delta_0^2\big) & = - \frac{\eta^{\sf u}\phi(\eta^{\sf u}) - \eta^{\sf l}\phi(\eta^{\sf l})}{v_{\sf C}^\mathtt{pri} + \delta_0^2}.
\end{align}
The joint distribution of $z_{\sf C}^\mathtt{pri}$ and $y$ follows from~\cite{GTurboSR} as
\begin{align}
	p\big(z_{\sf C}^\mathtt{pri}, y\big) = \mathcal{N}\big(z_{\sf C}^\mathtt{pri}; 0, v_z - v_{\sf C}^\mathtt{pri}\big) \Psi \big(y; z_{{\sf C}}^{\mathtt{pri}}, v_{\sf C}^{\mathtt{pri}} + \delta_0^2\big),
\end{align}
where $v_z = \sum_{m=1}^M\mathbb{E}\left[|z_m|^2\right] / M$.
The expectation in~\eqref{eqn:v_C_post_SE} can thus be written as an integral over the standard Gaussian measure $\mathrm{D}z = (1/\sqrt{2\pi})e^{-z^2/2}\mathrm{d}z$.
Since the first term inside the expectation evaluates to zero, we obtain
\begin{align} \label{eqn:v_C_post_SE_final}
	v_{\sf C}^\mathtt{post} &= v^\mathtt{pri}_{\sf C} - (v^\mathtt{pri}_{\sf C})^2 \times \nonumber \\
	&~~~~\underbrace{\sum_{y\in\mathcal{Q}} \int \mathrm{D}z \frac{\big(\Psi^\prime(y; \sqrt{v_z - v_{\sf C}^{\mathtt{pri}}}z, v_{\sf C}^{\mathtt{pri}} + \delta_0^2)\big)^2}{\Psi(y; \sqrt{v_z - v_{\sf C}^{\mathtt{pri}}}z, v_{\sf C}^{\mathtt{pri}} + \delta_0^2)}}_{\triangleq \vartheta \big(v_{\sf C}^\mathtt{pri}\big)}.
\end{align}
Substituting~\eqref{eqn:v_C_post_SE_final} into~\eqref{eqn:v_C_ext} gives the extrinsic variance passed from module~C to module~A:
\begin{align} \label{eqn:v_C_ext_se}
	v_{\sf C}^\mathtt{ext} = \frac{1}{\vartheta \big(v_{\sf C}^\mathtt{pri}\big)} - v_{\sf C}^\mathtt{pri}.
\end{align}
Finally, substituting~\eqref{eqn:v_C_ext_se} into the variance recursions of modules~A and~B, and using the identity $v_{\sf C}^\mathtt{pri} = v_{\sf A}^\mathtt{pri}$,
we obtain the \ac{SE} of \ac{Q-STMP} in the following theorem.

\begin{algorithm}[t]
	\caption{\ac{Q-STMP} Algorithm}
	\label{alg:quantized_stmp}
	\begin{algorithmic}[1]
		\STATE {\bfseries Input:} $\bsm{A}$, $\bsm{y}$, $\delta_0^2$, $\bsm{x}_{\sf A}^\mathtt{pri}$, $v_{\sf A}^\mathtt{pri}$
		\STATE {\bfseries Output:} $\bsm{x}_{\sf B}^\mathtt{post}$
		\STATE $\bsm{z}_{\sf C}^\mathtt{pri} = \bsm{A}\bsm{x}_{\sf A}^\mathtt{pri}$, $v_{\sf C}^\mathtt{pri} = \frac{\mathrm{tr}\big(v_{\sf A}^\mathtt{pri}\bsm{A}\bsm{A}^\top\big)}{M}$
		\REPEAT
		\STATE \% MMSE dequantizer (module~C)
		\STATE $z^\mathtt{post}_{{\sf C}, m} = z^\mathtt{pri}_{{\sf C}, m} - \frac{v^\mathtt{pri}_{\sf C}}{\sqrt{v^\mathtt{pri}_{\sf C}+ \delta_0^2}} \frac{\phi(\eta_m^{\sf u}) - \phi(\eta_m^{\sf l})}{\Phi(\eta_m^{\sf u}) - \Phi(\eta_m^{\sf l})}, \forall m$
		\STATE $v^\mathtt{post}_{{\sf C}}= v^\mathtt{pri}_{\sf C} - \frac{(v^\mathtt{pri}_{\sf C})^2}{M(v^\mathtt{pri}_{\sf C} + \delta_0^2)}
		\sum\limits_{m=1}^M \Big( 
		\frac{\eta_m^{\sf u}\phi(\eta_m^{\sf u}) - \eta_m^{\sf l}\phi(\eta_m^{\sf l})}{\Phi(\eta_m^{\sf u}) - \Phi(\eta_m^{\sf l})} + \left(\frac{\phi(\eta_m^{\sf u}) - \phi(\eta_m^{\sf l})}{\Phi(\eta_m^{\sf u}) - \Phi(\eta_m^{\sf l})}\right)^2 \Big)$
		\STATE 	$v^\mathtt{ext}_{\sf C} = \left(\frac{1}{v^\mathtt{post}_{\sf C}} - \frac{1}{v^\mathtt{pri}_{\sf C}}\right)^{-1}$
		\STATE $\bsm{z}^\mathtt{ext}_{\sf C} = v^\mathtt{ext}_{\sf C} \left(\frac{\bsm{z}^\mathtt{post}_{\sf C}}{v^\mathtt{post}_{\sf C}} - \frac{\bsm{z}^\mathtt{pri}_{\sf C}}{v^\mathtt{pri}_{\sf C}} \right)$
		\STATE \% LMMSE estimator (module~A)		
		\color{black}
		\STATE $\bsm{x}_{\sf A}^{\mathtt{post}} = \bsm{x}_{\sf A}^\mathtt{pri} + v_{\sf A}^\mathtt{pri}\bsm{A}^\top \big(v_{\sf A}^\mathtt{pri}\bsm{A}\bsm{A}^\top + v_{\sf C}^\mathtt{ext}\bsm{I}\big)^{-1} \big(\bsm{z}_{\sf C}^\mathtt{ext} - \bsm{A} \bsm{x}_{\sf A}^\mathtt{pri}\big)$
		\STATE $v_{\sf A}^\mathtt{post} = v_{\sf A}^\mathtt{pri} - \frac{(v_{\sf A}^\mathtt{pri})^2}{N} \mathrm{tr} \big(\bsm{A}^\top \big(v_{\sf A}^\mathtt{pri}\bsm{A} \bsm{A}^\top + v_{\sf C}^\mathtt{ext} \bsm{I}\big)^{-1}\bsm{A}\big)$\color{black}
		
		\STATE $v_{\sf B}^{\mathtt{pri}} = v_{\sf A}^\mathtt{ext} = \left(\frac{1}{v_{\sf A}^{\mathtt{post}}} - \frac{1}{v_{\sf A}^{\mathtt{pri}}}\right)^{-1}$ 
		\STATE $\bsm{x}_{\sf B}^{\mathtt{pri}} = \bsm{x}_{\sf A}^\mathtt{ext} = v_{\sf A}^{\mathtt{ext}} \left(\frac{\bsm{x}_{\sf A}^{\mathtt{post}}}{v_{\sf A}^{\mathtt{post}}} - \frac{\bsm{x}_{\sf A}^{\mathtt{pri}}}{v_{\sf A}^{\mathtt{pri}}}\right)$ 
		\STATE \% Score-based MMSE denoiser (module~B)
		\STATE $\bsm{x}_{\sf B}^\mathtt{post} = \bsm{x}_{\sf B}^\mathtt{pri} + v_{\sf B}^\mathtt{pri} \bsm{s}_{\bsm{\theta}} \big(\bsm{x}_{\sf B}^\mathtt{pri}, v_{\sf B}^\mathtt{pri}\big)$ 
		\STATE $v_{\sf B}^\mathtt{post} = v_{\sf B}^\mathtt{pri} + \frac{(v_{\sf B}^\mathtt{pri})^2}{N} \mathrm{tr} \Big(\bsm{S}_{\bsm{\phi}} \big(\bsm{x}_{\sf B}^\mathtt{pri}, v_{\sf B}^\mathtt{pri}\big) \Big)$ 
		\STATE $ v_{\sf A}^{\mathtt{pri}} = v_{\sf B}^\mathtt{ext} = \left(\frac{1}{v_{\sf B}^{\mathtt{post}}} - \frac{1}{v_{\sf B}^{\mathtt{pri}}}\right)^{-1}$ 
		\STATE $ \bsm{x}_{\sf A}^{\mathtt{pri}} = \bsm{x}_{\sf B}^\mathtt{ext} = v_{\sf B}^{\mathtt{ext}} \left(\frac{\bsm{x}_{\sf B}^{\mathtt{post}}}{v_{\sf B}^{\mathtt{post}}} - \frac{\bsm{x}_{\sf B}^{\mathtt{pri}}}{v_{\sf B}^{\mathtt{pri}}}\right)$ 
		\STATE $\bsm{z}_{\sf C}^\mathtt{pri} = \bsm{A}\bsm{x}_{\sf A}^\mathtt{pri}$, $v_{\sf C}^\mathtt{pri} = \frac{\mathrm{tr}\big(v_{\sf A}^\mathtt{pri}\bsm{A}\bsm{A}^\top\big)}{M}$
		\UNTIL{the stopping criterion is met}
	\end{algorithmic}
\end{algorithm}

\begin{theo}
	Under Assumptions~\ref{assump1}, \ref{assump2}, \ref{assump3}, 
	the asymptotic variance evolution of the \ac{Q-STMP} algorithm with partial orthogonal $\bsm{A}$ converges to the following scalar \ac{SE} recursion:
	\begin{align}
		\vartheta(v_{\sf A}^{\mathtt{pri}}) &= \sum_{y\in\mathcal{Q}} \int \mathrm{D}z \frac{\big(\Psi^\prime(y; \sqrt{v_z - v_{\sf A}^{\mathtt{pri}}}z, v_{\sf A}^{\mathtt{pri}} + \delta_0^2)\big)^2}{\Psi(y; \sqrt{v_z - v_{\sf A}^{\mathtt{pri}}}z, v_{\sf A}^{\mathtt{pri}} + \delta_0^2)}, \\
		v_{\sf B}^\mathtt{pri} &= \frac{N}{M} \frac{1}{\vartheta(v_{\sf A}^{\mathtt{pri}})} - v_{\sf A}^\mathtt{pri}, \\
		v_{\sf A}^\mathtt{pri} &= \left(\frac{1}{\mathrm{MSE} \big(v_{\sf B}^\mathtt{pri}\big)} - \frac{1}{v_{\sf B}^\mathtt{pri}}\right)^{-1}.
	\end{align}
\end{theo}

\section{Experiments} \label{sec:experiments}
In this section, we evaluate the performance of the proposed methods on the FFHQ~\cite{ffhq} and LDCT~\cite{ldct2021} datasets.
FFHQ is a high-quality human face dataset consisting of RGB images with a resolution of $3\times256\times256$.
LDCT is a medical imaging dataset composed of low-dose computed tomography scans of multiple anatomical sites (including head, chest, and abdomen) with a spatial resolution of $512\times 512$, which is commonly adopted for evaluating reconstruction algorithms in compressive imaging.
We adopt the pre-trained score models from~\cite{score_sde} and~\cite{score_mri} as the first-order score networks for the FFHQ and LDCT datasets, respectively.
However, to the best of our knowledge, no publicly available second-order score models exist for either dataset.
For training the second-order score model, we adopt the NCSN++ architecture~\cite{score_sde} that outputs only the diagonal entries of the second-order score, yielding the same output dimension as the first-order score model.
The focus on the diagonal is due to the fact that only the trace of the second-order score is pertinent to the posterior variance evaluation.
The training of the second-order score model follows the same default settings as its first-order counterpart~\cite{score_sde}.
For both datasets, we evaluate the proposed method and several benchmark approaches on a set of $1,000$ test images.

We set the noise-level weights in the first- and second-order score matching objectives to $\lambda_1(\sigma_i)=\sigma_i^2$ and $\lambda_2(\sigma_i)=\sigma_i^4$, respectively~\cite{lu2022maximum}.
During inference, \ac{STMP} and \ac{Q-STMP} are initialized with $\bsm{x}_{\sf A}^\mathtt{pri} = \frac{1}{2} \bsm{1}$, where $\bsm{1}$ is an all-one vector, and $v_{\sf A}^\mathtt{pri} = 0.25$.
The algorithms terminate when the relative change between two consecutive iterations satisfies
$\| \bsm{x}_{\sf B}^{\mathtt{post}, (t)} -  \bsm{x}_{\sf B}^{\mathtt{post}, (t-1)} \|_2 / \| \bsm{x}_{\sf B}^{\mathtt{post}, (t-1)} \|_2 \leq 10^{-4}$, or when the maximum number of $50$ iterations is reached.
For the baselines considered in this paper, we follow the default hyperparameter settings provided in the corresponding official implementation or the original paper whenever available.

\begin{figure*}[t]
	\centering
	\begin{minipage}{0.325\linewidth}
		\centering
		\includegraphics[width=\linewidth]{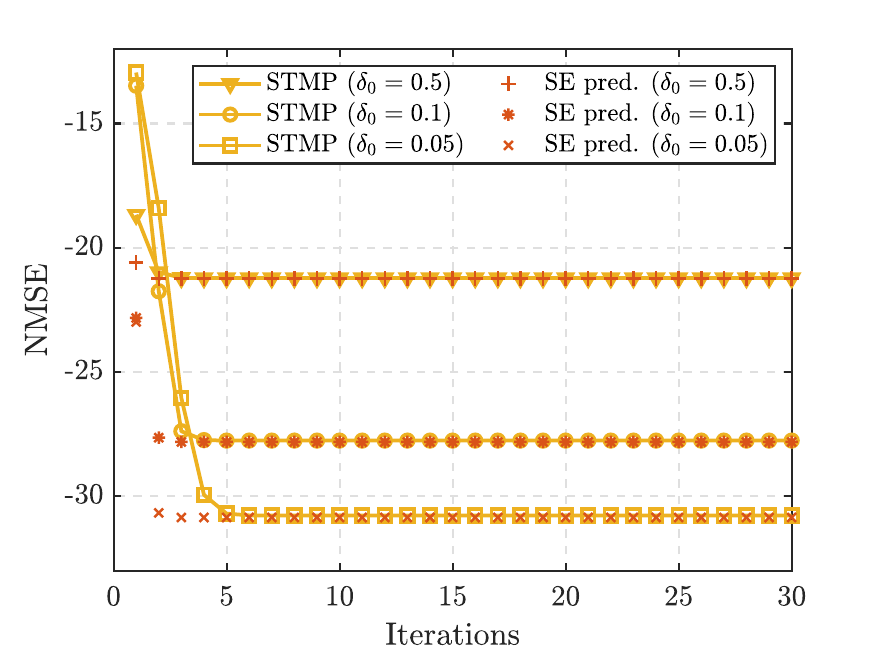}\\
		\footnotesize{(a) $M/N = 0.8$, $\beta = 1$}
	\end{minipage}
	\begin{minipage}{0.325\linewidth}
		\centering		
		\includegraphics[width=\linewidth]{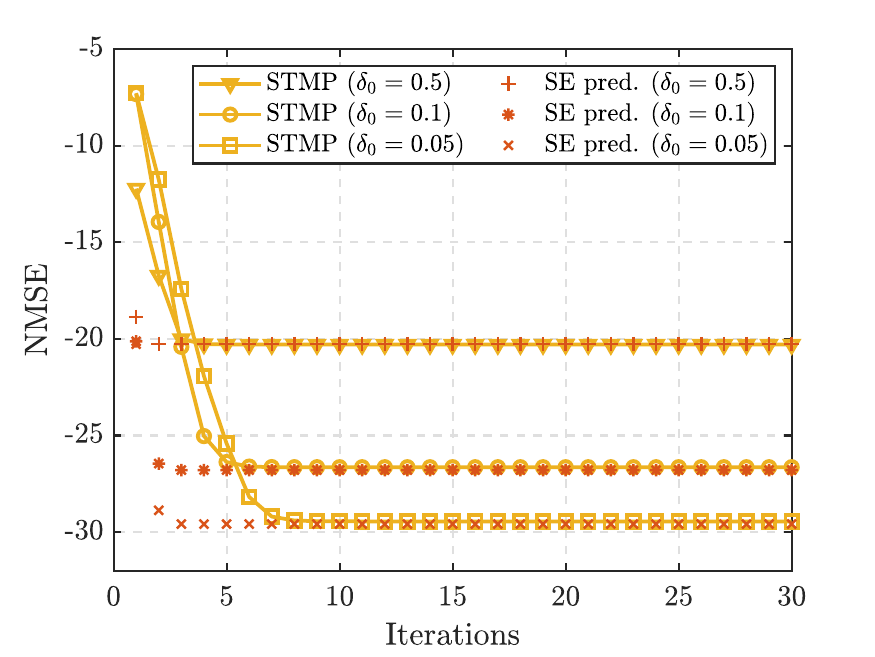}\\
		\footnotesize{(b) $M/N = 0.5$, $\beta = 0.8$}
	\end{minipage}
	\begin{minipage}{0.325\linewidth}
		\centering		
		\includegraphics[width=\linewidth]{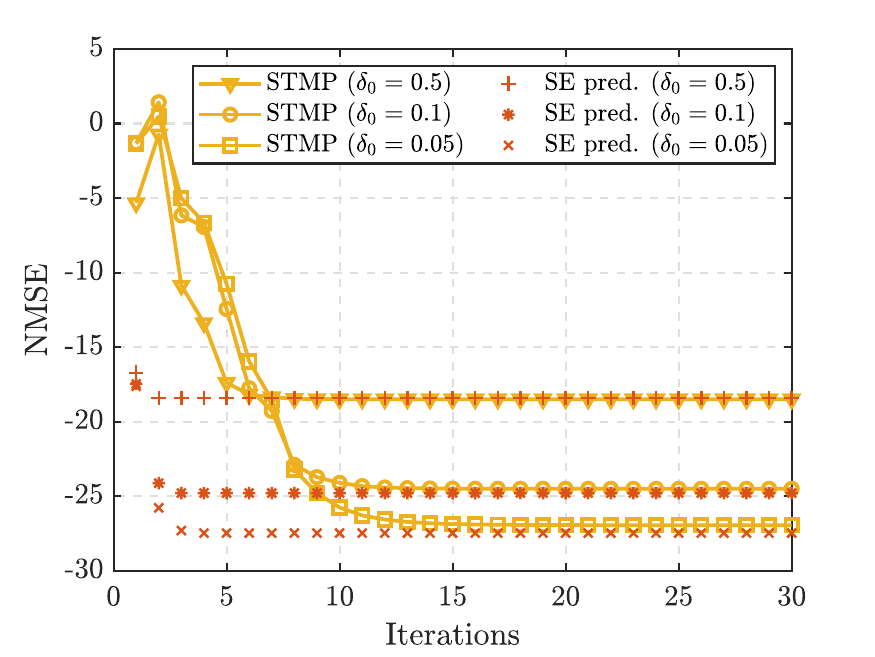}\\
		\footnotesize{(c) $M/N = 0.2$, $\beta = 0.6$}
	\end{minipage}
	\caption{\label{fig:stmp_convergence}
		Convergence behaviors and \ac{SE} predictions of \ac{STMP} on FFHQ using randomly row-selected \ac{DCT} sensing matrices.}
\end{figure*}

We consider the following types of measurement matrices:
\begin{itemize}
	\item \textbf{Randomly row-selected \ac{DCT} matrix~\cite{turbo_cs, oamp}:}
	The matrix takes the form of $\bsm{A} = \bsm{S} \bsm{W} \bsm{\Theta}$, where $\bsm{S} \in \mathbb{R}^{M \times N}$ is a random row selection matrix consisting of randomly selected rows from a permutation matrix, $\bsm{W} \in \mathbb{R}^{N \times N}$ is a \ac{DCT} matrix, and $\bsm{\Theta} \in \mathbb{R}^{N \times N}$ is a diagonal matrix with random signs ($1$ or $-1$) on the diagonal.
	
	\item \textbf{\Ac{DFT} matrix with equispaced Cartesian masks~\cite{score_mri}:} 
	The matrix takes the form $\bsm{A} = \bsm{S}\bsm{F}$, where $\bsm{F} \in \mathbb{C}^{N \times N}$ denotes the normalized \ac{DFT} matrix implemented via \ac{FFT}, and $\bsm{S} \in \mathbb{R}^{M \times N}$ is a row selection matrix determined by an equispaced Cartesian sampling mask. Specifically, the low-frequency columns are fully sampled, while the remaining high-frequency columns are uniformly subsampled.
	
	\item \textbf{General unitarily-invariant matrix:}
	Realizations of $\bsm{A}$ are constructed from the \ac{SVD} $\bsm{A} = \bsm{U}\bsm{\Sigma}\bsm{V}^\top$.
	The nonzero singular values are chosen to follow a geometric progression satisfying $\lambda_i/\lambda_{i+1} = \kappa^{1/M}, \forall i$, with the normalization $\sum_{i=1}^M \lambda_i = N$.
	Here, $\kappa \geq 1$ is the condition number of $\bsm{A}$.
	The singular vector matrices $\bsm{U} \in \mathbb{R}^{M\times M}$ and $\bsm{V} \in \mathbb{R}^{N\times N}$ are drawn uniformly at random from the group of orthogonal matrices, i.e., from the Haar distribution.
	
	\item \textbf{I.i.d. random Gaussian matrix:}
	The matrix $\bsm{A} \in \mathbb{R}^{M \times N}$ has entries drawn \ac{i.i.d.} from a Gaussian distribution $\mathcal{N}(0, 1/N)$.
\end{itemize}

\newcommand{\imwidth}{1.6cm}
\begin{figure*}
	\centering
	\def\arraystretch{0.7}
	\setlength\tabcolsep{0.03cm}
	\begin{tabular}{lccccccccc}
		\multirow{2}{*}{\raisebox{-0.5cm}[0pt][0pt]{\rotatebox{90}{\scriptsize $\delta_0 = 0.05$}}} &
		\includegraphics[width=\imwidth,height=\imwidth]{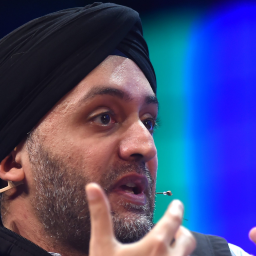}
		& \includegraphics[width=\imwidth,height=\imwidth]{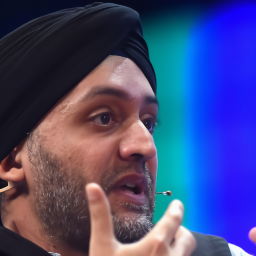}
		& \includegraphics[width=\imwidth,height=\imwidth]{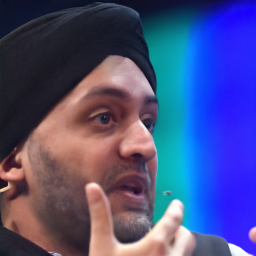}
		& \includegraphics[width=\imwidth,height=\imwidth]{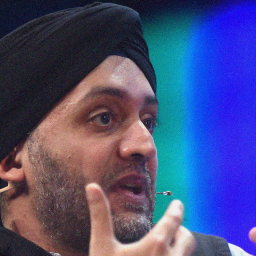}
		& \includegraphics[width=\imwidth,height=\imwidth]{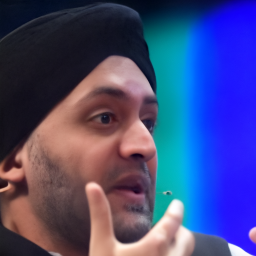}
		& \includegraphics[width=\imwidth,height=\imwidth]{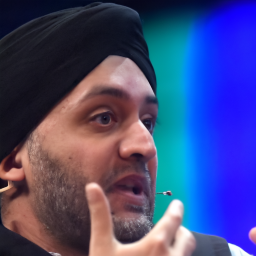}
		& \includegraphics[width=\imwidth,height=\imwidth]{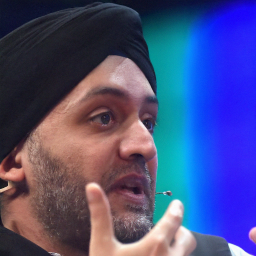}
		& \includegraphics[width=\imwidth,height=\imwidth]{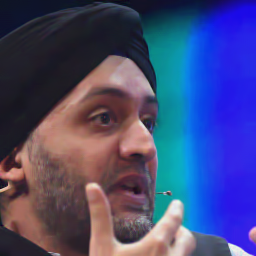}
		& \includegraphics[width=\imwidth,height=\imwidth]{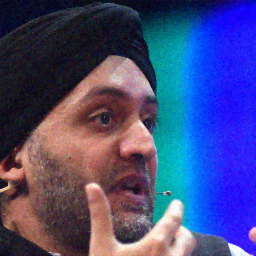}  \\
		&
		\includegraphics[width=\imwidth,height=\imwidth]{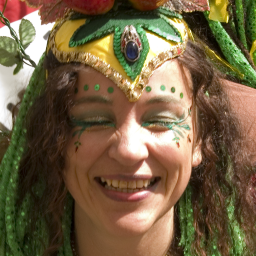}
		& \includegraphics[width=\imwidth,height=\imwidth]{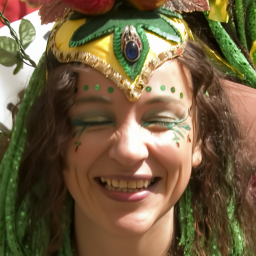}
		& \includegraphics[width=\imwidth,height=\imwidth]{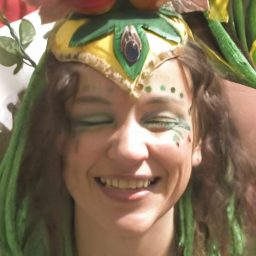}
		& \includegraphics[width=\imwidth,height=\imwidth]{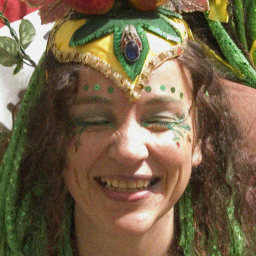}
		& \includegraphics[width=\imwidth,height=\imwidth]{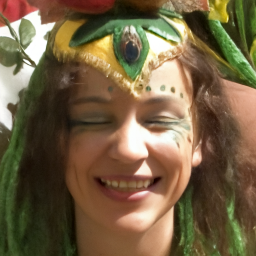}
		& \includegraphics[width=\imwidth,height=\imwidth]{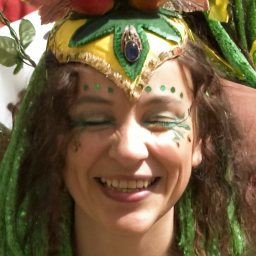}
		& \includegraphics[width=\imwidth,height=\imwidth]{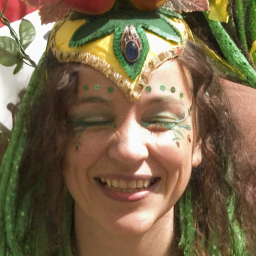}
		& \includegraphics[width=\imwidth,height=\imwidth]{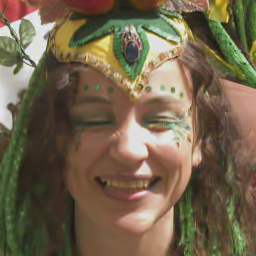}
		& \includegraphics[width=\imwidth,height=\imwidth]{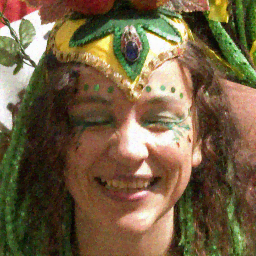} \\
		\multicolumn{10}{c}{\vspace{-.2em}} \\
		\multirow{2}{*}{\raisebox{-0.45cm}[0pt][0pt]{\rotatebox{90}{\scriptsize $\delta_0 = 0.5$}}} &
		\includegraphics[width=\imwidth,height=\imwidth]{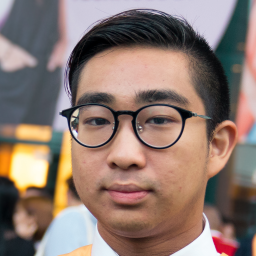}
		& \includegraphics[width=\imwidth,height=\imwidth]{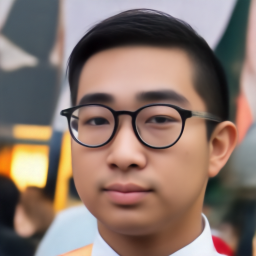}
		& \includegraphics[width=\imwidth,height=\imwidth]{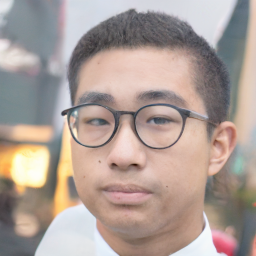}
		& \includegraphics[width=\imwidth,height=\imwidth]{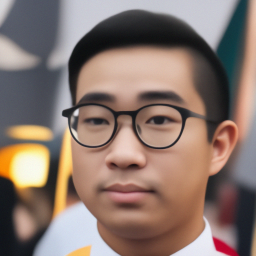}
		& \includegraphics[width=\imwidth,height=\imwidth]{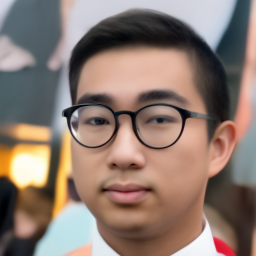}
		& \includegraphics[width=\imwidth,height=\imwidth]{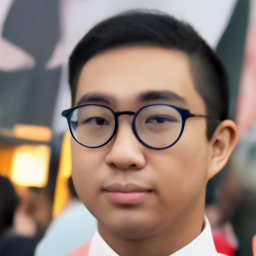}
		& \includegraphics[width=\imwidth,height=\imwidth]{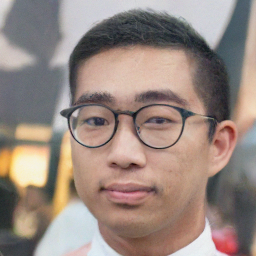}
		& \includegraphics[width=\imwidth,height=\imwidth]{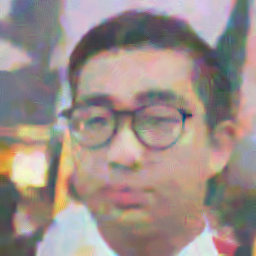}
		& \includegraphics[width=\imwidth,height=\imwidth]{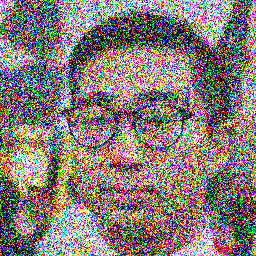} \\
		&
		\includegraphics[width=\imwidth,height=\imwidth]{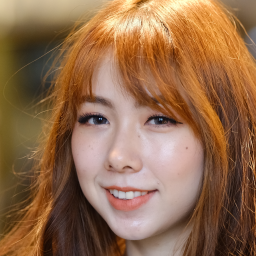}
		& \includegraphics[width=\imwidth,height=\imwidth]{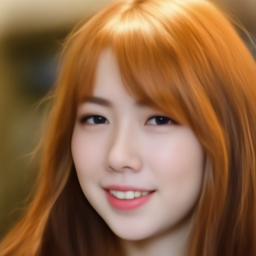}
		& \includegraphics[width=\imwidth,height=\imwidth]{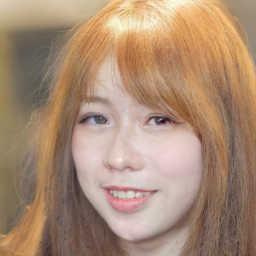}
		& \includegraphics[width=\imwidth,height=\imwidth]{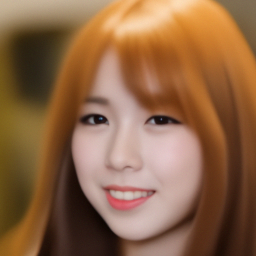}
		& \includegraphics[width=\imwidth,height=\imwidth]{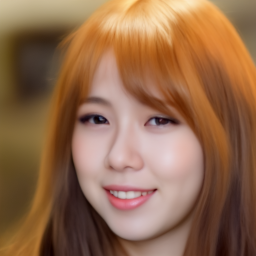}
		& \includegraphics[width=\imwidth,height=\imwidth]{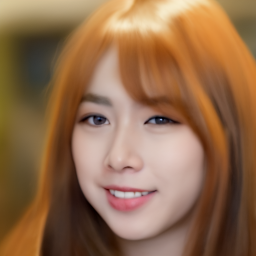}
		& \includegraphics[width=\imwidth,height=\imwidth]{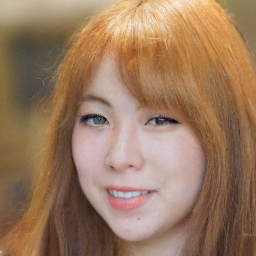}
		& \includegraphics[width=\imwidth,height=\imwidth]{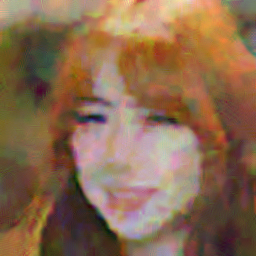}
		& \includegraphics[width=\imwidth,height=\imwidth]{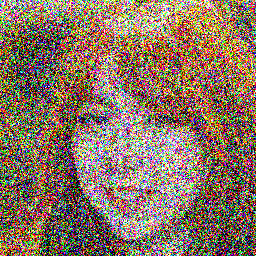}  \\
		& \scriptsize Ground truth & \scriptsize STMP & \scriptsize DPS & \scriptsize DiffPIR & \scriptsize DDRM (20) & \scriptsize DDRM (100) & \scriptsize DMPS & \scriptsize D-Turbo-CS & \scriptsize PnP-ADMM
	\end{tabular}
	\caption{Examples of compressive image recovery on FFHQ using randomly row-selected \ac{DCT} sensing matrices. The sampling ratio is $M/N = 0.7$ and we set $\beta = 1$ for \ac{STMP}.}
	\label{fig:representative_stmp_07}
\end{figure*}

\begin{figure*}
	\centering
	\def\arraystretch{0.7}
	\setlength\tabcolsep{0.03cm}
	\begin{tabular}{lccccccccc}
		\multirow{2}{*}{\raisebox{-0.5cm}[0pt][0pt]{\rotatebox{90}{\scriptsize $\delta_0 = 0.05$}}} &
		\includegraphics[width=\imwidth,height=\imwidth]{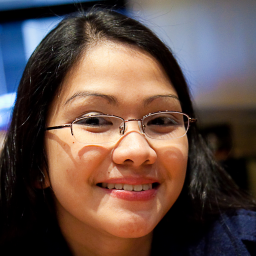}
		& \includegraphics[width=\imwidth,height=\imwidth]{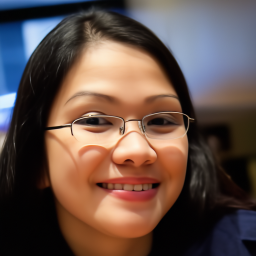}
		& \includegraphics[width=\imwidth,height=\imwidth]{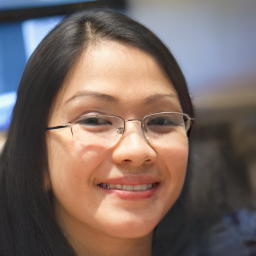}
		& \includegraphics[width=\imwidth,height=\imwidth]{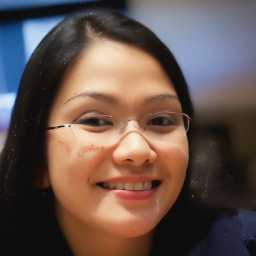}
		& \includegraphics[width=\imwidth,height=\imwidth]{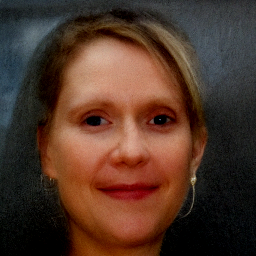}
		& \includegraphics[width=\imwidth,height=\imwidth]{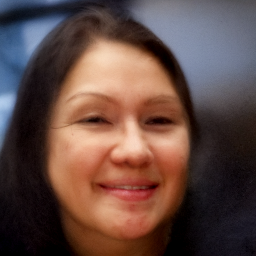}
		& \includegraphics[width=\imwidth,height=\imwidth]{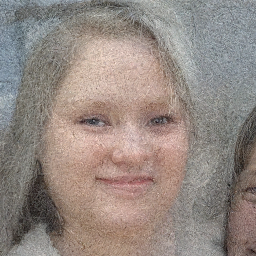}
		& \includegraphics[width=\imwidth,height=\imwidth]{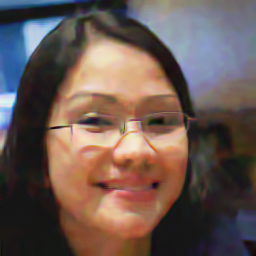}
		& \includegraphics[width=\imwidth,height=\imwidth]{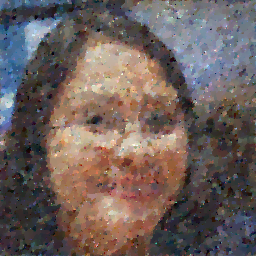}  \\
		&
		\includegraphics[width=\imwidth,height=\imwidth]{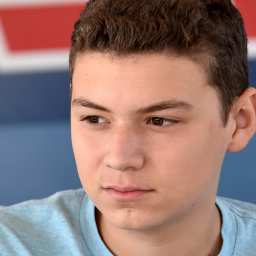}
		& \includegraphics[width=\imwidth,height=\imwidth]{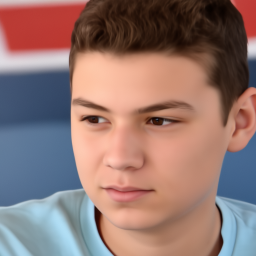}
		& \includegraphics[width=\imwidth,height=\imwidth]{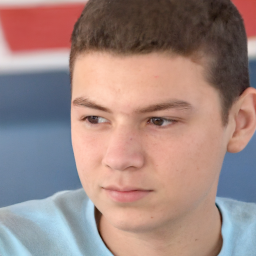}
		& \includegraphics[width=\imwidth,height=\imwidth]{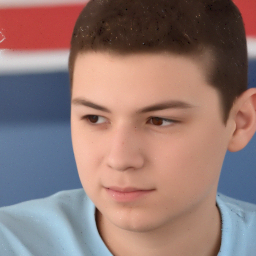}
		& \includegraphics[width=\imwidth,height=\imwidth]{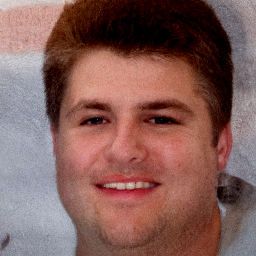}
		& \includegraphics[width=\imwidth,height=\imwidth]{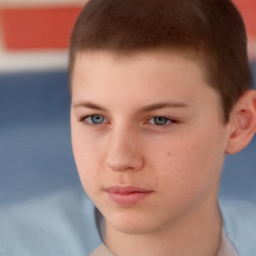}
		& \includegraphics[width=\imwidth,height=\imwidth]{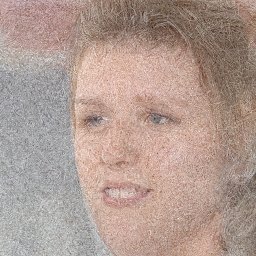}
		& \includegraphics[width=\imwidth,height=\imwidth]{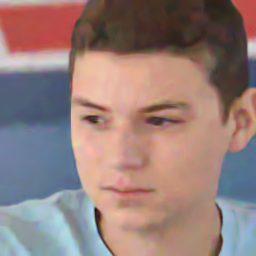}
		& \includegraphics[width=\imwidth,height=\imwidth]{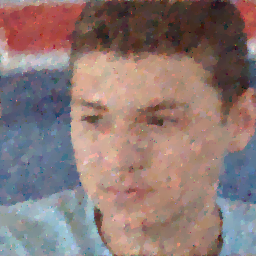} \\
		\multicolumn{10}{c}{\vspace{-.2em}} \\
		\multirow{2}{*}{\raisebox{-0.45cm}[0pt][0pt]{\rotatebox{90}{\scriptsize $\delta_0 = 0.5$}}} &
		\includegraphics[width=\imwidth,height=\imwidth]{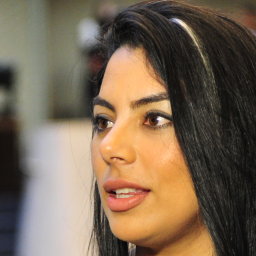}
		& \includegraphics[width=\imwidth,height=\imwidth]{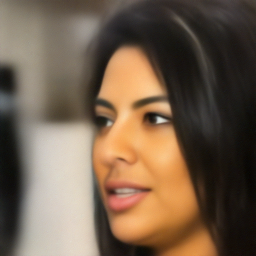}
		& \includegraphics[width=\imwidth,height=\imwidth]{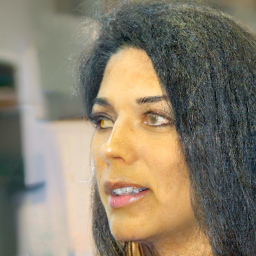}
		& \includegraphics[width=\imwidth,height=\imwidth]{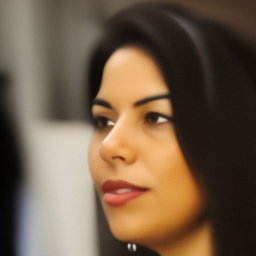}
		& \includegraphics[width=\imwidth,height=\imwidth]{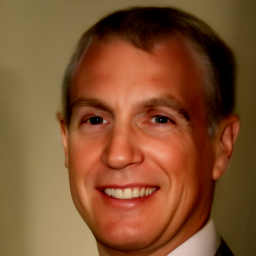}
		& \includegraphics[width=\imwidth,height=\imwidth]{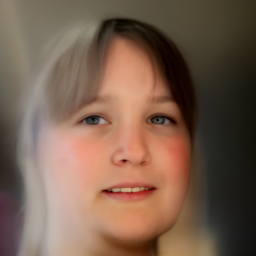}
		& \includegraphics[width=\imwidth,height=\imwidth]{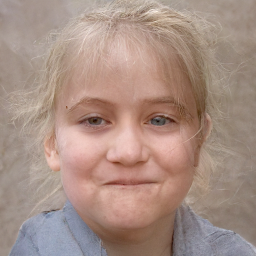}
		& \includegraphics[width=\imwidth,height=\imwidth]{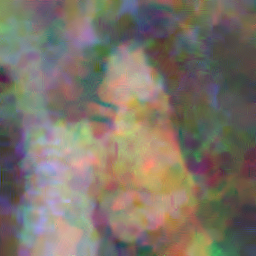}
		& \includegraphics[width=\imwidth,height=\imwidth]{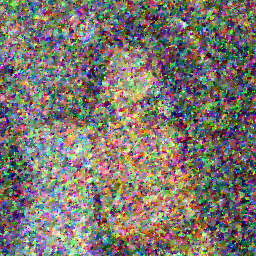} \\
		&
		\includegraphics[width=\imwidth,height=\imwidth]{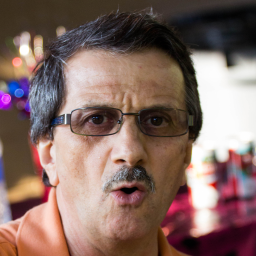}
		& \includegraphics[width=\imwidth,height=\imwidth]{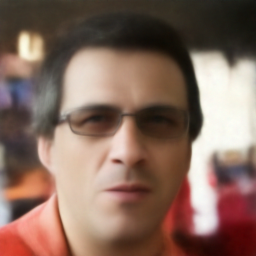}
		& \includegraphics[width=\imwidth,height=\imwidth]{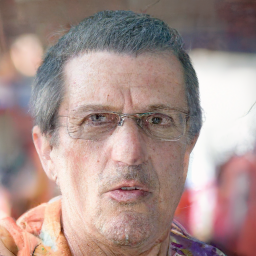}
		& \includegraphics[width=\imwidth,height=\imwidth]{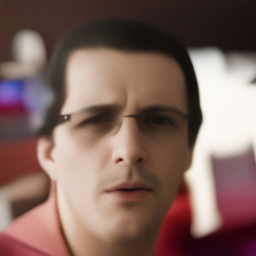}
		& \includegraphics[width=\imwidth,height=\imwidth]{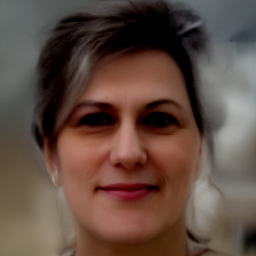}
		& \includegraphics[width=\imwidth,height=\imwidth]{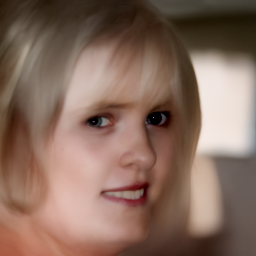}
		& \includegraphics[width=\imwidth,height=\imwidth]{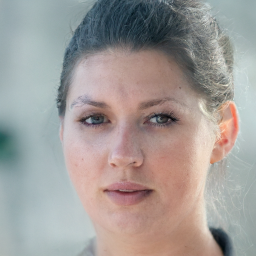}
		& \includegraphics[width=\imwidth,height=\imwidth]{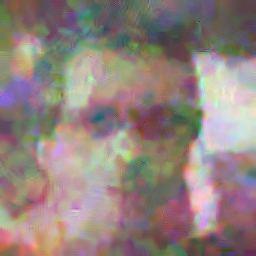}
		& \includegraphics[width=\imwidth,height=\imwidth]{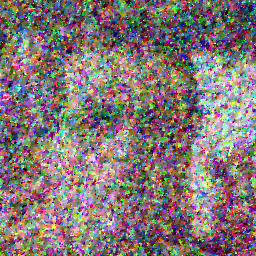}  \\
		& \scriptsize Ground truth & \scriptsize STMP & \scriptsize DPS & \scriptsize DiffPIR & \scriptsize DDRM (20) & \scriptsize DDRM (100) & \scriptsize DMPS & \scriptsize D-Turbo-CS & \scriptsize PnP-ADMM
	\end{tabular}
	\caption{Examples of compressive image recovery on FFHQ using randomly row-selected \ac{DCT} sensing matrices. The sampling ratio is $M/N = 0.1$ and we set $\beta = 0.5$ for \ac{STMP}.}
	\label{fig:representative_stmp_01}
\end{figure*}

\begin{table*}[t]
	\centering
	\caption{Quantitative Results for Compressive Image Recovery on FFHQ Using Randomly Row-selected \ac{DCT} Sensing Matrices \\
		(\textbf{Bold}: Best, \underline{Underline}: Second Best)}
	\vspace{-1em}
	\label{tab:stmp_quantitative_results}
	\begin{center}
		\begin{tabular}{c l p{0.0001\textwidth} cccc p{0.0001\textwidth} cccc}
			\toprule
			\multirow{2}{*}{Sampling ratio} & \multirow{2}{*}{Method} & ~ &  \multicolumn{4}{c}{$\delta_0 = 0.5$} & ~ & \multicolumn{4}{c}{$\delta_0 = 0.05$} \\ 
			\cline{4-7} \cline{9-12}
			& & ~ & PSNR$\uparrow$ & SSIM$\uparrow$ & FID$\downarrow$ & LPIPS$\downarrow$ & ~ & PSNR$\uparrow$ & SSIM$\uparrow$ & FID$\downarrow$ & LPIPS$\downarrow$ \\
			\midrule
			\multirow{8}{*}{$M/N = 0.7$} & STMP (Ours) & ~ & $\mathbf{25.82}$ & $\mathbf{0.7649}$ & $57.76$ & $\mathbf{0.2089}$ & ~ & $\mathbf{34.98}$ & $\mathbf{0.9420}$ & $\underline{23.42}$ & $\underline{0.0437}$ \\
			& DPS & ~ & $17.98$ & $0.5888$ & $\mathbf{39.98}$ & $0.2323$ & ~ & $28.33$ & $0.8596$ & $26.63$ & $0.0857$ \\
			& DiffPIR  & ~ & $24.19$ & $0.7015$ & $86.51$ & $0.2870$ & ~ & $28.05$ & $0.7728$ & $27.00$ & $0.0914$ \\
			& DDRM (20) & ~ &  $24.77$ & $0.7252$ & $58.90$ & $0.2253$ & ~ & $29.10$ & $0.8552$ & $37.92$ & $0.1192$ \\
			& DDRM (100) & ~ & $\underline{25.01}$ & $\underline{0.7339}$ & $57.00$ & $\underline{0.2184}$ & ~ &  $\underline{31.79}$ & $\underline{0.9108}$ & $25.42$ & $0.0636$ \\
			& DMPS & ~ & $18.68$ & $0.5878$ & $\underline{48.32}$ & $0.2383$ & ~ & $30.14$ & $0.8878$ & $\mathbf{13.78}$ & $\mathbf{0.0341}$ \\
			& D-Turbo-CS & ~ & $19.28$  & $0.5390$  & $146.73$ & $0.5131$ & ~ & $28.17$ & $0.8715$ & $37.09$ & $0.1141$ \\
			& PnP-ADMM  & ~ & $8.68$ & $0.0400$ & $341.78$ & $1.4445$ & ~ & $29.15$ & $0.7314$ & $43.26$ & $0.1735$ \\
			\midrule
			\multirow{8}{*}{$M/N = 0.4$} & STMP (Ours) & ~ & $\mathbf{24.73}$ & $\mathbf{0.7320}$ & $61.22$ & $\mathbf{0.2441}$ & ~ & $\mathbf{33.49}$ & $\mathbf{0.9249}$ & $31.35$ & $\mathbf{0.0633}$ \\
			& DPS & ~ & $16.75$ & $0.5466$ & $\mathbf{44.89}$ & $0.2632$ & ~ & $26.66$ & $0.8273$ & $\underline{26.46}$ & $0.0971$ \\
			& DiffPIR  & ~ & $23.18$ & $0.6461$ & $\underline{59.82}$ & $\underline{0.2490}$ & ~ & $28.23$ & $0.8100$ & $\mathbf{24.22}$ & $\underline{0.0667}$ \\
			& DDRM (20) & ~ & $22.37$ & $0.6527$ & $68.61$ & $0.2884$ & ~ & $25.62$ & $0.7673$ & $42.79$ & $0.1798$ \\
			& DDRM (100) & ~ & $\underline{23.58}$ & $\underline{0.6850}$ & $62.56$ & $0.2578$ & ~ & $\underline{29.47}$ & $\underline{0.8599}$ & $30.43$ & $0.0995$ \\
			& DMPS & ~ & $16.66$ & $0.5315$ & $63.92$ & $0.3130$ & ~ &  $25.73$ & $0.7990$ & $26.63$ & $0.0986$ \\
			& D-Turbo-CS & ~ & $18.05$ & $0.4883$ & $204.67$ & $0.5697$ & ~ &  $26.79$ & $0.8378$ & $42.84$ & $0.1524$ \\
			& PnP-ADMM  & ~ & $8.75$ & $0.0381$ & $352.42$ & $1.4230$ & ~ & $28.89$ & $0.7530$ & $56.15$ & $0.1846$  \\
			\midrule
			\multirow{8}{*}{$M/N = 0.1$} & STMP (Ours) & ~ & $\mathbf{22.07}$ & $\mathbf{0.6447}$ & $\underline{68.85}$ & $\mathbf{0.3388}$ & ~ & $\mathbf{29.86}$ & $\mathbf{0.8615}$ & $\underline{46.09}$ & $\mathbf{0.1228}$ \\
			& DPS & ~ & $14.01$ & $0.4447$ & $\mathbf{56.23}$ & $\underline{0.3470}$ & ~ & $22.15$ & $0.7170$ & $\mathbf{29.36}$ & $\underline{0.1446}$ \\
			& DiffPIR & ~ & $\underline{21.07}$ & $\underline{0.6149}$ & $97.38$ & $0.3572$ & ~ & $\underline{25.52}$ & $\underline{0.7539}$ & $56.30$ & $0.1622$ \\
			& DDRM (20) & ~ & $11.82$ & $0.3286$ & $95.36$ & $0.5249$ & ~ &  $12.66$ & $0.2819$ & $122.80$ & $0.6120$ \\
			& DDRM (100) & ~ & $18.48$ & $0.5225$ & $105.22$ & $0.4188$ & ~ &  $21.43$ & $0.6274$ & $68.89$ & $0.3072$ \\
			& DMPS & ~ & $11.31$ & $0.3188$ & $92.56$ & $0.5561$ & ~ & $12.31$ & $0.1787$ & $185.04$ & $0.8939$ \\
			& D-Turbo-CS & ~ & $15.47$ & $0.3913$ & $291.16$ & $0.6741$ & ~ & $23.98$ & $0.7324$ & $63.58$ & $0.2861$ \\
			& PnP-ADMM  & ~ & $11.46$ & $0.0638$ & $374.52$ & $1.0889$ & ~ & $19.16$ & $0.4646$ & $310.09$ & $0.5784$  \\
			\bottomrule
		\end{tabular}
	\end{center}
\end{table*}

\subsection{Results on Compressive Image Recovery}
For the experiments on the FFHQ dataset, we compare the proposed \ac{STMP} algorithm against both state-of-the-art score-based posterior sampling techniques and traditional \ac{PnP} methods.
The baselines include:
\begin{itemize}
	\item \textbf{DPS~\cite{dps}:} We follow the original paper and use $1,000$ reverse-diffusion steps, with the parameter $\zeta$ set to $1$. 
	\item \textbf{DiffPIR~\cite{diffpir}:} We follow the original paper and use $100$ reverse-diffusion steps, with the parameters $(\lambda, \zeta)$ set to $(5, 1)$.
	\item \textbf{DDRM~\cite{kawar2022denoising}:} We use both $20$ and $100$ reverse-diffusion steps. The parameters $(\eta, \eta_b)$ are set to $(0.85, 1)$.
	\item \textbf{DMPS~\cite{dmps}:} We follow the original paper and use $1,000$ reverse-diffusion steps, with the parameter $\lambda$ set to $1.75$.
	\item \textbf{D-Turbo-CS~\cite{xue2017access}:} We choose the BM3D denoiser for D-Turbo-CS. The algorithm is executed for $50$ iterations.
	\item \textbf{PnP-ADMM~\cite{pnp_admm}:}
	We choose the \ac{TV} denoiser for PnP-ADMM and the parameters $(\lambda, \rho, \gamma)$ are set to $(0.01, 1, 1)$. The algorithm is executed for $50$ iterations.
\end{itemize}

\color{black}
For the experiments on the LDCT dataset, we consider the following baselines:
\begin{itemize}
	\item \textbf{\Ac{Score-MI}~\cite{score_mri}:} We follow the original implementation and use the \ac{PC} sampler.
	Each of the $1,000$ reverse-diffusion steps includes one predictor and one corrector update, requiring a total of $2,000$ sequential \ac{NFEs}.
	
	\item \textbf{\Ac{Score-ALD}~\cite{score_ald2021nips}:} \Ac{Score-ALD} uses an ALD sampler with $1,000$ steps, where each step approximates the likelihood score using a matched-filter-type data-consistency term.

	\item \textbf{Score-SDE~\cite{score_sde}:} Score-SDE is based on a crude approximation to the conditional score function.
	It adopts the same \ac{PC} sampler as \ac{Score-MI}, requiring $2{,}000$ sequential \ac{NFEs}.
	
	\item \textbf{FISTA-TV~\cite{fista_tv2009}:} FISTA-TV is a fast iterative shrinkage-thresholding algorithm (FISTA) for solving linear inverse problems in image processing.
	It adopts a \ac{TV} term as the regularization in the optimization procedure.
	We run $300$ iterations for reconstructing each \ac{CT} image with regularization parameter $0.001$.
	
	\item \textbf{FISTA-wavelet:} FISTA-wavelet follows the same configuration as FISTA-TV, except that the \ac{TV} regularizer is replaced by a wavelet-domain sparsity prior.
\end{itemize}
\color{black}

Fig.~\ref{fig:stmp_convergence} shows the \ac{NMSE} trajectory of \ac{STMP} over iterations on the FFHQ dataset.
Overall, the proposed algorithm exhibits fast convergence across a wide range of sampling ratios and noise levels.
As expected, the number of iterations required for convergence increases as the sampling ratio $M/N$ decreases, since recovering an image from fewer measurements becomes more ill-posed.
In addition, the algorithm converges slightly more slowly when the measurement noise is small.
\color{black}
Regarding the \ac{SE} characterization, we observe that the \ac{SE} equations provide accurate predictions of the converged \ac{NMSE} achieved by \ac{STMP}.
However, before convergence, a noticeable mismatch arises between the \ac{SE} prediction and the actual \ac{STMP} trajectory.
As discussed in Remark~\ref{remk:empirical_gaussian}, this mismatch is mainly due to the gap between the empirical-Gaussian residual characterization in rigorous \ac{SE} analysis and the stronger i.i.d. Gaussian model used to construct the practical denoiser transfer function.
Moreover, damping may further drive the denoiser input away from the ideal i.i.d. Gaussian model,
since it forms each message as a weighted average of the current extrinsic estimate and its previous value, thereby introducing temporal correlation between consecutive messages.
The transient fluctuations observed in the first few \ac{STMP} iterations can be attributed to the same factors.
\color{black}

\newcommand{\imwidthlarge}{2.4cm}
\begin{figure*}
	\color{black}
	\centering
	\def\arraystretch{0.7}
	\setlength\tabcolsep{0.03cm}
	\begin{tabular}{lccccccc}
		& \includegraphics[width=\imwidthlarge]{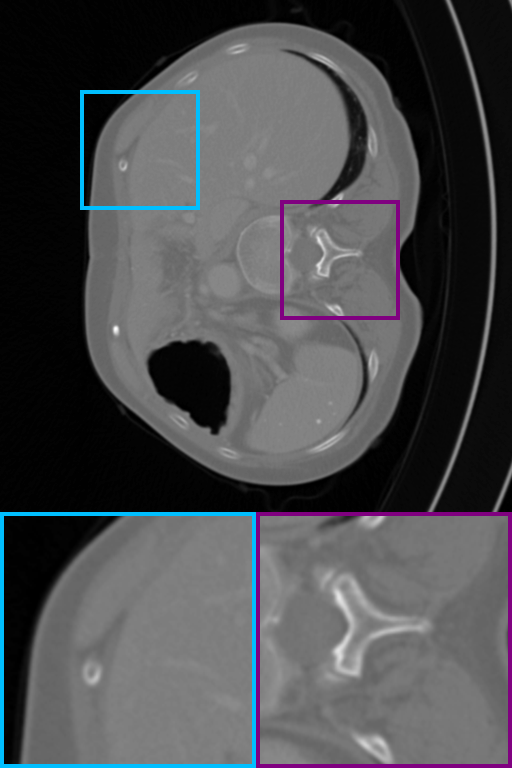}
		& \includegraphics[width=\imwidthlarge]{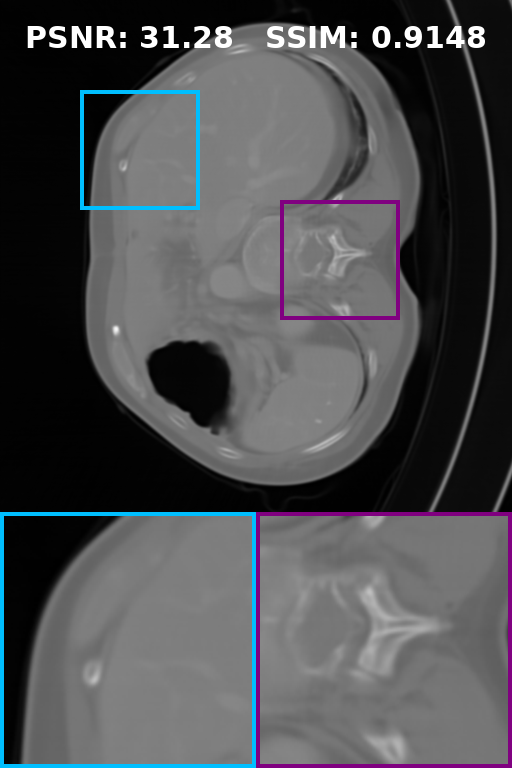}
		& \includegraphics[width=\imwidthlarge]{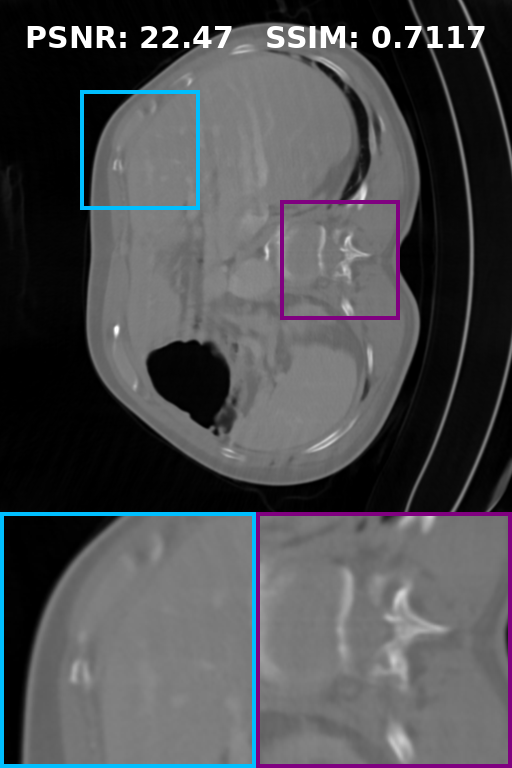}
		& \includegraphics[width=\imwidthlarge]{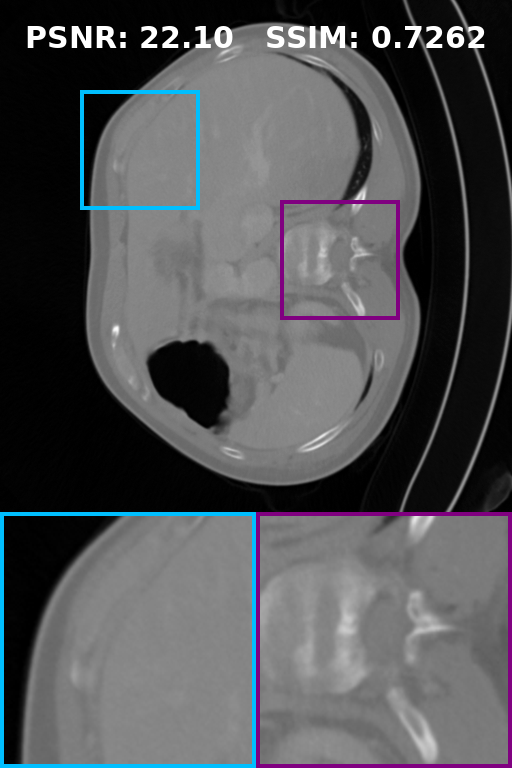}
		& \includegraphics[width=\imwidthlarge]{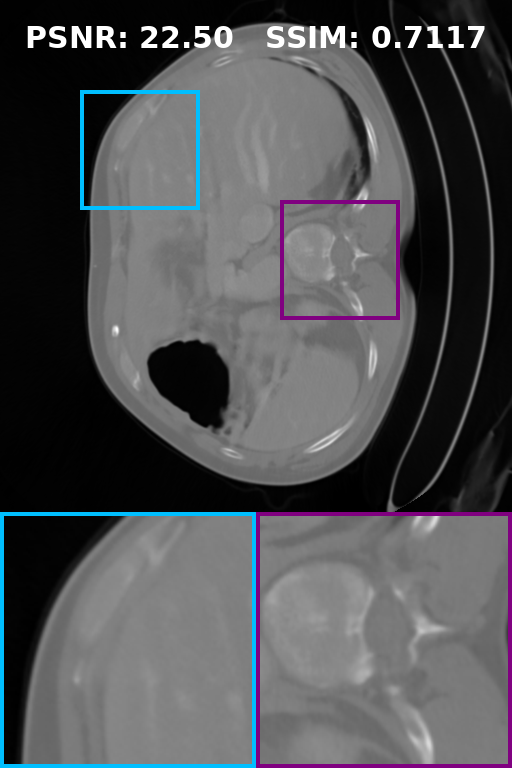}
		& \includegraphics[width=\imwidthlarge]{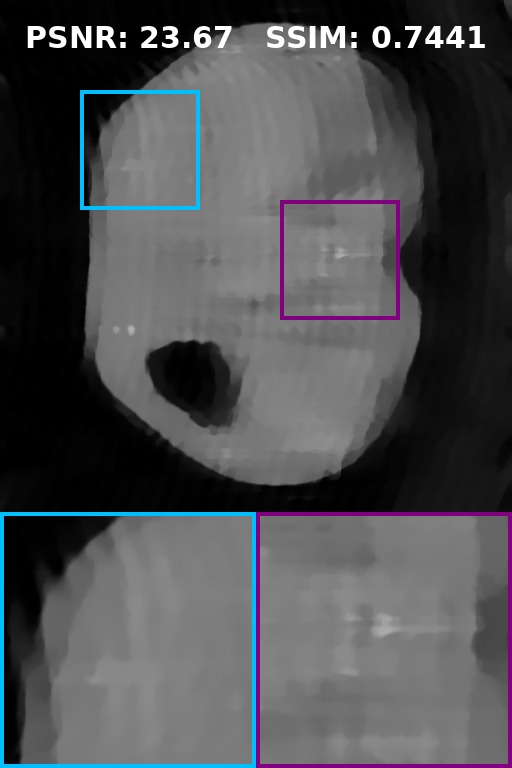}
		& \includegraphics[width=\imwidthlarge]{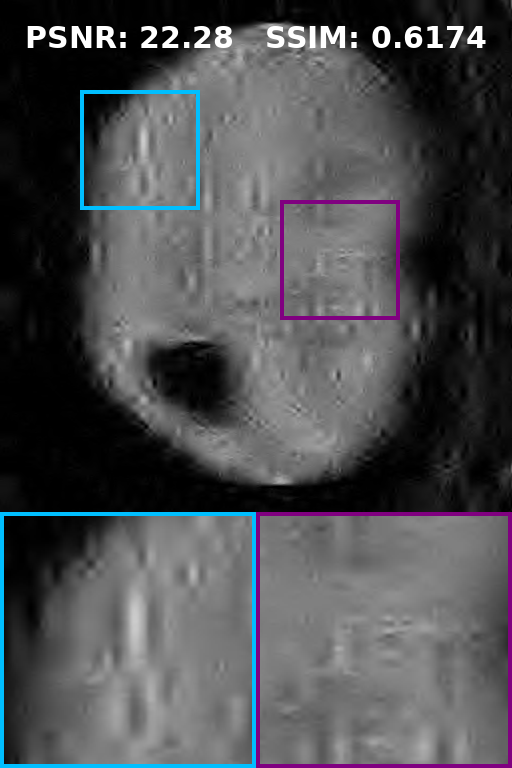}  \\
		&
		\includegraphics[width=\imwidthlarge]{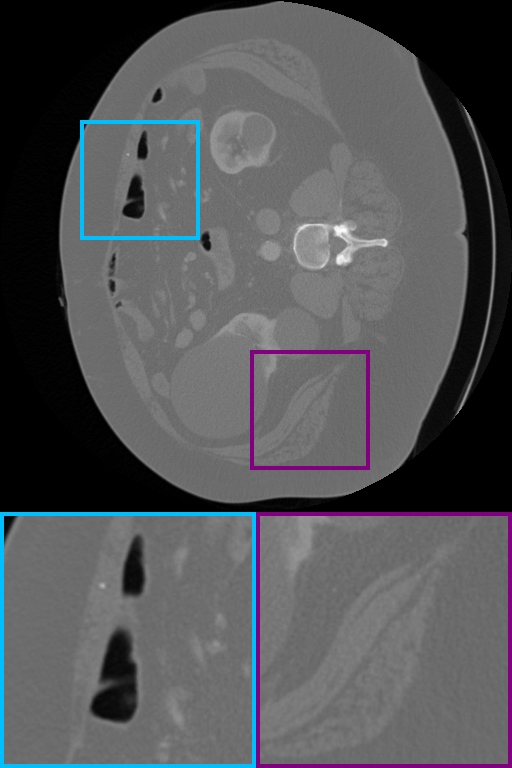}
& \includegraphics[width=\imwidthlarge]{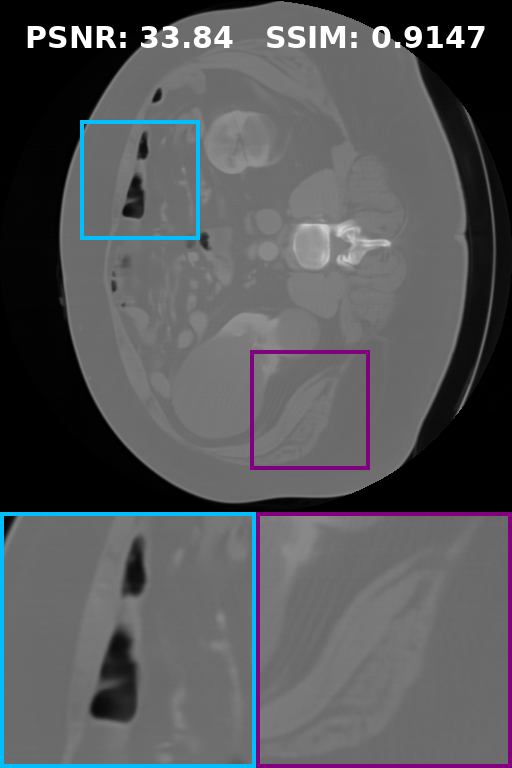}
& \includegraphics[width=\imwidthlarge]{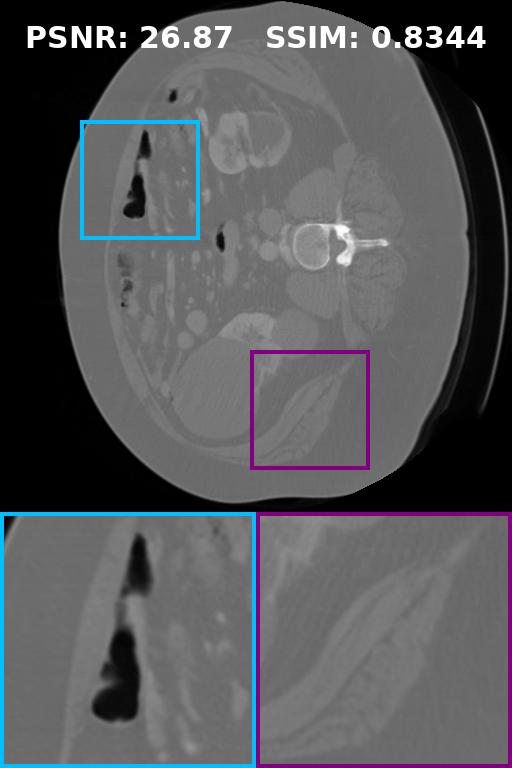}
& \includegraphics[width=\imwidthlarge]{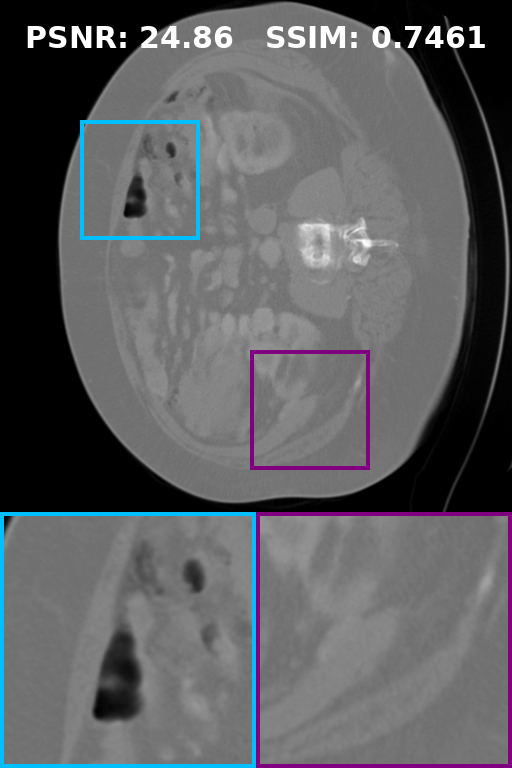}
& \includegraphics[width=\imwidthlarge]{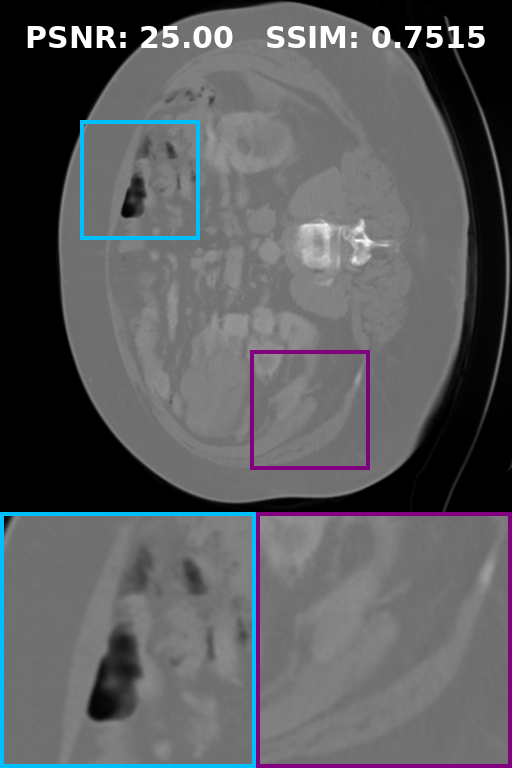}
& \includegraphics[width=\imwidthlarge]{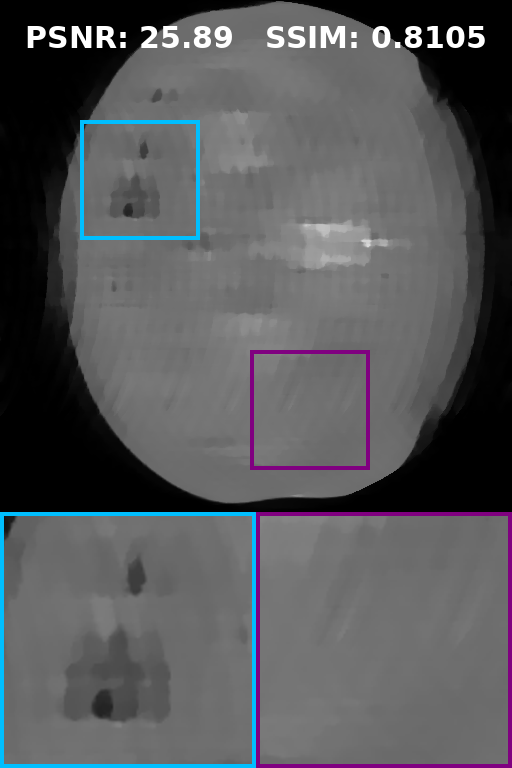}
& \includegraphics[width=\imwidthlarge]{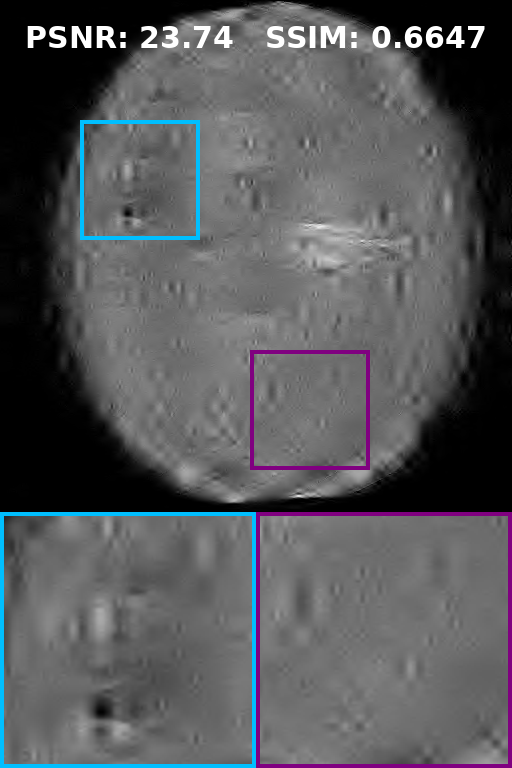}  \\
		& \scriptsize Ground truth & \scriptsize STMP & \scriptsize Score-MI & \scriptsize Score-ALD & \scriptsize Score-SDE & \scriptsize FISTA-TV & \scriptsize FISTA-wavelet
	\end{tabular}
	\caption{Examples of compressive image recovery on LDCT using \Ac{DFT} sensing matrices with equispaced Cartesian masks.
	The sampling ratio is $M/N = 1/32$, with noiseless observations.
	We set $\beta = 0.2$ for \ac{STMP}.}
	\label{fig:tsp_major_mri1}
\end{figure*}

\begin{figure*}
	\color{black}
	\centering
	\def\arraystretch{0.7}
	\setlength\tabcolsep{0.03cm}
	\begin{tabular}{lccccccc}
		& \includegraphics[width=\imwidthlarge]{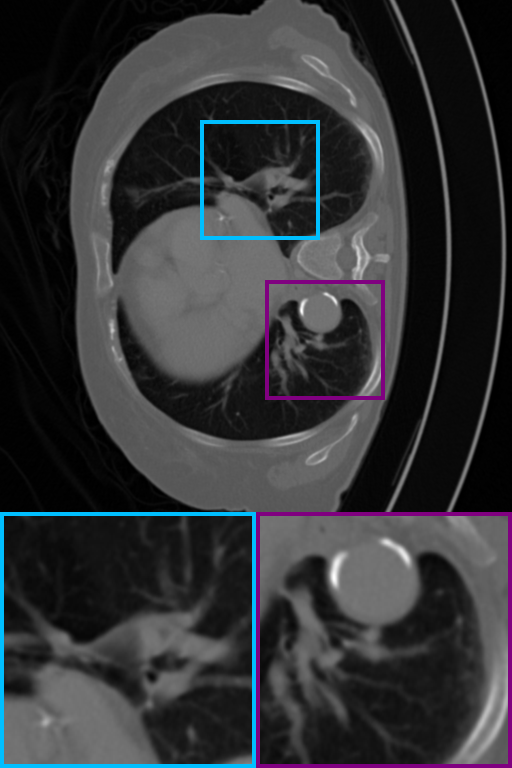}
		& \includegraphics[width=\imwidthlarge]{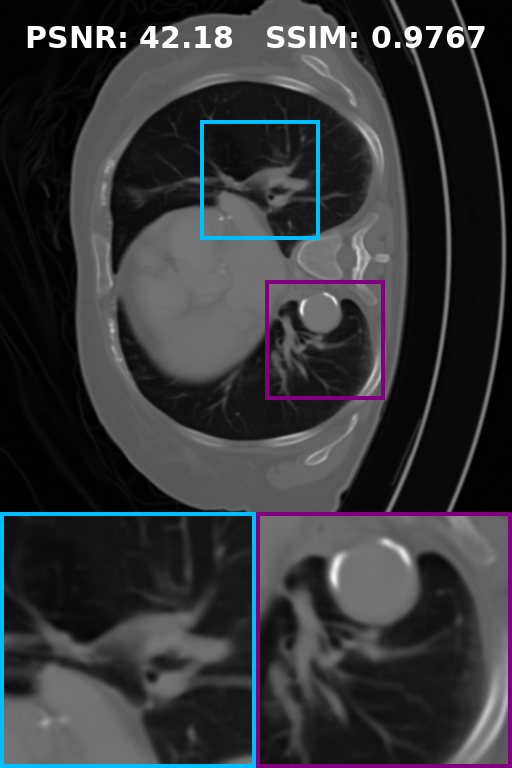}
		& \includegraphics[width=\imwidthlarge]{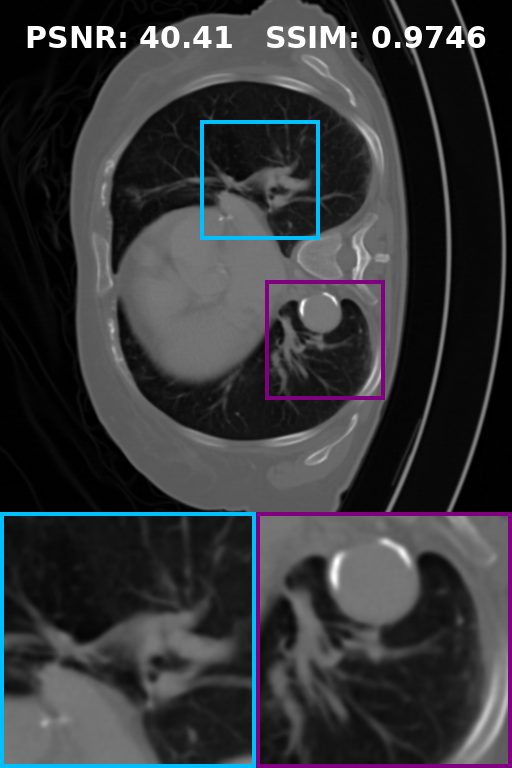}
		& \includegraphics[width=\imwidthlarge]{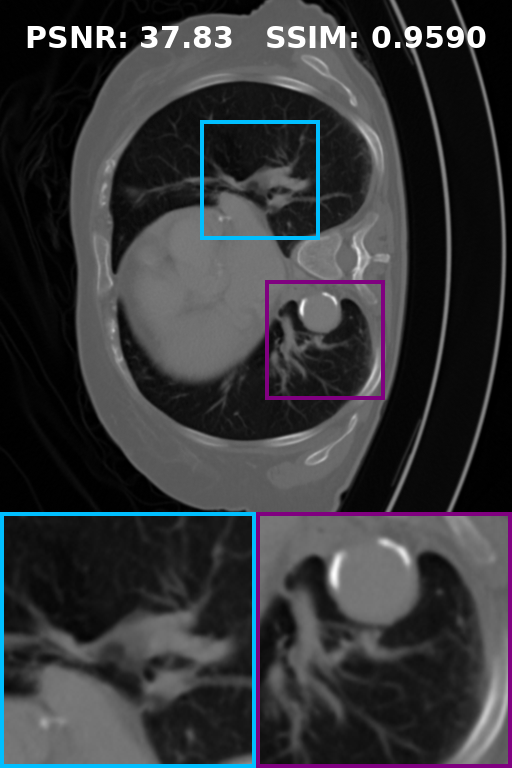}
		& \includegraphics[width=\imwidthlarge]{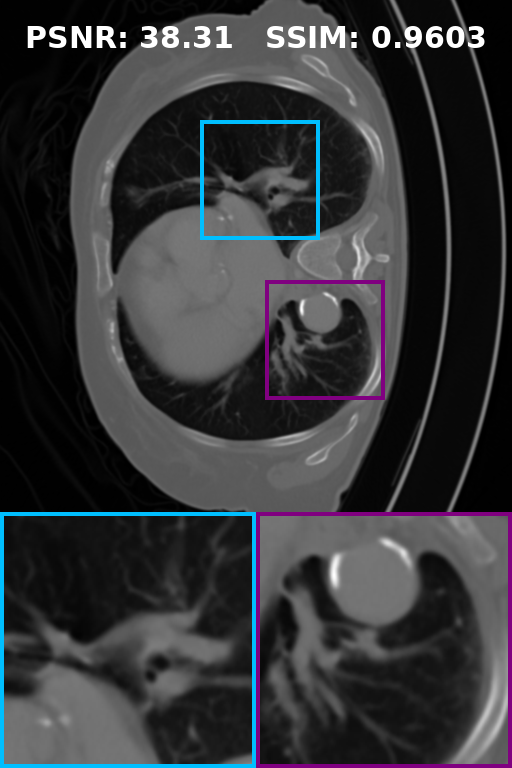}
		& \includegraphics[width=\imwidthlarge]{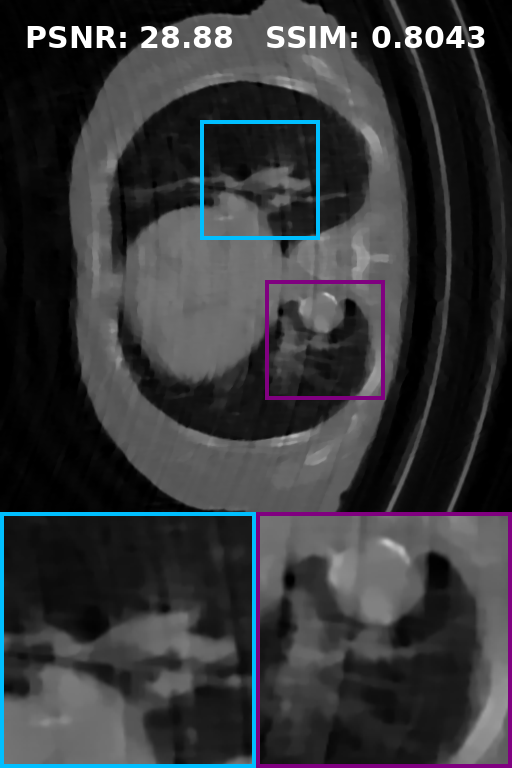}
		& \includegraphics[width=\imwidthlarge]{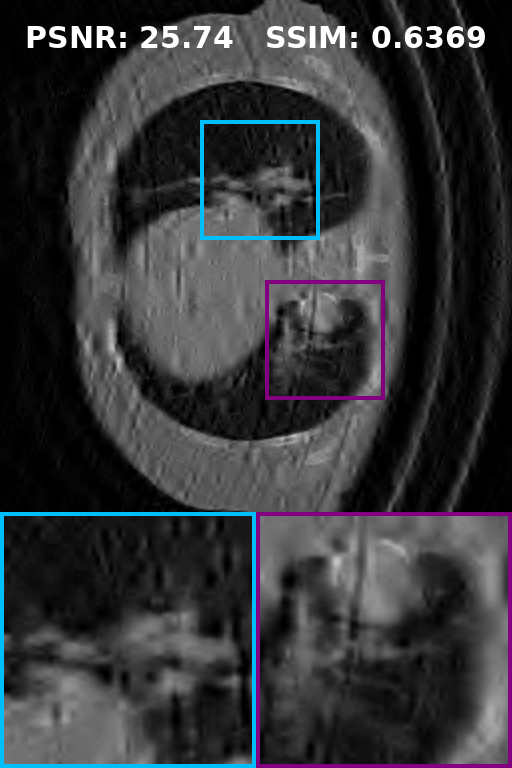}  \\
		&
		\includegraphics[width=\imwidthlarge]{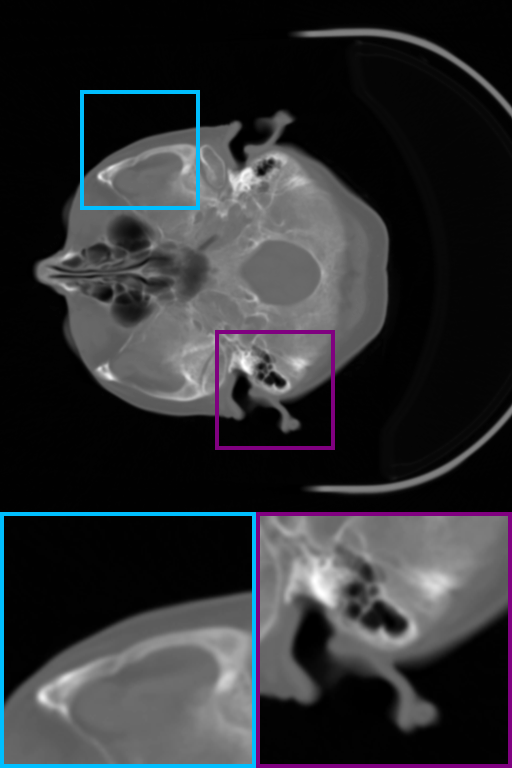}
	& \includegraphics[width=\imwidthlarge]{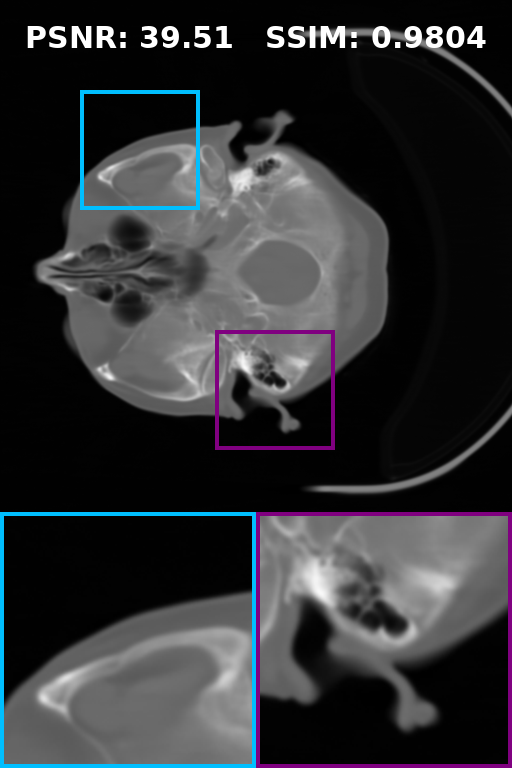}
	& \includegraphics[width=\imwidthlarge]{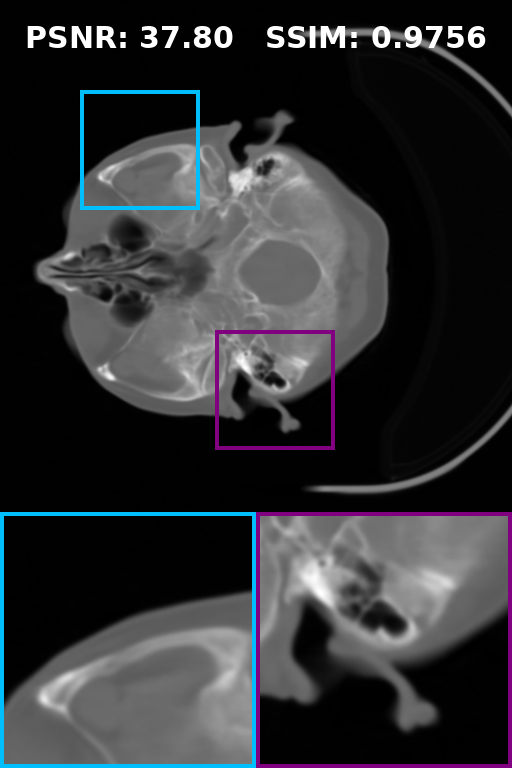}
	& \includegraphics[width=\imwidthlarge]{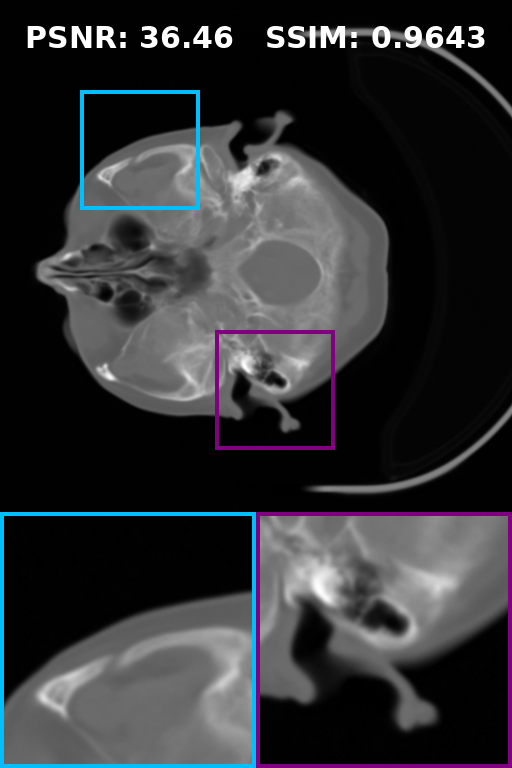}
	& \includegraphics[width=\imwidthlarge]{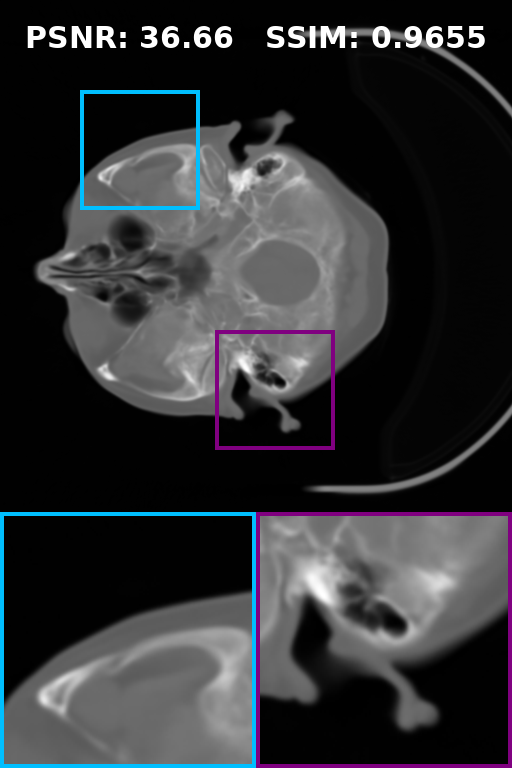}
	& \includegraphics[width=\imwidthlarge]{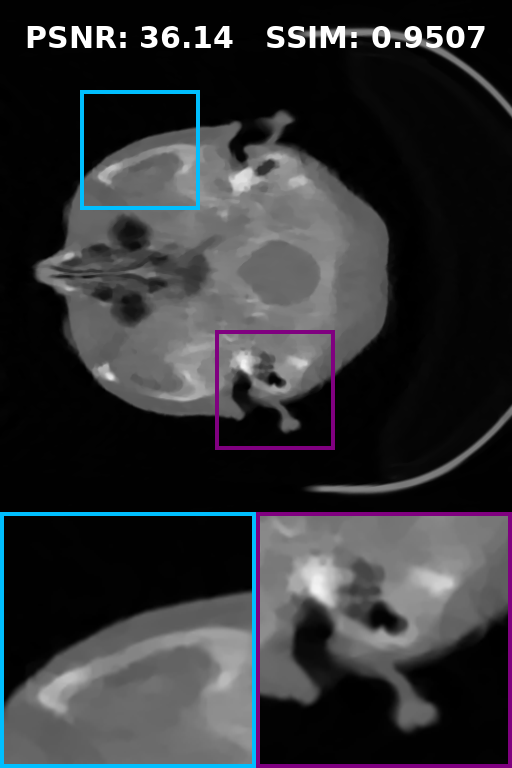}
	& \includegraphics[width=\imwidthlarge]{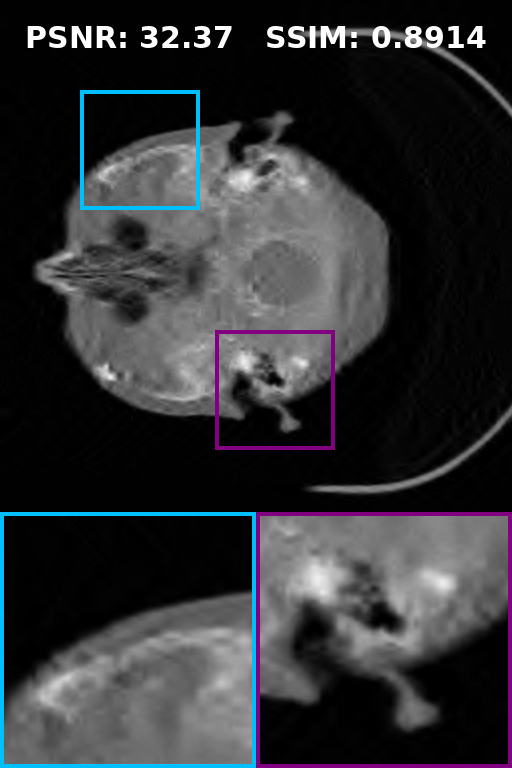}  \\
		& \scriptsize Ground truth & \scriptsize STMP & \scriptsize Score-MI & \scriptsize Score-ALD & \scriptsize Score-SDE & \scriptsize FISTA-TV & \scriptsize FISTA-wavelet
	\end{tabular}
	\caption{Examples of compressive image recovery on LDCT using \Ac{DFT} sensing matrices with equispaced Cartesian masks.
		The sampling ratio is $M/N = 1/8$, with noiseless observations.
		We set $\beta = 0.2$ for \ac{STMP}.}
	\label{fig:tsp_major_mri2}
\end{figure*}

\begin{table*}[t]
	\color{black}
	\centering
	\caption{Quantitative Results for Compressive Image Recovery on LDCT Using \Ac{DFT} Sensing Matrices with Equispaced Cartesian Masks \\
		(\textbf{Bold}: Best, \underline{Underline}: Second Best; $\delta_0 = 0$)}
	\vspace{-1em}
	\label{tab:tsp_major_mri}
	\begin{center}
		\begin{tabular}{l p{0.0001\textwidth} cc p{0.0001\textwidth} cc p{0.0001\textwidth} cc p{0.0001\textwidth} cc}
			\toprule
			\multirow{2}{*}{Method} & ~ &  \multicolumn{2}{c}{$M/N=1/32$} &~ & \multicolumn{2}{c}{$M/N=1/16$} & ~ & \multicolumn{2}{c}{$M/N=1/8$} & ~ & \multicolumn{2}{c}{$M/N=1/4$}  \\ 
			\cline{3-4} \cline{6-7} \cline{9-10} \cline{12-13}
			& ~ & PSNR$\uparrow$ & SSIM$\uparrow$ & ~ & PSNR$\uparrow$ & SSIM$\uparrow$  & ~ & PSNR$\uparrow$ & SSIM$\uparrow$ &~ & PSNR$\uparrow$ & SSIM$\uparrow$  \\
			\midrule
			STMP (Ours) & ~ &  $\mathbf{30.97}$ & $\mathbf{0.8581}$ & ~ & $\mathbf{37.09}$ & $\mathbf{0.9152}$ & ~ & $\mathbf{41.48}$ & $\mathbf{0.9534}$ & ~ & $\mathbf{44.15}$ & $\underline{0.9709}$  \\
			Score-MI & ~ & $\underline{30.48}$ & $\underline{0.8574}$ & ~ & $\underline{34.19}$ & $\underline{0.9018}$ & ~ & $\underline{39.98}$ & $\underline{0.9409}$ & ~ & $\underline{43.89}$ & $\mathbf{0.9778}$ \\
			Score-ALD & ~ & $26.58$ & $0.7831$ & ~ & $30.25$ & $0.8578$ & ~ & $35.89$ & $0.9337$ & ~ & $40.63$ & $0.9638$ \\
			Score-SDE & ~ & $26.75$ & $0.7870$ & ~ & $30.53$ & $0.8633$ & ~ & $36.33$ & $0.9382$ & ~ & $41.13$ & $0.9668$ \\
			FISTA-TV & ~ & $23.58$ & $0.6873$ & ~ & $26.08$ & $0.7337$ & ~ & $32.00$ & $0.8431$ & ~ & $40.76$ & $0.9337$ \\
			FISTA-wavelet & ~ & $22.61$ & $0.5934$ & ~ & $24.42$ & $0.6252$ & ~ & $27.95$ & $0.7127$ & ~ & $35.66$ & $0.8783$ \\
			\bottomrule
		\end{tabular}
	\end{center}
\end{table*}

Figs.~\ref{fig:representative_stmp_07} and~\ref{fig:representative_stmp_01} present representative qualitative results on FFHQ at sampling ratios $M/N = 0.7$ and $M/N = 0.1$, respectively.
When the sampling ratio is high, all score-based baselines, including \ac{STMP}, DPS, DiffPIR, DDRM, and DMPS, produce visually high-quality reconstructions that preserve facial geometry and texture details.
In contrast, the conventional \ac{PnP} approaches D-Turbo-CS and PnP-ADMM, which rely on generic denoisers rather than learned score priors, struggle to produce satisfactory outputs and exhibit significant artifacts.
When the sampling ratio is low, the differences between the methods become more pronounced. 
DDRM often produces samples that are inconsistent with the ground-truth measurement, yielding images that are visually plausible but unrelated to the underlying subject.
DPS outperforms DDRM but still fails to remain faithful to the observations when the noise level is high.
For instance, in the last row of Fig.~\ref{fig:representative_stmp_01}, DPS reconstructs a man wearing transparent glasses, whereas the ground truth clearly shows sunglasses, indicating a hallucination caused by its stochastic sampling dynamics.
This deviation highlights the superior fidelity of our proposed \ac{STMP} method, which consistently adheres to the measurement across all noise levels and sampling ratios.

In Table \ref{tab:stmp_quantitative_results},
we report the average peak signal-to-noise ratio (PSNR), structural similarity index measure (SSIM), Fréchet inception distance (FID), and learned perceptual image patch similarity (LPIPS) to assess fidelity to the ground-truth image and perceptual reconstruction quality.
Across different sampling ratios and noise levels, the proposed \ac{STMP} algorithm consistently outperforms competing baselines on the majority of these metrics.
In particular, \ac{STMP} achieves the highest PSNR and SSIM in all settings, indicating superior pixel-level accuracy and structural preservation.
At the same time, its lower FID and LPIPS scores demonstrate that \ac{STMP} also produces more realistic and perceptually convincing images.
These results collectively show that \ac{STMP} not only adheres closely to the underlying measurements but also delivers highly competitive perceptual quality.

\begin{figure}
	[t]
	\centering
	\vspace{-.8em}
	\includegraphics[width=.83\columnwidth]{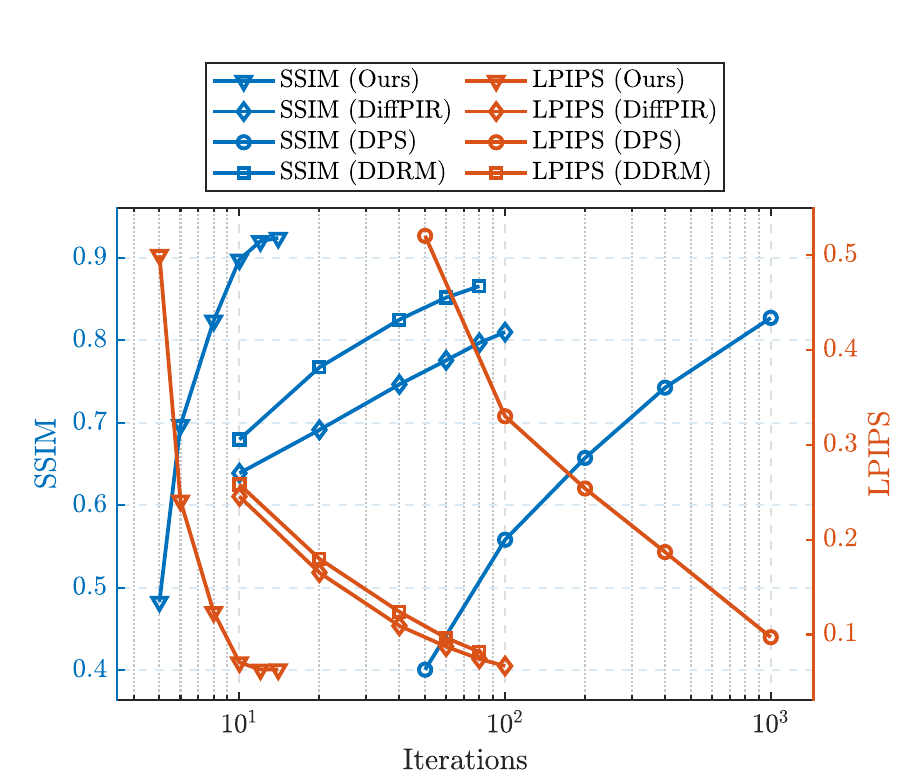}
	\caption{Tradeoff between reconstruction faithfulness / perceptual image quality and computational complexity on FFHQ using randomly row-selected \ac{DCT} sensing matrices. The sampling ratio is $M/N=0.4$ and the noise level is $\delta_0 = 0.05$. We set $\beta = 0.8$ for \ac{STMP}.}
	\label{fig_tradeoff}
\end{figure}

\begin{figure}[t]
	\centering
	\color{black}
	\begin{minipage}{0.493\linewidth}
		\centering
		\includegraphics[width=\linewidth]{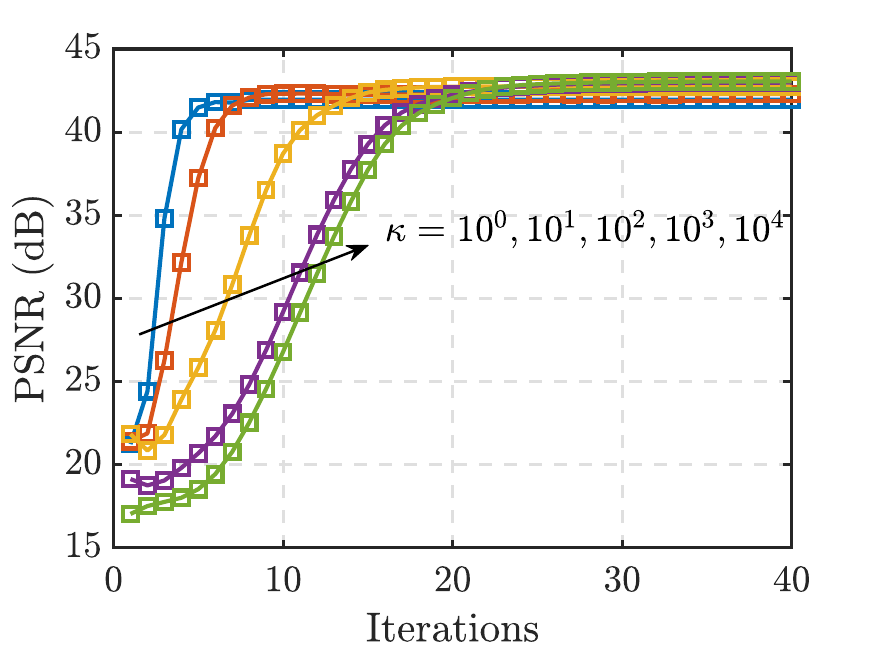}\\
		\footnotesize{(a) PSNR vs. iterations}
	\end{minipage}
	\begin{minipage}{0.493\linewidth}
		\centering		
		\includegraphics[width=\linewidth]{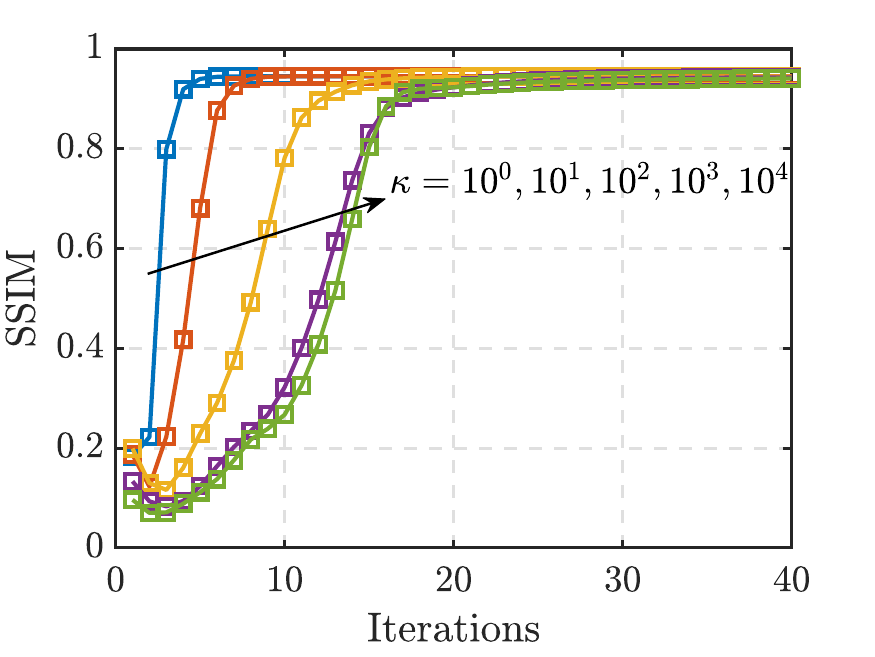}
		\footnotesize{(b) SSIM vs. iterations}
	\end{minipage}
	\caption{\label{fig:major_svd_kappa}
		Effect of the condition number on the convergence speed of \ac{STMP} with general unitarily-invariant sensing matrices constructed via \ac{SVD}. Results are obtained on the LDCT dataset with $M/N=0.7$ and $\delta_0=0.05$.
	}
\end{figure}

\begin{figure}[t]
	\centering
	\color{black}
	\begin{minipage}{0.493\linewidth}
		\centering
		\includegraphics[width=\linewidth]{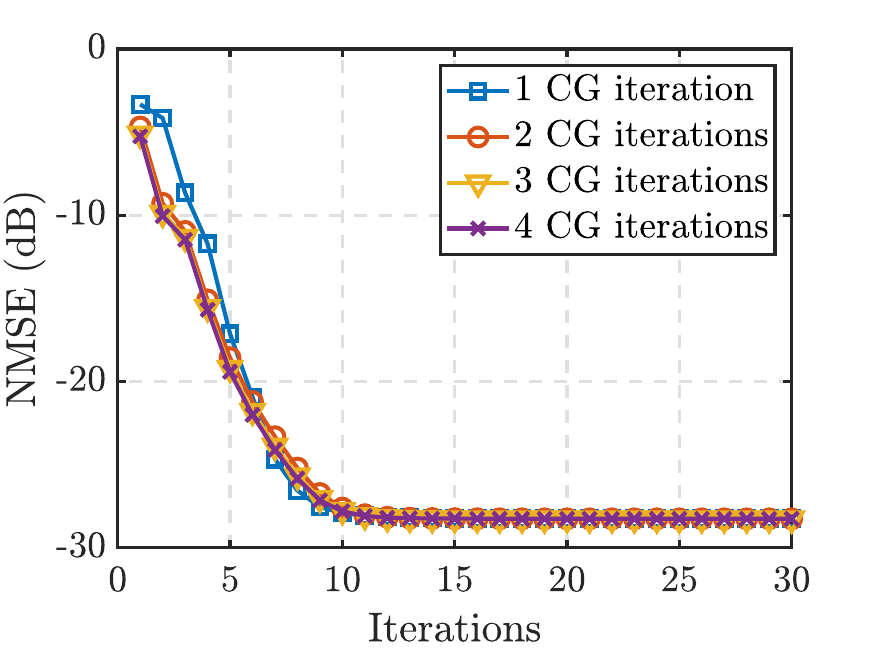}\\
		\footnotesize{(a) One probe vector}
	\end{minipage}
	\begin{minipage}{0.493\linewidth}
		\centering		
		\includegraphics[width=\linewidth]{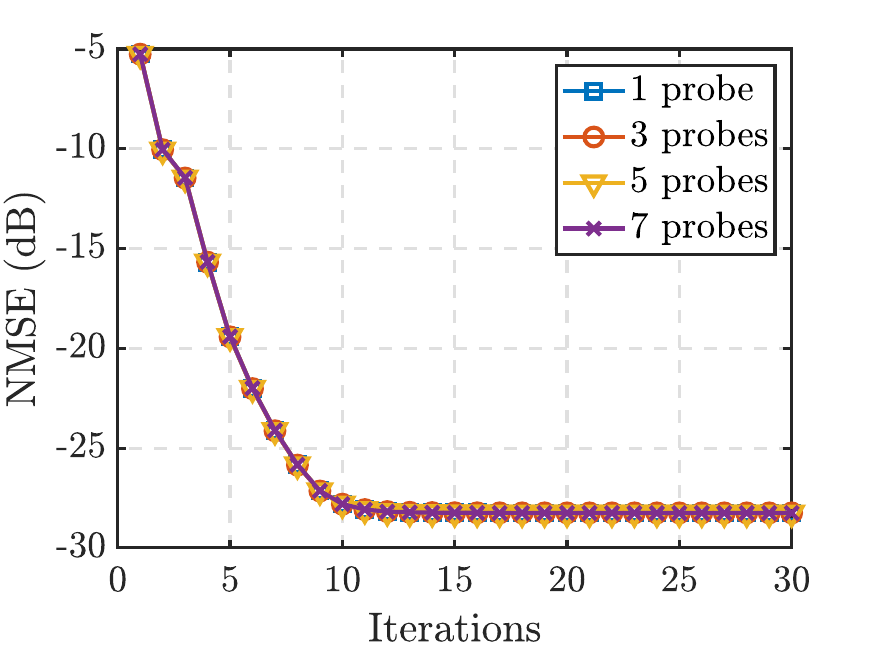}
		\footnotesize{(b) $4$ \ac{CG} iterations}
	\end{minipage}
	\caption{\label{fig:major_cg}
		Convergence behavior of \ac{STMP} with \ac{i.i.d.} Gaussian sensing matrices and \ac{CG}-based \ac{LMMSE} updates. Results are obtained on the FFHQ dataset with $M/N=0.4$, $\delta_0=0.05$, and $\beta=0.6$.}
\end{figure}

Next, we evaluate \ac{STMP} on the LDCT dataset using \ac{DFT} sensing matrices with equispaced Cartesian masks.
Since \ac{Score-MI} and Score-SDE only support noiseless observations, we set $\delta_0=0$ in the experiments reported in Fig.~\ref{fig:tsp_major_mri1}, Fig.~\ref{fig:tsp_major_mri2}, and Table~\ref{tab:tsp_major_mri}.
Overall, \ac{STMP} achieves the best reconstruction quality for almost all sampling ratios and consistently outperforms the optimization-based baselines by a clear margin.
Compared with score-based posterior sampling methods, \ac{STMP} provides comparable or better visual fidelity while preserving fine anatomical structures more faithfully, especially in highly compressed regimes.
Quantitatively, \ac{STMP} attains the highest PSNR across all sampling ratios and the best SSIM in most cases, demonstrating its strong reconstruction accuracy and robustness on the LDCT dataset.

Fig.~\ref{fig_tradeoff} illustrates the tradeoff between reconstruction performance and computational complexity on the FFHQ dataset.
For the baseline algorithms, we generate the tradeoff curves by varying the number of reverse-diffusion steps, whereas for \ac{STMP} we directly report the reconstruction performance at each iteration.
Compared with DiffPIR, DPS, and DDRM, the proposed \ac{STMP} algorithm achieves high SSIM and low LPIPS within only a small number of iterations, demonstrating its superior efficiency.
Regarding per-iteration complexity, the computational cost of these methods is dominated by the \ac{NFEs} of the score network.
In \ac{STMP}, each iteration requires one NFE from the first-order score network and one NFE from the second-order score network. 
However, these two NFEs can be executed in parallel, meaning that the overall time per iteration is comparable to that of DiffPIR and DDRM, both of which involve only a single NFE per step.
In contrast, DPS requires not only a forward pass for each NFE but also backpropagation through the score network.
This makes the iterative steps in DPS even more computationally expensive.

\begin{figure*}[t]
	\centering
	\begin{minipage}{0.325\linewidth}
		\centering
		\includegraphics[width=\linewidth]{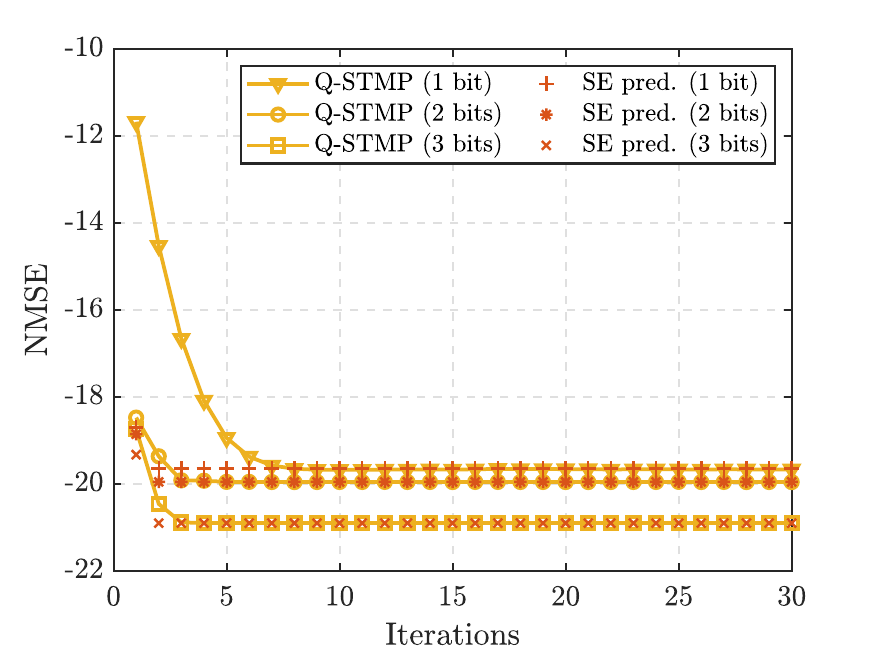}\\
		\footnotesize{(a) $M/N = 0.8$, $\beta = 1$}
	\end{minipage}
	\begin{minipage}{0.325\linewidth}
		\centering		
		\includegraphics[width=\linewidth]{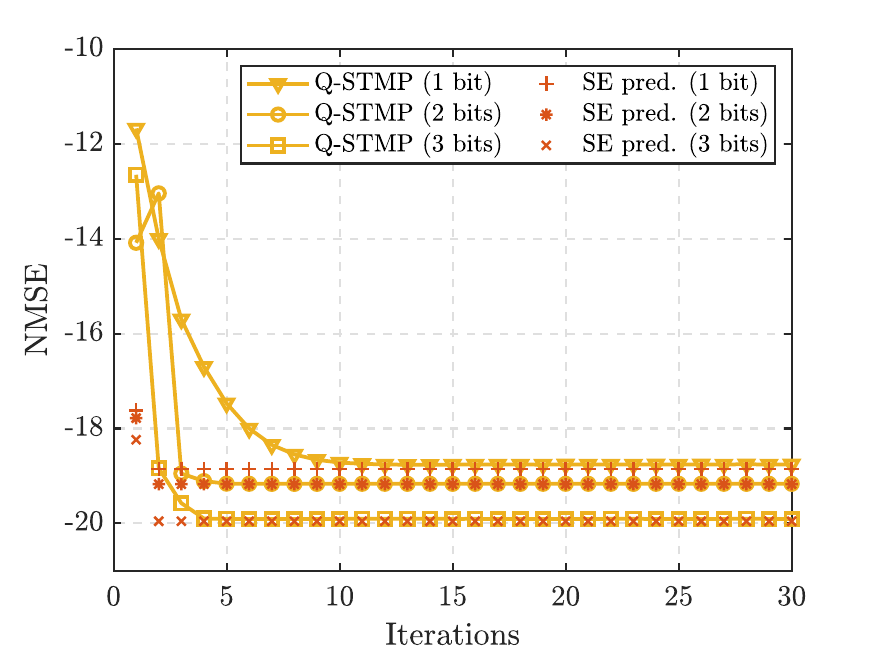}\\
		\footnotesize{(b) $M/N = 0.5$, $\beta = 0.8$}
	\end{minipage}
	\begin{minipage}{0.325\linewidth}
		\centering		
		\includegraphics[width=\linewidth]{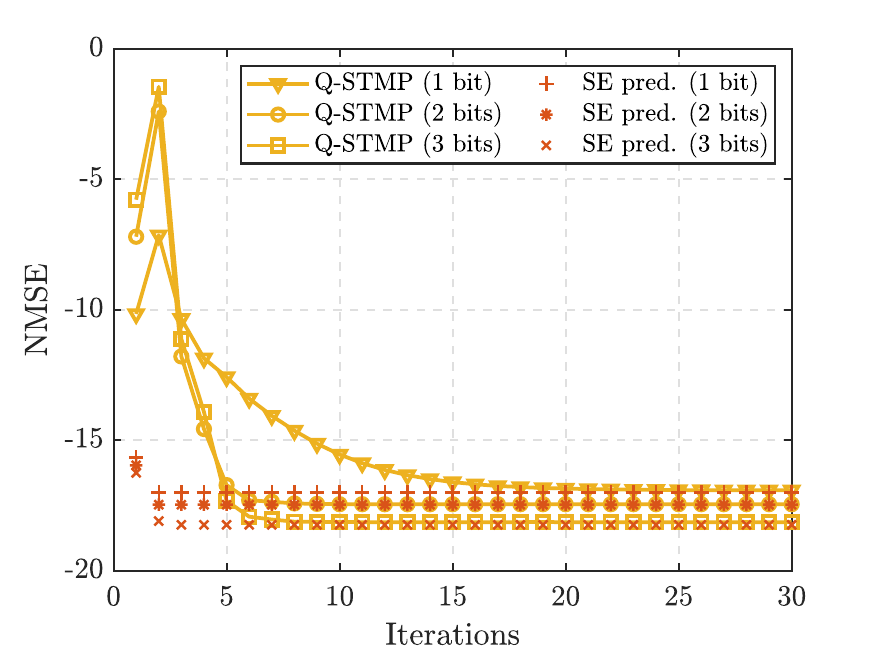}\\
		\footnotesize{(c) $M/N = 0.2$, $\beta = 0.6$}
	\end{minipage}
	\caption{\label{fig:quantized_stmp_convergence}
		Convergence behaviors and \ac{SE} predictions of \ac{Q-STMP} on FFHQ under different sampling ratios and quantization bit depths, where $\delta_0 = 0.5$.
		Randomly row-selected \ac{DCT} sensing matrices are used.}
\end{figure*}

\begin{figure*}
	\centering
	\def\arraystretch{0.7}
	\setlength\tabcolsep{0.03cm}
	\begin{tabular}{lccccclcccc}
		& \multicolumn{4}{c}{\scriptsize $M/N = 0.8$} & \scriptsize ~~~~~~~~ & & \multicolumn{4}{c}{\scriptsize $M/N = 0.5$} \\
		\multirow{2}{*}{\raisebox{-.15cm}[0pt][0pt]{\rotatebox{90}{\scriptsize 3 bits}}} &
		\includegraphics[width=\imwidth,height=\imwidth]{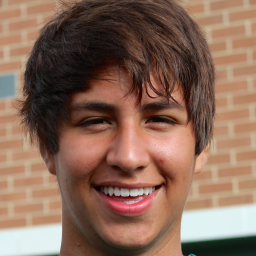}
		& \includegraphics[width=\imwidth,height=\imwidth]{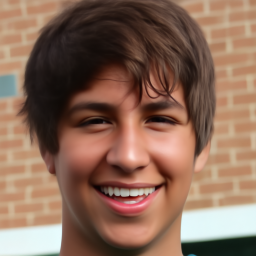}
		& \includegraphics[width=\imwidth,height=\imwidth]{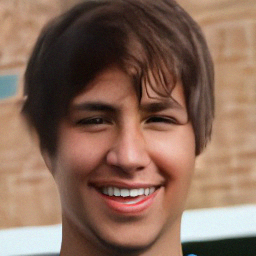}
		& \includegraphics[width=\imwidth,height=\imwidth]{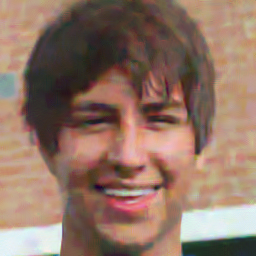}
		& 
		& \multirow{2}{*}{\raisebox{-.15cm}[0pt][0pt]{\rotatebox{90}{\scriptsize 3 bits}}}
		& \includegraphics[width=\imwidth,height=\imwidth]{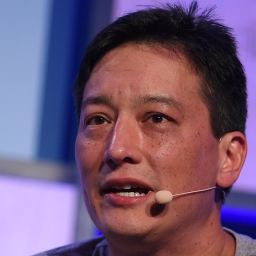}
		& \includegraphics[width=\imwidth,height=\imwidth]{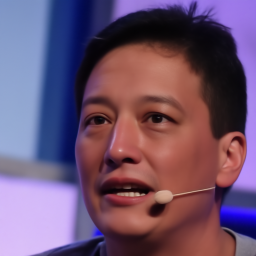}
		& \includegraphics[width=\imwidth,height=\imwidth]{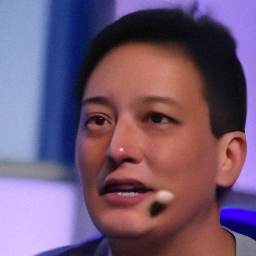}
		& \includegraphics[width=\imwidth,height=\imwidth]{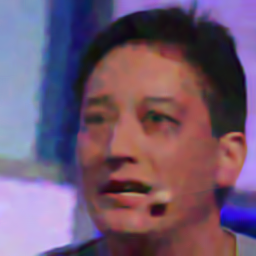}  \\
		& 
		\includegraphics[width=\imwidth,height=\imwidth]{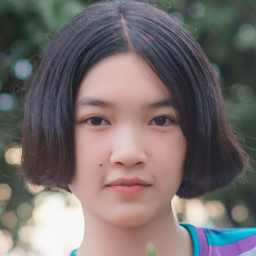}
		& \includegraphics[width=\imwidth,height=\imwidth]{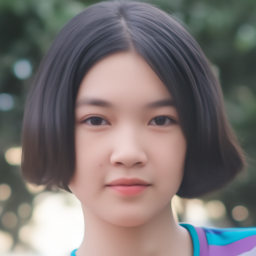}
		& \includegraphics[width=\imwidth,height=\imwidth]{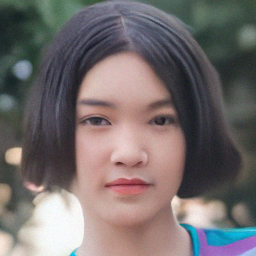}
		& \includegraphics[width=\imwidth,height=\imwidth]{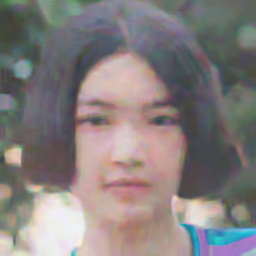}
		& & 
		& \includegraphics[width=\imwidth,height=\imwidth]{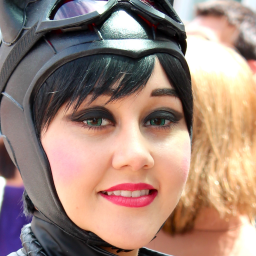}
		& \includegraphics[width=\imwidth,height=\imwidth]{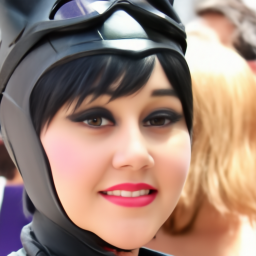}
		& \includegraphics[width=\imwidth,height=\imwidth]{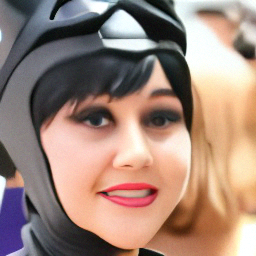}
		& \includegraphics[width=\imwidth,height=\imwidth]{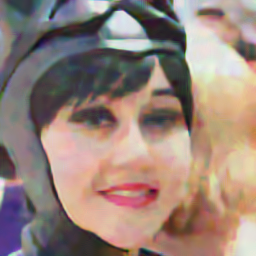} \\
		\multicolumn{11}{c}{\vspace{-.2em}} \\
		\multirow{2}{*}{\raisebox{-.15cm}[0pt][0pt]{\rotatebox{90}{\scriptsize 2 bits}}} &
		\includegraphics[width=\imwidth,height=\imwidth]{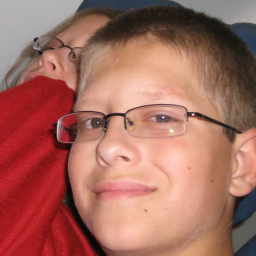}
		& \includegraphics[width=\imwidth,height=\imwidth]{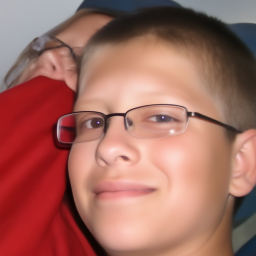}
		& \includegraphics[width=\imwidth,height=\imwidth]{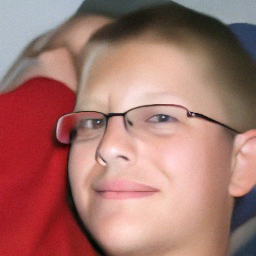}
		& \includegraphics[width=\imwidth,height=\imwidth]{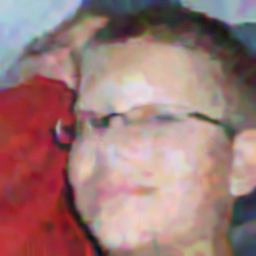}
		& 
		& \multirow{2}{*}{\raisebox{-.15cm}[0pt][0pt]{\rotatebox{90}{\scriptsize 2 bits}}}
		& \includegraphics[width=\imwidth,height=\imwidth]{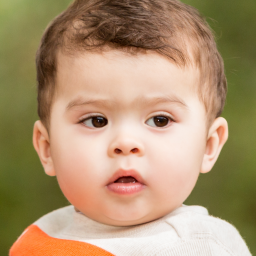}
		& \includegraphics[width=\imwidth,height=\imwidth]{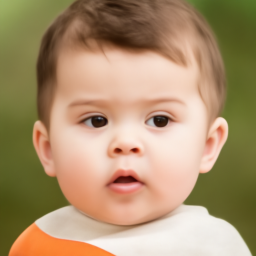}
		& \includegraphics[width=\imwidth,height=\imwidth]{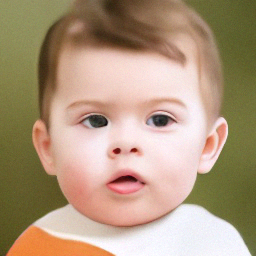}
		& \includegraphics[width=\imwidth,height=\imwidth]{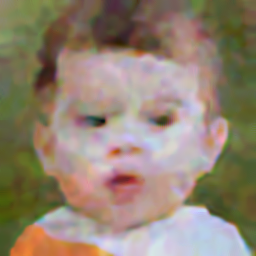} \\
		& 
		\includegraphics[width=\imwidth,height=\imwidth]{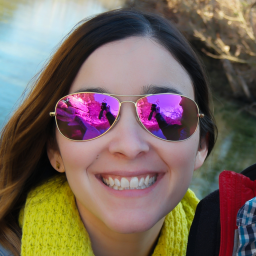}
		& \includegraphics[width=\imwidth,height=\imwidth]{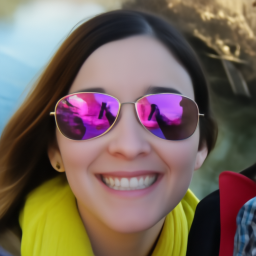}
		& \includegraphics[width=\imwidth,height=\imwidth]{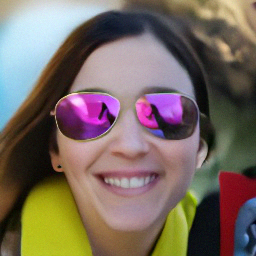}
		& \includegraphics[width=\imwidth,height=\imwidth]{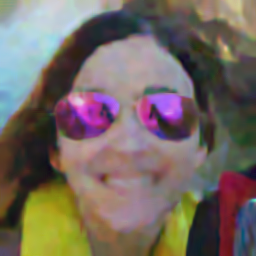}
		& & 
		& \includegraphics[width=\imwidth,height=\imwidth]{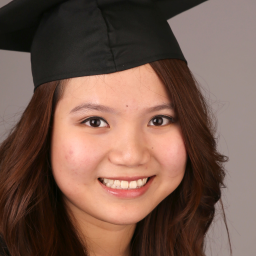}
		& \includegraphics[width=\imwidth,height=\imwidth]{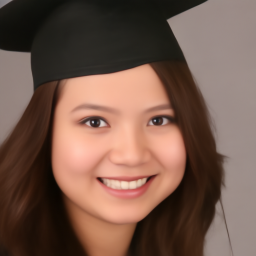}
		& \includegraphics[width=\imwidth,height=\imwidth]{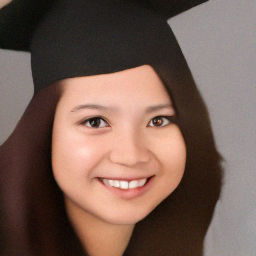}
		& \includegraphics[width=\imwidth,height=\imwidth]{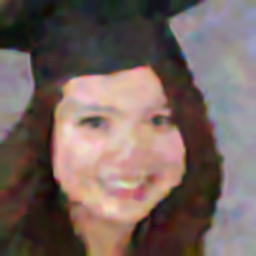}  \\
		\multicolumn{11}{c}{\vspace{-.2em}} \\
		\multirow{2}{*}{\raisebox{-.15cm}[0pt][0pt]{\rotatebox{90}{\scriptsize 1 bit}}} &
		\includegraphics[width=\imwidth,height=\imwidth]{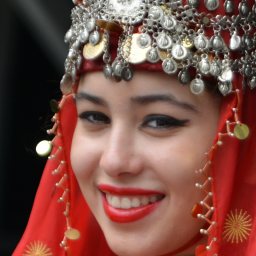}
		& \includegraphics[width=\imwidth,height=\imwidth]{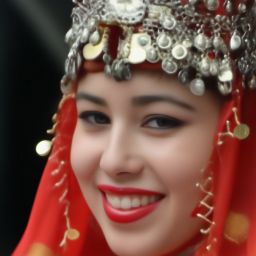}
		& \includegraphics[width=\imwidth,height=\imwidth]{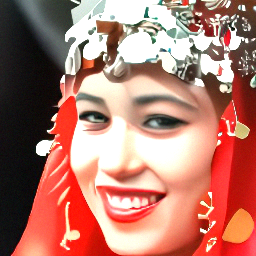}
		& \includegraphics[width=\imwidth,height=\imwidth]{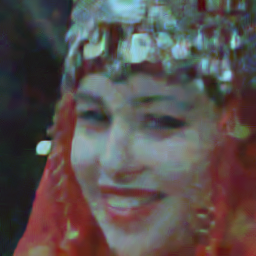}
		& 
		& \multirow{2}{*}{\raisebox{-.15cm}[0pt][0pt]{\rotatebox{90}{\scriptsize 1 bit}}}
		& \includegraphics[width=\imwidth,height=\imwidth]{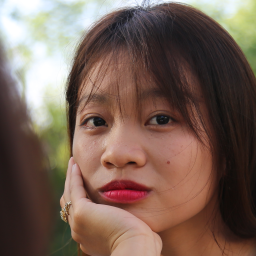}
		& \includegraphics[width=\imwidth,height=\imwidth]{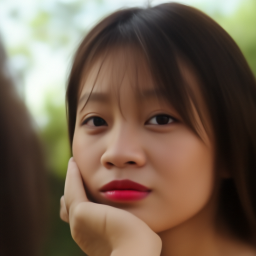}
		& \includegraphics[width=\imwidth,height=\imwidth]{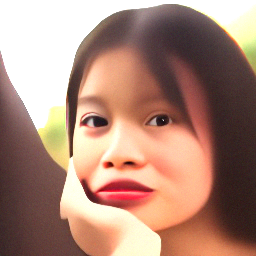}
		& \includegraphics[width=\imwidth,height=\imwidth]{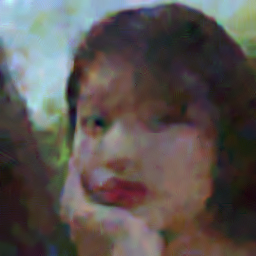} \\
		& 
		\includegraphics[width=\imwidth,height=\imwidth]{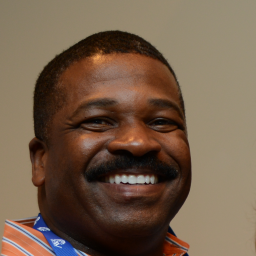}
		& \includegraphics[width=\imwidth,height=\imwidth]{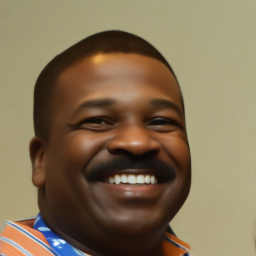}
		& \includegraphics[width=\imwidth,height=\imwidth]{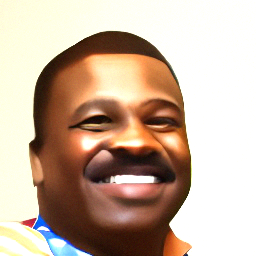}
		& \includegraphics[width=\imwidth,height=\imwidth]{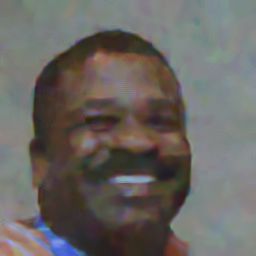}
		& & 
		& \includegraphics[width=\imwidth,height=\imwidth]{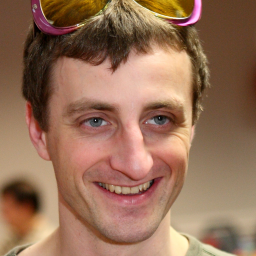}
		& \includegraphics[width=\imwidth,height=\imwidth]{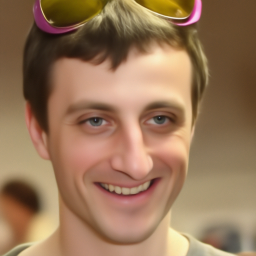}
		& \includegraphics[width=\imwidth,height=\imwidth]{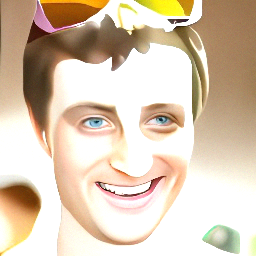}
		& \includegraphics[width=\imwidth,height=\imwidth]{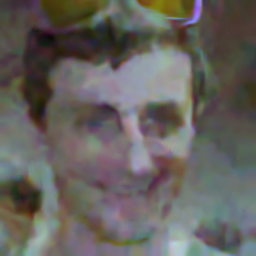}  \\
		& \scriptsize Ground truth & \scriptsize \ac{Q-STMP} & \scriptsize QCS-SGM & \scriptsize GTurbo-SR & \scriptsize ~~~~~~~~ & & \scriptsize Ground truth & \scriptsize \ac{Q-STMP} & \scriptsize QCS-SGM & \scriptsize GTurbo-SR
	\end{tabular}
	\caption{Examples of quantized compressive image recovery on FFHQ using randomly row-selected \ac{DCT} sensing matrices, where $\delta_0 = 0.1$.
		We set $\beta = 0.6$ for \ac{Q-STMP}.}
	\label{fig:quantized}
\end{figure*}

{\color{black}
	In Fig.~\ref{fig:major_svd_kappa}, we investigate the effect of the condition number $\kappa$ on the convergence behavior of \ac{STMP} for general unitarily-invariant sensing matrices constructed via \ac{SVD}.
	As $\kappa$ increases, the singular values become more ill-conditioned, leading to a slower convergence rate. 
	Nevertheless, \ac{STMP} remains stable and eventually attains similar reconstruction performance for all tested $\kappa$, demonstrating its robustness to non-partial-orthogonal sensing matrices.
	Fig.~\ref{fig:major_cg} further evaluates the practical implementation of the \ac{LMMSE} update using the \ac{CG} method.
	The results show that, for \ac{i.i.d.} Gaussian sensing matrices, only a few \ac{CG} iterations are sufficient to achieve nearly identical convergence behavior.
	Moreover, increasing the number of Hutchinson probe vectors brings negligible improvement.
	These observations confirm that the proposed \ac{CG}-based implementation provides an efficient and accurate alternative to explicit matrix inversion, with very low computational overhead.
}

\subsection{Results on Quantized Compressive Image Recovery}
In this subsection, we compare the \ac{Q-STMP} algorithm with the following baselines:
\begin{itemize}
	\item \textbf{QCS-SGM~\cite{qcs_sgm}:} We follow the original paper and use $2,311$ reverse-diffusion steps.
	We choose $\epsilon = 0.00005$ for the $1$-bit setting, and $\epsilon = 0.00002$ for all higher-bit quantization levels.
	\item \textbf{GTurbo-SR~\cite{GTurboSR}:} We choose the BM3D denoiser for GTurbo-SR. The algorithm is executed for $50$ iterations.
\end{itemize}

Fig.~\ref{fig:quantized_stmp_convergence} shows the \ac{NMSE} trajectories and corresponding \ac{SE} predictions for \ac{Q-STMP} on the FFHQ dataset under different sampling ratios and quantization bit depths.
Similar to \ac{STMP}, \ac{Q-STMP} exhibits rapid convergence, and the converged performance aligns closely with the \ac{SE} predictions.
Notably, increasing the number of quantization bits from $1$ to $3$ yields only marginal improvements in reconstruction performance. 
This suggests that the FFHQ image data possess significant redundancy, allowing \ac{Q-STMP} to recover meaningful image content even from extremely coarsely quantized observations.

Fig.~\ref{fig:quantized} presents representative results for quantized compressive image recovery on the FFHQ dataset under different sampling ratios and quantization bit depths.
Across all settings, \ac{Q-STMP} consistently produces sharp and faithful reconstructions, even at extremely low bit levels such as $1$ or $2$ bits.
In contrast, competing approaches including QCS-SGM and GTurbo-SR often introduce noticeable artifacts, oversmoothing, or hallucinated details, with degradation becoming more severe as the quantization becomes coarser.
The robustness of \ac{Q-STMP} to both low sampling ratios and aggressive quantization highlights the strong expressive power of the score-based generative prior and the effectiveness of the \ac{TMP} and \ac{EP} frameworks in handling heavily distorted measurements.
Table~\ref{table:q_stmp} summarizes the quantitative performance of all methods in terms of PSNR, SSIM, FID, and LPIPS.

\begin{table*}[t]
	\centering
	\caption{Quantitative Results for Quantized Compressive Image Recovery on FFHQ Using Randomly Row-selected \ac{DCT} Sensing Matrices 
		\\ (\textbf{Bold}: Best, \underline{Underline}: Second Best; $\delta_0 = 0.1$)}
	\vspace{-1em}
	\begin{center}
		\begin{tabular}{c l p{0.0001\textwidth} cccc p{0.0001\textwidth} cccc}
			\toprule
			\multirow{2}{*}{Quantization} & \multirow{2}{*}{Method} & ~ &  \multicolumn{4}{c}{$M/N = 0.8$} & ~ & \multicolumn{4}{c}{$M/N = 0.5$} \\
			\cline{4-7} \cline{9-12} 
			& & ~ & PSNR$\uparrow$ & SSIM$\uparrow$ & FID$\downarrow$ & LPIPS$\downarrow$ & ~ & PSNR$\uparrow$ & SSIM$\uparrow$ & FID$\downarrow$ & LPIPS$\downarrow$ \\
			\midrule
			\multirow{3}{*}{3 bits} & \ac{Q-STMP} (Ours)  & ~ & $\mathbf{29.88}$ & $\mathbf{0.8660}$ & $\mathbf{44.98}$ & $\mathbf{0.1149}$ & ~ & $\mathbf{28.93}$ & $\mathbf{0.8453}$ & $\mathbf{48.74}$ & $\mathbf{0.1353}$ \\
			& QCS-SGM  & ~ &  $\underline{26.88}$ & $\underline{0.7497}$ & $\underline{48.33}$ & $\underline{0.1514}$ & ~ & $\underline{26.02}$ & $\underline{0.7206}$ & $\underline{56.55}$ & $\underline{0.1879}$  \\
			& GTurbo-SR  & ~ & $22.81$ & $0.6969$ & $64.55$ & $0.3158$ & ~ & $22.98$ & $0.6811$ & $88.68$ & $0.3998$  \\
			\midrule
			\multirow{3}{*}{2 bits} & \ac{Q-STMP} (Ours)  & ~ &  $\mathbf{28.52}$ & $\mathbf{0.8364}$ & $\mathbf{48.78}$ & $\mathbf{0.1421}$ & ~ &  $\mathbf{27.13}$ & $\mathbf{0.7919}$ & $\mathbf{48.65}$ & $\mathbf{0.1674}$  \\
			& QCS-SGM  & ~ & $\underline{24.86}$ & $\underline{0.7121}$ & $\underline{56.84}$ & $\underline{0.1943}$ & ~ & $\underline{23.95}$ & $\underline{0.6791}$ & $\underline{65.90}$ & $\underline{0.2351}$  \\
			& GTurbo-SR  & ~ & $20.70$ & $0.6332$ & $116.84$ & $0.4669$ & ~ &  $20.14$ & $0.5884$ & $164.07$ & $0.5530$  \\
			\midrule
			\multirow{3}{*}{1 bit} & \ac{Q-STMP} (Ours)  & ~ & $\mathbf{27.39}$ & $\mathbf{0.8268}$ & $\mathbf{48.04}$ & $\mathbf{0.1482}$ & ~ & $\mathbf{25.79}$ & $\mathbf{0.7411}$ & $\mathbf{47.10}$ & $\mathbf{0.1957}$  \\
			& QCS-SGM  & ~ & $12.17$ & $0.5685$ & $\underline{99.61}$ & $\underline{0.3573}$ & ~ &  $11.34$ & $0.5230$ & $\underline{117.85}$ & $\underline{0.4109}$  \\
			& GTurbo-SR  & ~ & $\underline{18.72}$ & $\underline{0.6061}$ & $133.15$ & $0.4946$ & ~ & $\underline{18.10}$ & $\underline{0.5610}$ & $185.61$ & $0.5626$  \\
			\bottomrule
		\end{tabular}
	\end{center}
	\label{table:q_stmp}
\end{table*}

\begin{figure}[t]
	\centering
	\begin{minipage}{0.493\linewidth}
		\centering
		\includegraphics[width=\linewidth]{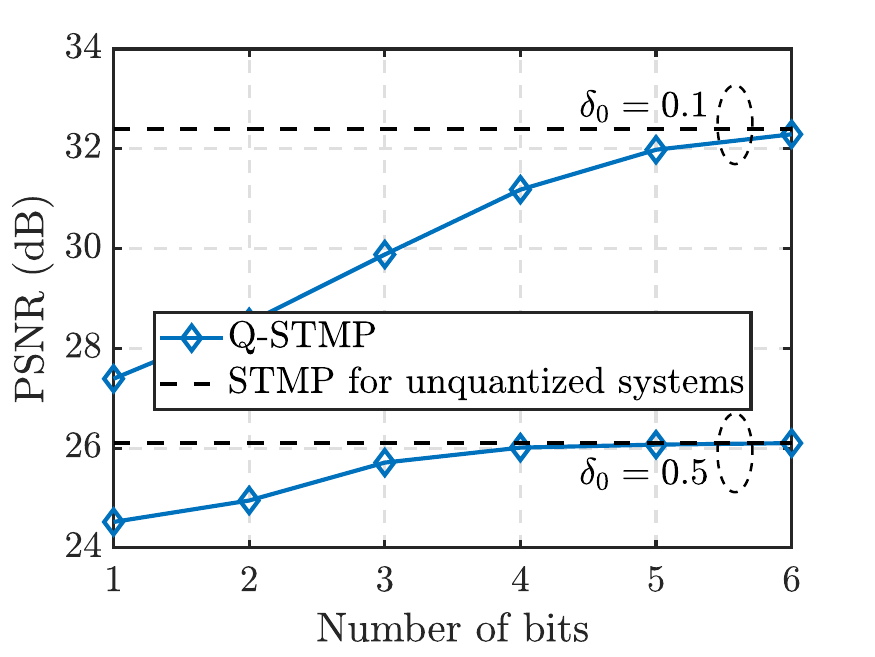}\\
		\footnotesize{(a) PSNR vs. quantization bit depth}
	\end{minipage}
	\begin{minipage}{0.493\linewidth}
		\centering		
		\includegraphics[width=\linewidth]{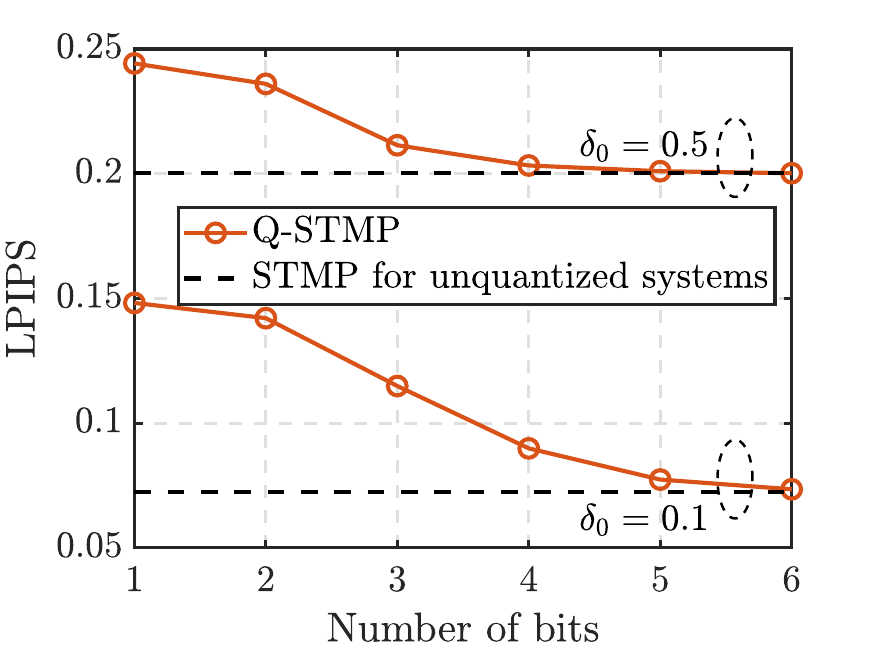}\\
		\footnotesize{(b) LPIPS vs. quantization bit depth}
	\end{minipage}
	\caption{\label{fig:qstmp_bits} 
		Performance of \ac{Q-STMP} on FFHQ under different quantization bit depths for a sampling ratio of $M/N = 0.8$.
		Randomly row-selected \ac{DCT} sensing matrices are used.
		Dashed lines indicate the performance of \ac{STMP} in the unquantized setting.}
\end{figure}

Fig.~\ref{fig:qstmp_bits} illustrates how the reconstruction quality of Q\mbox{-}\ac{STMP} varies with the quantization bit depth.
As the number of bits increases, both PSNR and LPIPS improve steadily, with notable gains from $1$ to $4$ bits and diminishing returns thereafter.
Notably, \ac{Q-STMP} already approaches the performance of \ac{STMP} for unquantized systems even with moderate bit depths ($4$ to $6$ bits), especially for the higher-noise setting $\delta_0 = 0.5$.
These results highlight the robustness of \ac{Q-STMP} to coarse quantization and demonstrate that only a small number of bits is sufficient to achieve near-optimal reconstruction fidelity.

\section{Conclusions} \label{sec:conclusion}
In this paper, we introduced \ac{STMP}, a novel approach for compressive image recovery using unconditional score networks.
STMP inherits the fast convergence of message-passing algorithms while taking full advantage of the expressive prior through the integration of score-based generative models.
To deal with quantized measurements, we further proposed Q-STMP, which augments STMP with a component-wise MMSE dequantizer.
We presented the \ac{SE} equations to characterize the asymptotic performance of \ac{STMP} and \ac{Q-STMP}.
Experimental results on the FFHQ and LDCT datasets under diverse sensing models highlight the efficiency and state-of-the-art performance of \ac{STMP} and \ac{Q-STMP} against a variety of benchmarks.

\bibliographystyle{IEEEtran}
\bibliography{example_paper}

\begin{IEEEbiography}[{\includegraphics[width=1in,height=1.25in,clip,keepaspectratio]{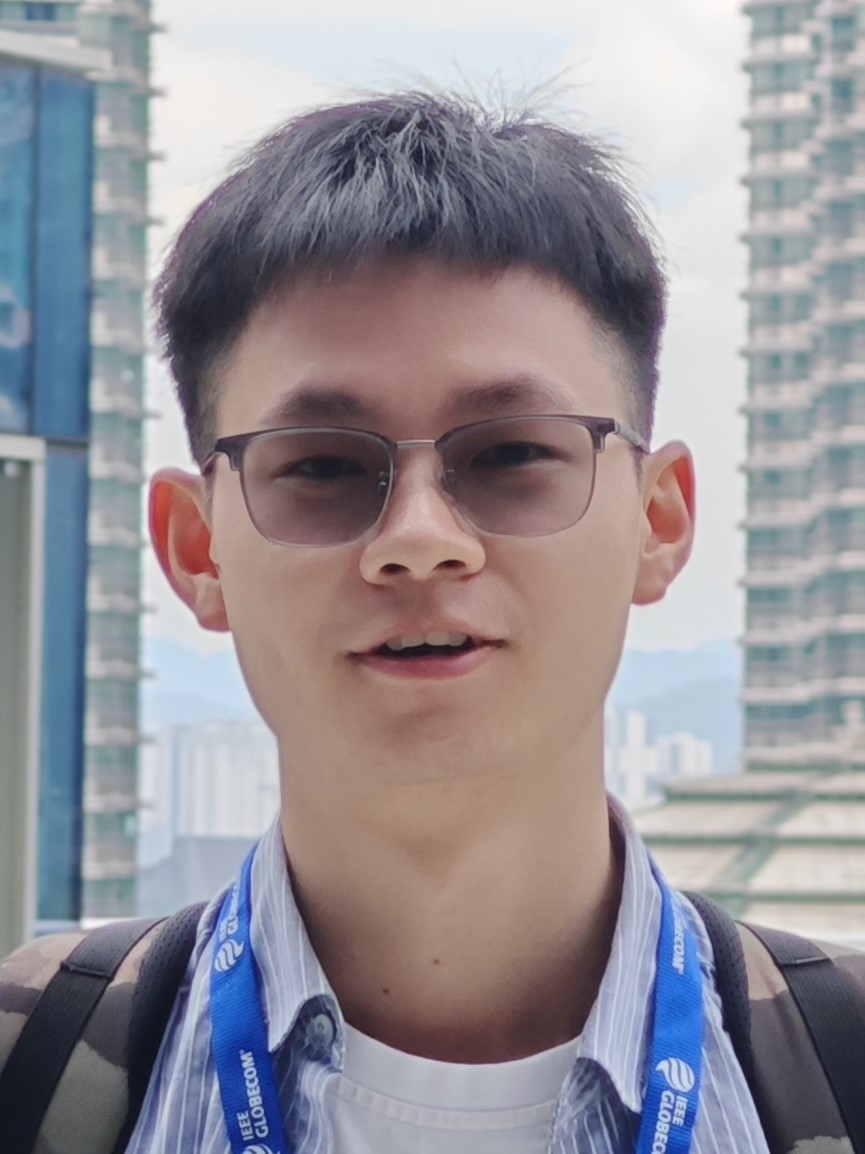}}]{Chang Cai}
	(Member, IEEE) received the B.Eng. degree from the Yingcai Honors College, University of Electronic Science and Technology of China, in 2021,
	and the Ph.D. degree from the Department of Information Engineering, The Chinese University of Hong Kong, in 2025.
	He is currently a Research Assistant Professor with the Department of Electrical and Computer Engineering, The University of Hong Kong.
	His research interests include generative AI, Bayesian inference, and semantic communications.
	He served as the Managing Editor for the IEEE Open Journal of the Communications Society from 2022 to 2023. 
\end{IEEEbiography}

\begin{IEEEbiography}[{\includegraphics[width=1in,height=1.25in,clip,keepaspectratio]{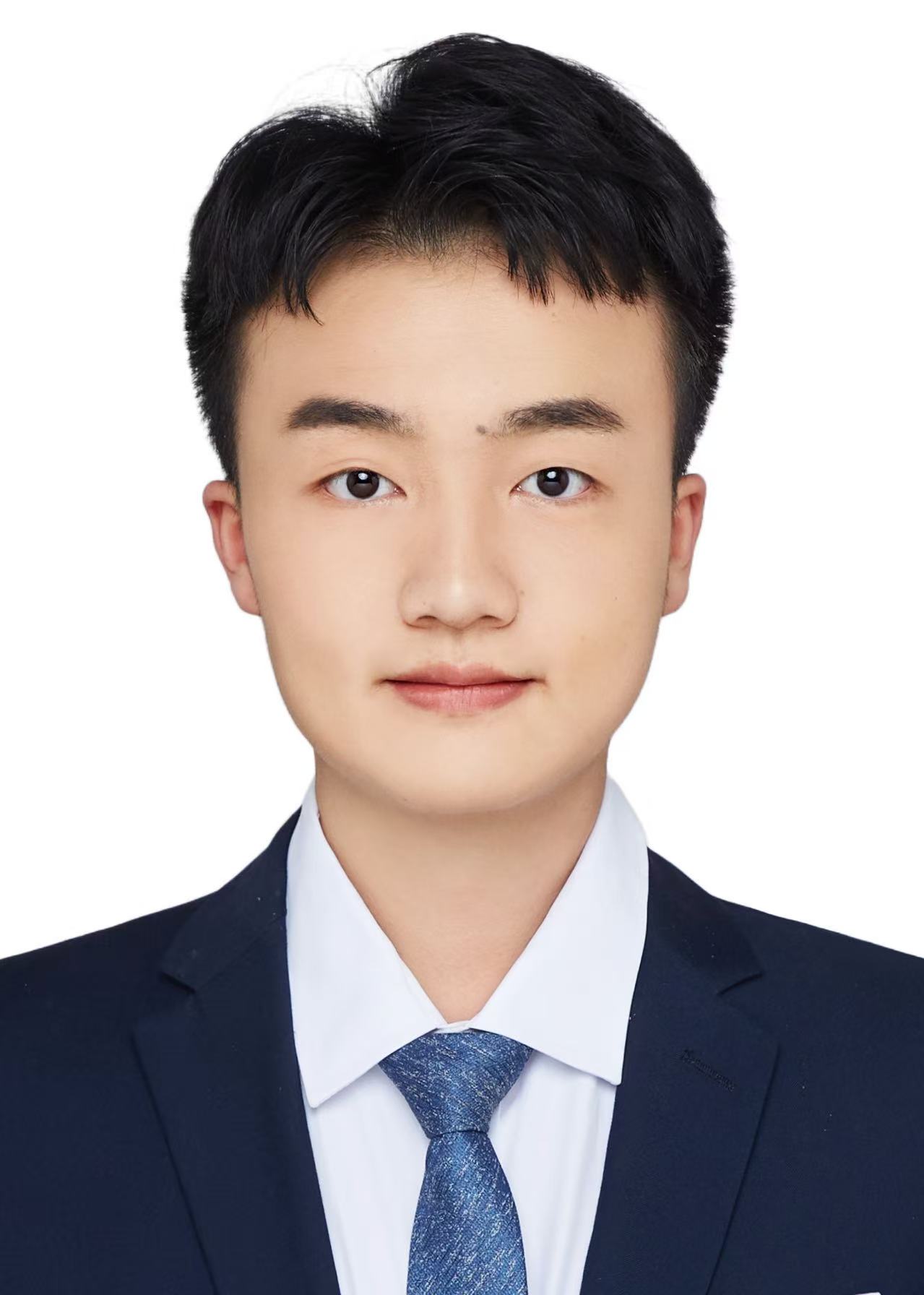}}]{Hao Jiang}
	(Graduate Student Member, IEEE) received the B.Eng. degree in electronic information engineering and the B.Mgmt. degree in electronic commerce from the University of Electronic Science and Technology of China, Chengdu, China, in 2022, where he is currently working toward the Ph.D. degree with the National Key Laboratory of Wireless Communications. From 2025 to 2026, he was a visiting student with the Singapore University of Technology and Design, Singapore. His research interests include signal processing and machine learning for 6G wireless communications.
\end{IEEEbiography}

\begin{IEEEbiography}[{\includegraphics[width=1in,height=1.25in,clip,keepaspectratio]{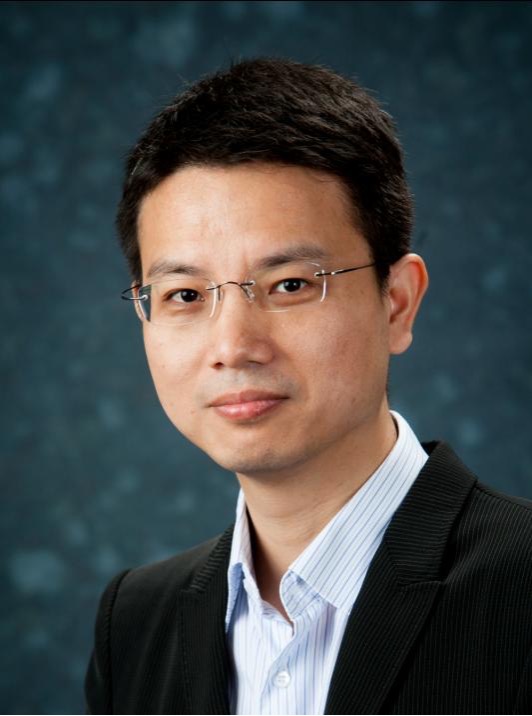}}]{Xiaojun Yuan}
	(Fellow, IEEE) received the Ph.D. degree in Electrical Engineering from the City University of Hong Kong in 2009. From 2009 to 2011, he was a research fellow at the Department of Electronic Engineering, the City University of Hong Kong. He was also a visiting scholar at the Department of Electrical Engineering, the University of Hawaii at Manoa in spring and summer 2009, as well as in the same period of 2010. From 2011 to 2014, he was a research assistant professor with the Institute of Network Coding, The Chinese University of Hong Kong. From 2014 to 2017, he was an assistant professor with the School of Information Science and Technology, ShanghaiTech University. He is now a professor with the National Key Laboratory of Wireless Communications, the University of Electronic Science and Technology of China.
	
	His research interests cover a broad range of signal processing, machine learning, and wireless communications, including but not limited to intelligent communications, structured signal reconstruction, Bayesian approximate inference, distributed learning, etc. He has published over 300 peer-reviewed research papers in the leading international journals and conferences in the related areas. He has served on many technical programs for international conferences. He was an editor of IEEE leading journals, including IEEE Transactions on Wireless Communications and IEEE Transactions on Communications. He was a co-recipient of IEEE Heinrich Hertz Award 2022, and a co-recipient of IEEE Jack Neubauer Memorial Award 2025.
\end{IEEEbiography}

\begin{IEEEbiography}[{\includegraphics[width=1in,height=1.25in,clip,keepaspectratio]{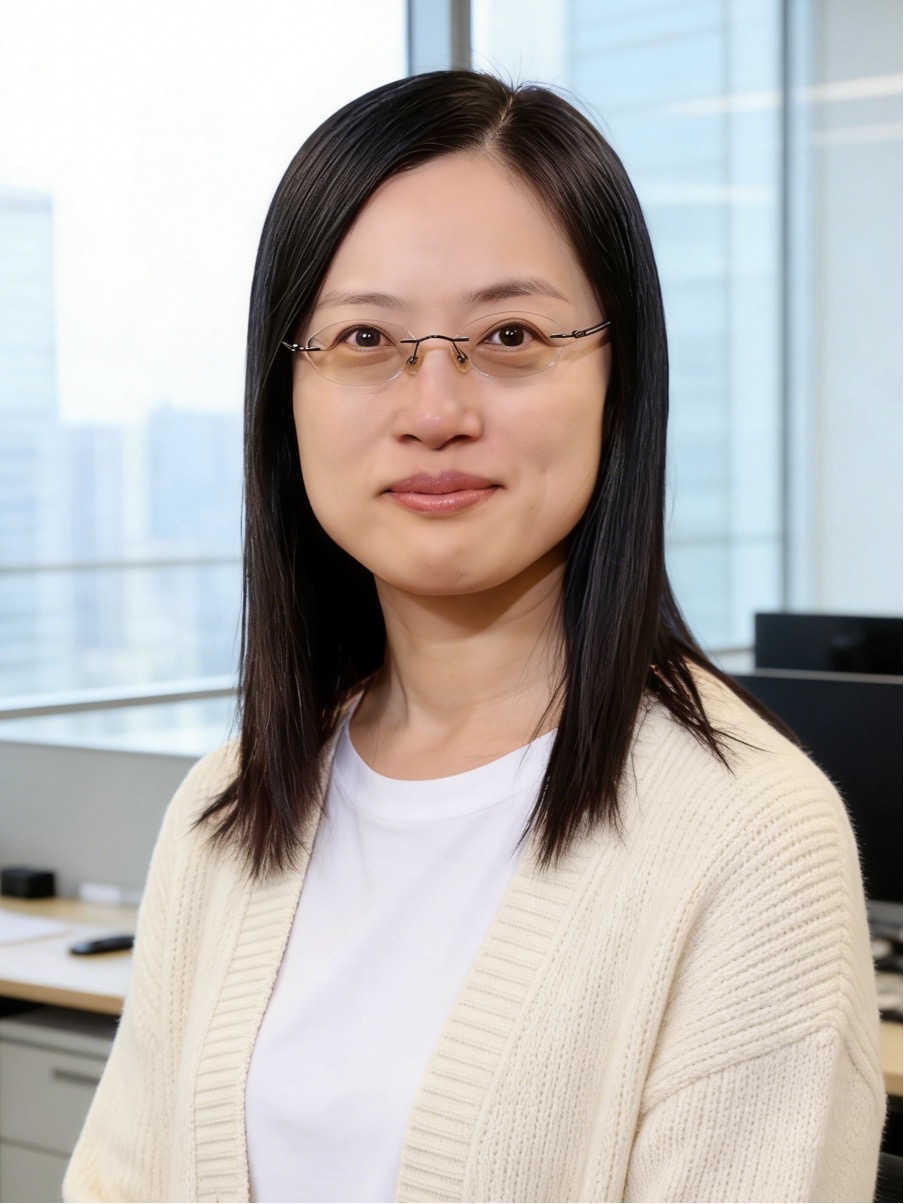}}]{Ying-Jun Angela Zhang}
	(Fellow, IEEE) received her Ph.D. degree from the Department of Electrical and Electronic Engineering, The Hong Kong University of Science and Technology. She joined the Department of Information Engineering, The Chinese University of Hong Kong in 2005, where she is now a professor. 
	Prof. Zhang is now the Steering Committee Chair of IEEE Wireless Communication Letters, Director of IEEE ComSoc Asia Pacific Region Board, and member of IEEE VTS Fellow Evaluation Standing Committee. Previously, she served as Editor-in-Chief of IEEE Open Journal of the Communications Society, Chair of the Executive Editor Committee of IEEE Transactions on Wireless Communications, and member of IEEE ComSoc Fellow Evaluation Standing Committee. Prof. Zhang has served on the Organizing Committees of many top conferences, such as IEEE GLOBECOM, ICC, VTC, SmartgridComm, and was the Chair of IEEE ComSoc Technical Committee of Smart Grid Communications.
	
	Prof. Zhang is a co-recipient of 2021 and 2014 IEEE ComSoc Asia Pacific Outstanding Paper Awards, 2013 IEEE SmartgridComm Best Paper Award, and 2011 IEEE Marconi Prize Paper Award on Wireless Communications. As the only winner from engineering science, Prof. Zhang won the Hong Kong Young Scientist Award 2006, conferred by the Hong Kong Institute of Science.
\end{IEEEbiography}

\end{document}